\documentclass{article} 
\usepackage[final]{colm2026_conference}

\usepackage{microtype}
\usepackage{hyperref}
\usepackage{tcolorbox}
\tcbuselibrary{skins, breakable}
\usetikzlibrary{shadows}

\usepackage{url}

\fancypagestyle{firstpagefooter}{%
  \fancyhf{}%
  \lhead{Paper Under Review}%
  \renewcommand{\headrulewidth}{1.5pt}%
}
\usepackage{xcolor}
\usepackage{array}
\usepackage{amsmath}
\usepackage{amssymb}

\usepackage{colortbl}
\usepackage{longtable}
\usepackage{booktabs}
\usepackage{graphicx}
\usepackage{caption}
\usepackage{float}
\usepackage{placeins}
\usepackage{algorithm}
\usepackage{algpseudocode}
\definecolor{darkblue}{rgb}{0, 0, 0.5}
\definecolor{botred}{RGB}{180, 50, 50}
\definecolor{botredbg}{RGB}{255, 241, 241}
\definecolor{Anchor87}{RGB}{255, 210, 120}
\definecolor{Anchor83}{RGB}{255, 185, 155}
\definecolor{Anchor82}{RGB}{205, 190, 255}
\definecolor{Anchor77}{RGB}{160, 215, 235}
\definecolor{Anchor74}{RGB}{185, 195, 225}
\definecolor{Anchor62}{RGB}{140, 210, 175}
\definecolor{AnchorOver}{RGB}{255, 130, 115}
\definecolor{AnchorUnder}{RGB}{95, 210, 160}
\hypersetup{colorlinks=true, citecolor=darkblue, linkcolor=darkblue, urlcolor=darkblue}

\title{HumorRank: A Tournament-Based Leaderboard for Evaluating Humor Generation in Large Language Models\thanks{Live leaderboard: \url{https://humorrank-leaderboard.pages.dev/}.}}

\author{
Edward Ajayi \\
Carnegie Mellon University Africa \\
Kigali, Rwanda \\
\texttt{eaajayi@andrew.cmu.edu}
\And
Prasenjit Mitra \\
Carnegie Mellon University Africa \\
Kigali, Rwanda \\
\texttt{prasenjm@andrew.cmu.edu}
}

\begin{document}
\raggedbottom


\maketitle
\thispagestyle{firstpagefooter}


\begin{abstract}
Evaluating humor in large language models (LLMs) is an open challenge because existing approaches yield isolated, incomparable metrics rather than unified model rankings, making it difficult to track progress across systems. We introduce HumorRank, a tournament-based evaluation framework and leaderboard for textual humor generation. On two public benchmarks (SemEval-2026 MWAHAHA and Humor Transfer Bench), we conduct extensive automated pairwise evaluation across nine models spanning proprietary, open-weight, and specialized systems. Pairwise judgments are produced by LLM judges grounded in the General Theory of Verbal Humor (GTVH): each judge integrates structured comedic analysis into adjudication, jointly yielding a preference decision, an interpretable rationale, and mechanism, delivery, and failure tags rather than a black-box funniness score. Judgments are aggregated via an Adaptive Swiss tournament, with Bradley--Terry Maximum Likelihood Estimation (MLE) producing globally consistent humor generation capability rankings. Rankings are \textit{cross-judge stable}: independent LLM judges (Llama 3.3 70B and Qwen 2.5 72B) yield Kendall $\tau = 0.889$ on both benchmarks, and a human calibration study shows human--LLM agreement tracking human--human agreement on hard funny-versus-funny pairs. Our results demonstrate that HumorRank yields statistically grounded model stratifications, showing that humor quality is associated with mastery of comedic mechanisms such as incongruity, conciseness, escalation, and absurdity rather than model scale alone, with specialized fine-tuned models reaching parity with far larger systems. HumorRank thus provides a scalable, interpretable, and reproducible methodology for benchmarking and understanding LLM-generated humor.
\end{abstract}

\section{Introduction}

Humor generation is a domain that requires a highly nuanced understanding of language, context, and pragmatic reasoning~\citep{quan2025can, kim2025ai}, posing a significant challenge for evaluating the capabilities of large language models (LLMs) in generating humor~\citep{narad2025llms}. This difficulty is reflected in the fragmented landscape of existing evaluation methods, where different studies adopt incompatible paradigms~\cite{ajayiautomatic}, including punchline detection~\citep{romanowski2025punchlines}, scalar scoring~\citep{goes2022crowd}, humor classification~\citep{wu2025humour}, LLM-as-a-Judge approaches~\citep{shafiei2025not}, and costly human preference evaluations~\citep{romanowski2025punchlines, horvitz2024getting}.

A central limitation of these approaches is the lack of a unified and scalable framework for comparing different humor-generation systems. Existing methods evaluate different aspects or types of generated humor but do not produce comparable system-level rankings, making it difficult to track progress in computational humor generation. Scaling comparisons across humor-generation systems is further hindered by systematic failures when LLMs judge generated jokes: \textit{score anchoring} and \textit{ranking collapse} under absolute rubrics, \textit{family bias} and self-preference when judges evaluate jokes produced by their own model family for the same prompt, and \textit{verbosity bias} and \textit{spurious ties} under naive pairwise joke comparison. We document these pathologies across judge paradigms and model families in Appendix~\ref{sec:appendix_judge_failures}.

As LLMs increasingly power chatbots and conversational assistants, appropriate humor can support rapport, engagement, and more natural human--AI interaction. A reliable and comparable protocol for evaluating the humor-generation capabilities of the underlying LLMs therefore becomes essential. To address this gap, we introduce \textbf{HumorRank}, a leaderboard-oriented framework that combines GTVH-structured pairwise judging, budget-aware Adaptive Swiss tournament scheduling, and Bradley--Terry aggregation to produce scalable, globally consistent model rankings with structured analyses of comedic strengths and failure modes. To our knowledge, HumorRank is the first end-to-end, fully automated, theory-grounded framework designed specifically for global capability ranking of humor-generation models and reuse across benchmarks. The central design choice is to separate \emph{which comparisons are performed} from \emph{how the final ranking is estimated}: a budget-aware Adaptive Swiss Pairing schedule uses provisional standings to prioritize close, under-sampled matchups while avoiding repeated pairings, whereas final ratings are estimated globally using Bradley--Terry maximum likelihood estimation (MLE). This separates adaptive pairing during the tournament from order-independent final rating estimation, enabling reduced-budget schedules without using provisional pairing scores as final ratings. Each GTVH-grounded duel yields not only an A/B/TIE preference, but also a brief comparative rationale, humor-mechanism and delivery tags attributed to the winning joke, and failure-mode tags attributed to the loser. Aggregated across the tournament, these annotations produce model-level comedic profiles that indicate how systems tend to succeed or fail, rather than providing only an ordinal ranking. We evaluate nine models on Humor Transfer Bench (HTB)~\citep{ajayi2026humorgen} and SemEval-2026 Task 1: MWAHAHA~\citep{semeval2026mwahaha}, showing that reduced-budget Swiss scheduling retains strong rank fidelity while independent Llama and Qwen judges produce stable model orderings.

    
    

Our contributions are as follows:
\begin{enumerate}
    \item We introduce \textbf{HumorRank}, the first end-to-end, fully automated, theory-grounded framework for global capability ranking of humor-generation models, designed for reuse across models, benchmarks, and humor domains.
    
    \item We formalize humor assessment as a \textbf{pairwise preference learning task} and use Bradley--Terry estimation for stable, comparable global rankings. To make this practical at scale, we pair it with a \textbf{budget-aware Adaptive Swiss Pairing} strategy and demonstrate strong rank fidelity under reduced comparison budgets.
    
    \item We develop a \textbf{GTVH-grounded pairwise LLM-judge protocol for humor evaluation} that jointly outputs a preference decision, brief comparative rationale, winner mechanism and delivery annotations, and loser failure modes. Aggregating these signals yields interpretable model-level comedic profiles beyond scalar ratings or ordinal ranks.
\end{enumerate}

\section{Related Works}



\subsection{Model Evaluation in Humor Generation Systems}
\label{sec:related_humor_eval}

Despite growing interest in LLM humor capabilities, evaluation protocols remain inconsistent and difficult to compare across studies. Prior work spans automated metrics for human-AI co-creative humor~\cite{wu2025one}, crowd-sourced AI voting panels~\cite{goes2022crowd}, Best-Worst Scaling (BWS)~\cite{yamane2024generic}, Likert-style funniness templates~\cite{gorenz2024funny}, and fully human evaluation, which is costly and usually limited to small validation sets~\cite{zhang2024humor, goel2024automating, wang_innovative_2025, jain2024ai}. Broader evaluations of LLM humor understanding and generation~\cite{ajayiautomatic, zhou2025bridging, song2025large} extend these paradigms across task formulations. However, these approaches typically yield task-specific scores rather than \emph{ranked preference orderings} across multiple humor generation systems, and they rarely provide interpretable comparative rationales. As a result, evaluation is often a one-off measurement rather than a scalable comparative framework. Existing humor benchmarks and evaluation paradigms have explored increasingly diverse tasks, from humor understanding to generation and ranking.

\subsection{Computational Humor: Datasets, Theory, and Generation}
\label{sec:related_humor_benchmarks}

Humor is rooted in psychology~\cite{larkin2017overview} and linguistics~\cite{attardo2024linguistic}, with classical theories such as superiority, relief, and incongruity explaining humor~\cite{veatch1998theory}. These frameworks motivate interpretable dimensions of humor, such as expectation violation, tension release, and social positioning. However, they do not provide a deterministic recipe for generation~\cite{larkin2017overview}, as humor varies across context, culture, and individual perception. Linguistic and pragmatic analysis identifies relatively stable cues such as timing, delivery, ambiguity, and form--meaning incongruity, which support dataset construction and automated evaluation. Building on these foundations, prior work has introduced a range of humor benchmarks, historically focused on text-based tasks. More recently, advances in large language models have expanded this landscape to include multimodal datasets and evaluation settings spanning humor generation, understanding, and ranking~\cite{zhong_lets_2024,zhang2024humor,he2024chumor,ryan2025humor,jain2024ai}. These developments broaden the empirical scope of computational humor, yet they still leave open how to aggregate pairwise outcomes into a stable, cross-model leaderboard under scalable automated judging.

\subsection{LLM Leaderboard Rating Systems in NLP Tasks}
\label{sec:related_leaderboards}

Leaderboard-based evaluation has become prevalent in NLP, providing a standardized framework for comparing model performance across tasks and benchmarks~\cite{toloka_llm_leaderboard_2023, chiang2024chatbot, myrzakhan2024open}. Modern leaderboard platforms, such as Chatbot Arena~\cite{chiang2024chatbot} and the Open LLM Leaderboard~\cite{silva2026clarin}, often leverage \textit{LLM-as-a-Judge} paradigms~\cite{zheng2023judging} to enable scalable evaluation of model outputs. This approach supports both human and model-based preference judgments, enabling flexible evaluation. Furthermore, leaderboard-based systems facilitate direct comparison of models under consistent conditions, making them suitable for benchmarking progress in NLP~\cite{federiakin2025improving, myrzakhan2024open}. Prior work suggests that such ranking frameworks provide a reliable proxy for model quality and can be adapted to diverse settings, including multilingual and domain-specific evaluation scenarios~\cite{park2024open, silva2026clarin}. HumorRank applies a GTVH-grounded pairwise tournament pipeline with budget-aware Adaptive Swiss pairing to this setting. Appendix~\ref{sec:appendix_judge_failures} documents judge configurations that fail on humor and the validation criteria applied in our experiments.

\section{HumorRank}

The subjective and multidimensional nature of humor presents fundamental challenges for absolute quality scoring. To address this, we operationalize humor as a continuous \textit{cognitive reward} arising from the successful resolution of deliberately constructed linguistic incongruities; the full definition and derivation are in Appendix~\ref{sec:appendix_humor_definition}. Because lexical and semantic humor features (e.g., comedic delivery)~\citep{romanowski2025punchlines, kim2025ai} interact in ways that resist direct quantification~\citep{winters2025evaluating}, pairwise comparison mitigates these limitations~\citep{ravi2024small} by constraining evaluation to a relative preference judgment between two model-generated jokes conditioned on the same prompt~\citep{hossain2020semeval}. This formulation reduces cognitive load on the evaluator and is more robust to inter-annotator variance than uncalibrated scalar annotation.

While pairwise comparisons provide high-fidelity local signal, they are still discrete and unordered, and thus insufficient on their own to support a system-level leaderboard. To transform a collection of $\binom{K}{2}$ pairwise outcomes over $K$ competing models into a globally consistent capability ranking, an aggregation framework must resolve local inconsistencies and propagate information across the full tournament graph.
\textbf{HumorRank} addresses this through a two-stage pipeline: an Adaptive Swiss Tournament that efficiently builds the pairwise comparison graph, followed by global Bradley--Terry (BT) Maximum Likelihood Estimation (MLE) that maps observed outcomes to statistically grounded, continuous capability estimates. We additionally report Stable Elo ratings as a secondary reference metric for cross-validation.

\begin{figure}[!ht]
    \centering
    \includegraphics[width=0.95\columnwidth]{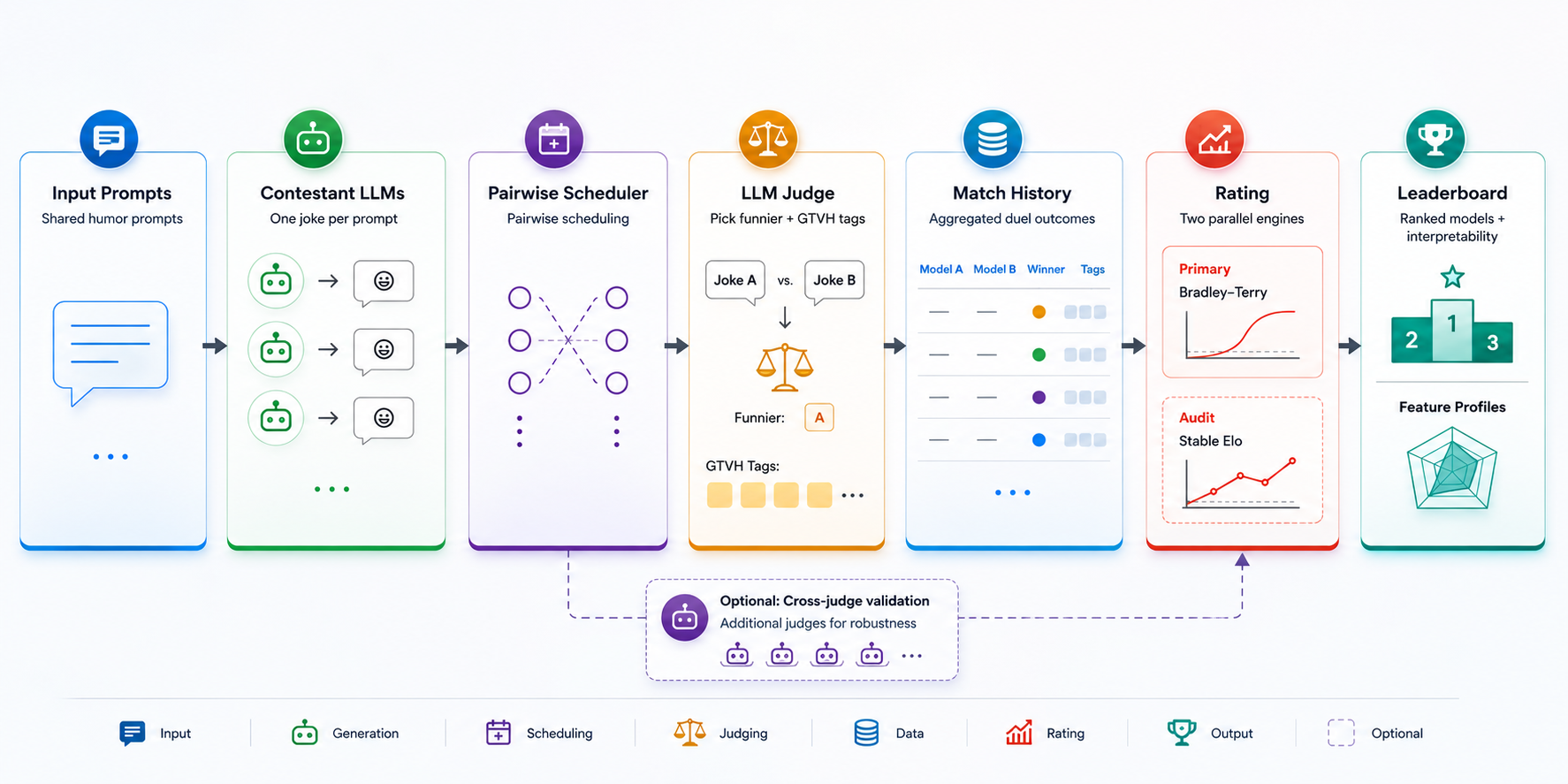}
    \caption{Overview of the HumorRank evaluation pipeline. Contestant models generate jokes on shared prompts, Adaptive Swiss pairing schedules pairwise duels, a GTVH-grounded LLM judge adjudicates each duel, and outcomes are aggregated into match history and converted to global Bradley--Terry ratings (primary) and Stable Elo scores (audit), producing a ranked leaderboard with interpretable humor-feature profiles.}
    \label{fig:architecture}
\end{figure}

\subsection{Pairwise Comparison}
\label{sec:why_pairwise}

Humor evaluation in this setting differs from standard LLM-as-a-Judge tasks (e.g., instruction following, summarization): SemEval-2026 MWAHAHA requires ranking \textit{multiple model-generated jokes on the same input}, which is inherently relative. For each input, every model generates a joke. We therefore collect pairwise preferences over those jokes (Model A vs.\ Model B on the same input) and aggregate outcomes with Bradley--Terry MLE~\cite{hunter2004mm} into a global leaderboard. Alternative judge configurations and ablation details are documented in Appendix~\ref{sec:appendix_judge_failures}.

\paragraph{General Theory of Verbal Humor-Guided Pairwise Formulation:}
\label{sec:gtvh_formulation}

To ensure the LLM-as-a-judge protocol transcends arbitrary preference, HumorRank's evaluation is formally grounded in the General Theory of Verbal Humor (GTVH)~\citep{ruch1993toward}. The GTVH parameterizes a joke $J$ not as a monolithic text, but as a hierarchical tuple of six Knowledge Resources (KRs): $J = \{SO, LM, SI, TA, NS, LA\}$. Here, the highest-order KRs are the \textit{Script Opposition} ($SO$, core semantic incongruity) and \textit{Logical Mechanism} ($LM$, cognitive resolution). Lower-order KRs govern surface presentation, such as \textit{Narrative Strategy} ($NS$, structure) and \textit{Language} ($LA$, lexical delivery).

Because LLMs can struggle with holistic, zero-shot humor evaluation due to alignment biases (Appendix~\ref{sec:appendix_judge_failures}), HumorRank does not prompt the judge for a black-box subjective scalar $P(J_A \succ J_B)$. Instead, we formulate the judge as an explicit feature-extraction function $E_\theta(J)$ over the theoretical KR space. 

The judge must instantiate categorical tags that map directly to the GTVH hierarchy (prompt details in Appendix~\ref{sec:appendix_judge_prompts}):
\begin{itemize}
    \item \textbf{Deep Structure ($SO, LM$):} The \texttt{HUMOR\_MECHANISMS} array captures $SO$ (e.g., \textit{incongruity}) and $LM$ (e.g., \textit{wordplay}, \textit{absurdity}).
    \item \textbf{Surface Presentation ($NS, LA$):} The \texttt{DELIVERY\_FEATURES} array captures $NS$ (e.g., \textit{framing commitment}) and $LA$ (e.g., \textit{timing}, \textit{conciseness}).
\end{itemize}

Thus, the pairwise decision is formulated over the same theoretical feature space: $P(J_A \succ J_B) = f_\theta(E_\theta(J_A), E_\theta(J_B) \mid \text{prompt})$. Feature extraction is integrated into the adjudication prompt rather than applied post hoc. The judge jointly returns the preference decision, comparative rationale, and GTVH-aligned feature annotations in one structured response.

\subsection{Bradley-Terry Global Maximum Likelihood Estimation}

The Bradley-Terry model~\citep{bradley1952rank} serves as our primary, order-independent ranking algorithm. By maximizing the likelihood of observed pairwise outcomes across the full tournament graph, BT estimates a latent ``humor capability'' score for each model. Preference-based BT-style rating has also been widely used in non-humor LLM evaluation settings, including arena-style leaderboards~\citep{chiang2024chatbot, myrzakhan2024open}.

Given models $i$ and $j$ with latent ratings $R_i$ and $R_j$, the probability that model $i$ wins over model $j$ is formulated as an Elo-scaled logistic function:
\begin{equation}
    P(\text{i wins against j}) = \frac{1}{1 + 10^{(R_j - R_i)/400}}
\end{equation}

Instead of sequential updates, HumorRank fits global MLE using the iterative Minorization-Maximization (MM) algorithm~\cite{hunter2004mm} until convergence ($\epsilon < 10^{-6}$), with ratings anchored at 1000. To quantify uncertainty in model separation, we report 95\% confidence intervals via 200 bootstrap resamples of the match history. Resampling and tournament configuration details are listed in Appendix~\ref{sec:appendix_hyperparameters} (Table~\ref{tab:hyperparams_tournament}).

\subsection{Stable Elo (Sequential Reference)}

While the BT model provides the global MLE, we simultaneously compute a sequential Elo rating~\cite{albers2001elo} to track dynamic stability and provide a secondary reference metric. The generalized sequential update rule is:
\begin{equation}
R_{\text{new}} = R_{\text{curr}} + K_{fac} \cdot (S - E)
\end{equation}
where $K_{fac}=32$ specifies the maximum volatility factor, $S$ denotes the observed outcome ($1.0$ for a win, $0.5$ for a tie, $0.0$ for a loss), and $E$ is the expected probability derived from Equation 1.

A known deficiency of standard Elo is order dependence, wherein the specific sequence of matches heavily influences the final ratings. HumorRank mitigates this vulnerability by implementing \textit{Stable Elo}: the entire tournament history is evaluated across $N{=}10$ randomly shuffled topological orderings. The final assigned score is the arithmetic mean of the resulting terminal ratings, yielding strong empirical sequence robustness. Shuffle-audit details are in Appendix~\ref{sec:appendix_elo_stability}.

\subsection{Adaptive Swiss Pairing}
\label{sec:pairing}

For large model pools, exhaustive $\mathcal{O}(K^2)$ pairwise comparisons become computationally expensive. HumorRank resolves this through \textit{Adaptive Swiss Pairing} (ASP): a single scheduling engine controlled by budget parameter $C_{\max}$ that preferentially matches models of similar standing while avoiding repeat pairings. ASP uses a temporary online strength score only for matchmaking. Final leaderboard ratings are always computed with global Bradley--Terry MLE. A pair $(i,j)$ is \textit{under-sampled} when its observed duel count in the current match graph $G$ falls below the target count implied by $C_{\max}$. ASP prioritizes such pairs in each scheduling round until the budget is exhausted.


At maximum budget, $C_{\max}$ recovers exhaustive round-robin (\textbf{Full RR}). Reduced-budget modes subsample duels via the same ASP engine: \textbf{Swiss 2RR} fixes two rounds per model, while \textbf{Swiss 3RR} fixes three rounds per model. Under a Swiss schedule with $R(K)$ rounds per model, the per-prompt comparison count is approximately $\frac{K\,R(K)}{2}$: therefore, fixed-round modes ($R\in\{2,3\}$) are $\mathcal{O}(K)$, and only schedules with $R(K)=\Theta(\log K)$ yield $\mathcal{O}(K \log K)$ comparisons per prompt. We abbreviate all round-robin schedules as RR throughout. We do not claim a formal convergence proof for BT under ASP. Empirical budget trade-offs are in Section~\ref{sec:budget_ablation} and Appendix~\ref{sec:appendix_budget_ablation}; a synthetic large-$K$ check is in Appendix~\ref{sec:appendix_asp_scaling}. Algorithm~\ref{alg:swiss} is in Appendix~\ref{sec:appendix_asp_algorithm}.

\section{Experimental Setup}

To empirically validate the HumorRank methodology, we execute a large-scale evaluation on two headline-conditioned humor generation benchmarks. Our experimental design tests discriminative power across varying model architectures, access paradigms, and parameter scales. Full reproducibility details, including hyperparameters and computational budget, are provided in Appendix \ref{sec:appendix_hyperparameters}.

\subsection{Benchmarks}

We evaluate on two publicly available humor generation benchmarks:

\textbf{SemEval-2026 MWAHAHA}~\citep{semeval2026mwahaha}: The official Task~1 test set ($300$ prompts) from the SemEval-2026 MWAHAHA shared task, which targets English joke generation conditioned specifically on news headlines. This provides a baseline evaluation on a narrow, single-domain input distribution.

\textbf{Humor Transfer Bench (HTB)}~\citep{ajayi2026humorgen}: A comprehensive evaluation set of $400$ prompts designed to assess cross-domain humor generalization. To contrast with SemEval's headline-centric focus, HTB spans eight structurally distinct input domains ($50$ prompts each): Neutral Facts, Everyday Life, Abstract Concepts, Dialogic Quotations, Scenario Inputs, Analogical Prompts, Direct Instructional, and News Headlines. 

\subsection{Model Evaluation Suite}

We evaluate a deliberately diverse suite of 9 language models to assess the leaderboard's capacity to resolve fine-grained capability differences. The inclusion criteria strictly span multiple model lineages and access paradigms:
\begin{itemize}
    \item \textbf{Frontier Proprietary Models:} GPT-5~\cite{singh2025openai}, Gemini 2.5 Pro~\cite{comanici2025gemini}, Claude 3.5 Haiku~\cite{anthropic2024claude3}, and Kimi K2~\cite{team2026kimi}.
    \item \textbf{Open-Weight Models:} Llama 3.3 70B Instruct~\cite{grattafiori2024llama}, Qwen 3 32B~\cite{bai2025qwen3}, GPT OSS 120B~\cite{agarwal2025gpt}, and Qwen 2.5 7B Instruct~\citep{qwen2-5}.
    \item \textbf{Humor-Specialized Model:} \textit{HumorGen-7B}~\citep{ajayi2026humorgen}, a humor fine-tuned model trained via Cognitive Synergy Framework (CSF) and supervised fine-tuning (SFT).
\end{itemize}
This suite reflects practical compute and API budget constraints while preserving representation across frontier proprietary APIs, open-weight models, and the humor fine-tuned \textit{HumorGen-7B}.

\subsection{Evaluation Protocol and LLM-as-Judge Ablation}

HumorRank employs LLM judges for pairwise comparison of contestant models on each benchmark ($K{=}9$ in our experiments), with all duels scheduled by Adaptive Swiss Pairing (Section~\ref{sec:pairing}). At the maximum budget, ASP becomes equivalent to exhaustive round-robin evaluation (\textbf{Full RR}), which we use for the main leaderboards, while reduced-budget Swiss modes are evaluated in Section~\ref{sec:budget_ablation}. Because judge quality depends heavily on the model performing the evaluation, we evaluated both proprietary and open-weight models before selecting the final configuration. This process revealed a broader limitation: no LLM judge reliably evaluates humor by default, and different judge models exhibit distinct failure modes, including score anchoring under absolute rubrics and family bias toward their own model outputs under pairwise comparison. We therefore designed a dedicated ablation study to characterize these failure modes before finalizing the judge configuration (Appendix~\ref{sec:appendix_judge_failures}). Configurations exhibiting these failure modes were excluded, and the final judge models are described below.

\textbf{LLM Judges:} Llama 3.3 70B Instruct serves as the primary judge for reported leaderboard ratings, and Qwen 2.5 72B Instruct serves as an independent secondary judge that re-labels the same duel set for cross-judge validation. Both judges follow the GTVH-grounded pairwise evaluation protocol in Section~\ref{sec:gtvh_formulation}, rather than an unconstrained funniness assessment. For each duel, the judge returns one structured response containing brief reasoning, a winner label (A, B, or TIE), and categorical annotations drawn from three closed vocabularies aligned with GTVH: humor mechanisms, delivery features, and loser failure modes. Tag-level definitions are provided in the evaluation prompt (Appendix~\ref{sec:appendix_judge_prompts}) and Appendix Table~\ref{tab:feature_definitions}, ensuring that comparisons are grounded in interpretable comedic attributes rather than unconstrained preference signals. The resulting annotations capture both the final preference decision and the mechanisms or shortcomings supporting that decision. This evaluation template was selected based on ablation results showing that unconstrained pairwise prompts and absolute-scoring formulations led to ranking instability, excessive ties, or model-specific preference biases (Appendix~\ref{sec:appendix_judge_failures}). Prompt order is swapped across comparisons to mitigate position bias.

\textbf{Judge Ablation \& Validity Check:} We report SemEval Qwen-judge Full RR ratings in Appendix~\ref{sec:appendix_qwen_leaderboard}, along with cross-benchmark Kendall $\tau$ analysis and budget ablations on both benchmarks (SemEval-2026 MWAHAHA and Humor Transfer Bench) in Appendix~\ref{sec:appendix_budget_ablation}. Large-scale human ranking of the full tournament is impractical because humor preference is subjective, and contestant models generate multiple jokes per prompt that often share similar setups and wording. Exhaustive pairwise human comparison is therefore costly and may yield only moderate inter-annotator agreement. We conduct a blind annotation study on a 90-pair evaluation set to assess whether our LLM judges align with human preferences on closely matched humor comparisons. This study serves as a reliability check rather than a substitute for large-scale human evaluation of the full tournament (Appendix~\ref{sec:appendix_human_eval}, Section~\ref{sec:human_eval}).

\section{Results}

Our evaluation yields an extensive empirical profile of humor capability across current language models. We present the system-level Bradley-Terry (BT) leaderboard, validate its stability across independent LLM judges, and subsequently decompose these ratings into interpretable psychometric features.

\subsection{HumorRank Leaderboard}

The Full RR tournaments, judged by the primary Llama 3.3 70B judge, reveal clear stratification on both benchmarks. Table~\ref{tab:leaderboard} reports Bradley--Terry ratings, Stable Elo reference scores, 95\% confidence intervals, and win rates. HTB (14,400 judgments) appears above SemEval (10,800 judgments). SemEval win-rate heatmaps are in Appendix Figure~\ref{fig:appendix_full_rr_llama}, and HTB budget-mode tables and figures are in Appendix~\ref{sec:appendix_htb_budget_llama}.

\begin{table}[t]
\centering
\footnotesize
\setlength{\tabcolsep}{2.5pt}
\begin{tabular}{@{}rlrrrr@{}}
\toprule
\multicolumn{6}{l}{\textit{Humor Transfer Bench (HTB): Full RR, Llama 3.3 70B judge, 14{,}400 judgments}} \\
\midrule
\textbf{Rank} & \textbf{Model} & \textbf{BT Rating} & \textbf{St.\ Elo} & \textbf{95\% CI} & \textbf{Win \%} \\
\midrule
1 & GPT-5 & 1314.7 & 1300.4 & $[1301.3, 1329.0]$ & 84.1\% \\
2 & Kimi K2 & 1242.0 & 1239.1 & $[1229.5, 1255.6]$ & 77.2\% \\
3 & HumorGen-7B\footnotemark[1] & 1097.7 & 1122.8 & $[1084.4, 1109.9]$ & 60.6\% \\
4 & Claude 3.5 Haiku & 1054.2 & 1058.3 & $[1043.2, 1068.5]$ & 55.1\% \\
5 & Gemini 2.5 Pro & 1024.0 & 1009.0 & $[1010.6, 1038.2]$ & 51.3\% \\
6 & GPT OSS 120B & 1009.6 & 1017.2 & $[998.4, 1021.6]$ & 49.5\% \\
7 & Qwen 3 32B & 942.5 & 946.2 & $[928.1, 955.7]$ & 41.3\% \\
8 & Llama 3.3 70B & 791.9 & 795.4 & $[776.4, 808.0]$ & 24.9\% \\
9 & Qwen 2.5 7B Instruct & 523.5 & 511.8 & $[501.8, 549.8]$ & 5.9\% \\
\midrule
\multicolumn{6}{l}{\textit{SemEval-2026 MWAHAHA: Full RR, Llama 3.3 70B judge, 10{,}800 judgments}} \\
\midrule
\textbf{Rank} & \textbf{Model} & \textbf{BT Rating} & \textbf{St.\ Elo} & \textbf{95\% CI} & \textbf{Win \%} \\
\midrule
1 & GPT-5 & 1307.5 & 1317.6 & $[1289.9, 1325.9]$ & 84.0\% \\
2 & Kimi-K2 & 1156.9 & 1175.7 & $[1139.9, 1170.7]$ & 67.8\% \\
3 & Gemini 2.5 Pro & 1115.1 & 1115.3 & $[1099.0, 1128.1]$ & 62.6\% \\
4 & HumorGen-7B\footnotemark[1] & 1092.8 & 1102.2 & $[1078.7, 1108.5]$ & 59.8\% \\
5 & Claude 3.5 Haiku & 1037.5 & 1027.3 & $[1024.1, 1050.9]$ & 52.7\% \\
6 & GPT OSS 120B & 1015.0 & 1002.2 & $[1001.7, 1030.6]$ & 49.8\% \\
7 & Qwen 3 32B & 976.9 & 966.2 & $[964.7, 988.8]$ & 45.0\% \\
8 & Llama 3.3 70B & 761.0 & 754.0 & $[743.9, 780.6]$ & 21.8\% \\
9 & Qwen 2.5 7B Instruct & 537.4 & 539.4 & $[513.5, 563.5]$ & 6.5\% \\
\bottomrule
\end{tabular}
\caption{Full round-robin HumorRank leaderboards on HTB (top) and SemEval (bottom), both judged by the primary Llama 3.3 70B judge. BT ratings are the primary metric, and Stable Elo (10 shuffle runs) is a sequence-robust audit. Anchor ranks are stable across benchmarks (GPT-5 \#1, Llama 3.3 70B and Qwen 2.5 7B Instruct \#8/\#9). SemEval win-rate heatmap: Appendix Figure~\ref{fig:appendix_full_rr_llama}. HTB extended budget ablations: Appendix~\ref{sec:appendix_htb_budget_llama}.}
\label{tab:leaderboard}
\end{table}
\footnotetext[1]{Referred to as HumorGen SFT 7B in plots and figures.}

On HTB, \textit{HumorGen-7B} ranks 3rd (BT = 1097.7), above Gemini 2.5 Pro and GPT OSS 120B. On SemEval it ranks 4th (BT = 1092.8). The primary Llama judge ranks its own generations 8th on both benchmarks (SemEval BT = 761.0, HTB BT = 791.9), providing no clear evidence of self-favoring in the resulting rankings. On SemEval, two-sided binomial tests reject a 50\% null win rate for 32/36 pairings at $\alpha{=}0.05$, with the remaining four concentrated in close mid-tier matchups.

\subsection{Cross-LLM-Judge Validity and Rank Stability}

Evaluating subjective data is inherently sensitive to the choice of the primary LLM judge. We replicate all full round-robin duels on both benchmarks with Qwen 2.5 72B Instruct as an independent secondary LLM judge (SemEval ratings in Appendix~\ref{sec:appendix_qwen_leaderboard}, $\tau$ tables in Appendix~\ref{sec:appendix_budget_ablation}). Bradley--Terry ratings from the Qwen judge correlate strongly with the primary Llama 3.3 70B leaderboard on SemEval and HTB: \textbf{Kendall's $\tau = 0.889$} ($p = 0.0002$) in each benchmark, and the same value when pooling all 25,200 Llama--Qwen LLM-judge pairwise labels (82.9\% agreement, Krippendorff $\alpha = 0.658$). Contestant ordering is \textit{cross-judge stable}: GPT-5 and Kimi K2 remain at ranks 1--2, and Llama 3.3 70B and Qwen 2.5 7B Instruct remain at ranks 8--9, with modest mid-tier reordering only. We also report a \textbf{transitivity score}: among all model triples with a clear pairwise winner on each edge, the fraction with no directed 3-cycle in the win graph ($1.0$ = no intransitivity). This score is $1.0$ under both LLM judges on both benchmarks.

\subsection{Tournament Budget Ablation}
\label{sec:budget_ablation}

Full round-robin is expensive as $K$ grows. Because ASP is the same engine at every budget, we ablate Swiss 2RR and Swiss 3RR schedules using the same pairing logic and fixed budget constraints. Table~\ref{tab:budget_ablation} summarizes comparison budget and cross-LLM-judge Kendall $\tau$ averaged over four cells (SemEval/HTB $\times$ Llama 3.3 70B/Qwen judges). At $K{=}9$, Swiss 3RR uses $12$ pairs/prompt ($\sim$33\% of Full RR) and restores SemEval Llama$\leftrightarrow$Qwen agreement to $\tau = 0.889$, matching Full RR. Ranks \#1 (GPT-5), \#8 (Llama 3.3 70B), and \#9 (Qwen 2.5 7B Instruct) are stable across Full RR, Swiss 2RR, and Swiss 3RR in all four evaluation cells. We treat Swiss 3RR as the practical scaling mode when exhaustive coverage is infeasible. Cross-judge stability is a budget effect: both benchmarks agree at Full RR and 3RR ($\tau = 0.889$), and 2RR is an under-budget stress test where rankings become volatile. Per-benchmark breakdowns are in Appendix~\ref{sec:appendix_semeval_budget} (SemEval) and Appendix~\ref{sec:appendix_htb_budget_llama}--\ref{sec:appendix_htb_qwen_budget} (HTB); a synthetic large-$K$ ASP check is in Appendix~\ref{sec:appendix_asp_scaling}.

\begin{table}[H]
\centering
\small
\begin{tabular}{lccc}
\toprule
\textbf{Schedule} & \textbf{Pairs / prompt} ($K{=}9$) & \textbf{Budget} & \textbf{Avg.\ cross-LLM-judge $\tau$} \\
\midrule
Full RR & 36 & 100\% & 0.889 \\
Swiss 3RR & 12 & $\sim$33\% & 0.861 \\
Swiss 2RR & 8 & $\sim$22\% & 0.806 \\
\bottomrule
\end{tabular}
\caption{Adaptive Swiss Pairing budget ablation across Full RR, Swiss 3RR, and Swiss 2RR. SemEval Llama 3.3 70B$\leftrightarrow$Qwen 2.5 72B LLM-judge $\tau = 0.889$ under both Full RR and Swiss 3RR.}
\label{tab:budget_ablation}
\end{table}

\FloatBarrier
\subsection{Human and LLM Judge Agreement}
\label{sec:human_eval}

To assess reliability of our LLM-as-a-Judge pipeline, we conducted a blind annotation study with three human evaluators on a 90-pair set (75 unique headlines) of curated funny-versus-funny comparisons. Pairs were stratified by comparison type (Table~\ref{tab:human_eval_design}): cross-tier, within-tier, scale, rank-spanning frontier, and alignment contrasts, rather than sampled exhaustively from the full tournament, limiting annotator fatigue from repeated setups.

\begin{table}[H]
\centering
\footnotesize
\setlength{\tabcolsep}{6pt}
\begin{tabular}{@{}l r@{}}
\toprule
\textbf{Comparison axis} & \textbf{Pairs} \\
\midrule
Cross-tier (rank span) & 32 \\
Within-tier, top quartile & 10 \\
Within-tier, lower ranks & 18 \\
Scale contrast & 10 \\
Rank-spanning frontier matchups & 15 \\
Alignment contrast & 5 \\
\midrule
\textbf{Total} & \textbf{90} \\
\bottomrule
\end{tabular}
\caption{Human evaluation set design (90 pairs, 75 unique headlines). Axes mirror tournament stratification, and all pairs are funny-versus-funny only.}
\label{tab:human_eval_design}
\end{table}

Annotators re-rated anonymized pairs against our production Llama 3.3 70B and Qwen 2.5 72B Instruct judges. Because humor preference is inherently subjective and has no single ground-truth label, we quantify reliability as agreement beyond chance using Krippendorff's $\alpha$~\citep{krippendorff2011computing} with nominal winner-model labels, supporting multi-rater cohorts with incomplete overlap. Table~\ref{tab:alpha_breakdown} reports cohort-level $\alpha$ across human-only, human--LLM, and LLM--LLM rater pools. Human--Llama alignment on H$_2$+H$_3$ is not significantly different from human--human agreement (Fisher exact $p = 0.808$). Protocol details are in Appendix~\ref{sec:appendix_human_eval}.

\begin{table}[H]
\centering
\footnotesize
\setlength{\tabcolsep}{4pt}
\renewcommand{\arraystretch}{0.92}
\begin{tabular}{@{}l r@{}}
\toprule
\textbf{Cohort} & \textbf{Krippendorff's $\alpha$} \\
\midrule
\multicolumn{2}{@{}l}{\textit{Human-only dyads}} \\
H$_{1}$ + H$_{3}$ & 0.446 \\
H$_{2}$ + H$_{3}$ & 0.436 \\
H$_{1}$ + H$_{2}$ & 0.334 \\
H$_{1}$ + H$_{2}$ + H$_{3}$ & 0.416 \\
\midrule
\multicolumn{2}{@{}l}{\textit{Human--Llama judge dyads}} \\
Llama + H$_{1}$ & 0.434 \\
Llama + H$_{2}$ & 0.441 \\
Llama + H$_{3}$ & 0.458 \\
Llama + H$_{2}$ + H$_{3}$ & 0.446 \\
Llama + H$_{1}$ + H$_{2}$ + H$_{3}$ & 0.432 \\
\midrule
\multicolumn{2}{@{}l}{\textit{Human--Qwen judge dyads}} \\
Qwen + H$_{2}$ + H$_{3}$ & 0.421 \\
Qwen + H$_{1}$ + H$_{2}$ + H$_{3}$ & 0.407 \\
\midrule
\multicolumn{2}{@{}l}{\textit{LLM--LLM judge dyad}} \\
Llama + Qwen & 0.505 \\
\bottomrule
\end{tabular}
\caption{Krippendorff's $\alpha$ (nominal winner-model labels) on the 90-pair blind evaluation set. H$_{i}$: human annotator $i$, \textit{Llama}/\textit{Qwen}: production LLM judges. Human--Llama alignment on H$_2$+H$_3$ is not significantly different from human--human agreement (Fisher exact $p = 0.808$).}
\label{tab:alpha_breakdown}
\end{table}

\FloatBarrier
\subsection{Theory-Grounded Feature Interpretability}

Beyond scalar Elo ratings, HumorRank's structured judge co-emits GTVH-grounded tags~\citep{attardo2017general} with each pairwise decision: \textit{humor mechanisms}, \textit{delivery features}, and \textit{failure modes}. Table~\ref{tab:gtvh_features} summarizes the primary tags in our judge prompt.

\begin{table}[H]
\centering
\small
\setlength{\tabcolsep}{4pt}
\renewcommand{\arraystretch}{0.95}
\begin{tabular}{@{}lcp{0.52\linewidth}@{}}
\toprule
\textbf{Feature} & \textbf{GTVH} & \textbf{Description} \\
\midrule
\textit{Incongruity} & LM & Conflicting scripts or ideas. \\
\textit{Absurdity} & SI & Breaks physical/social expectations. \\
\textit{Sarcasm} & TA/LM & Irony with an explicit target. \\
\textit{Wordplay} & LA & Puns and lexical/syntactic ambiguity. \\
\textit{Conciseness} & LA & Efficient buildup, comedic timing. \\
\bottomrule
\end{tabular}
\caption{GTVH-grounded humor features tagged in structured LLM judge responses.}
\label{tab:gtvh_features}
\end{table}

\begin{figure}[t]
    \centering
    \includegraphics[width=0.7\columnwidth]{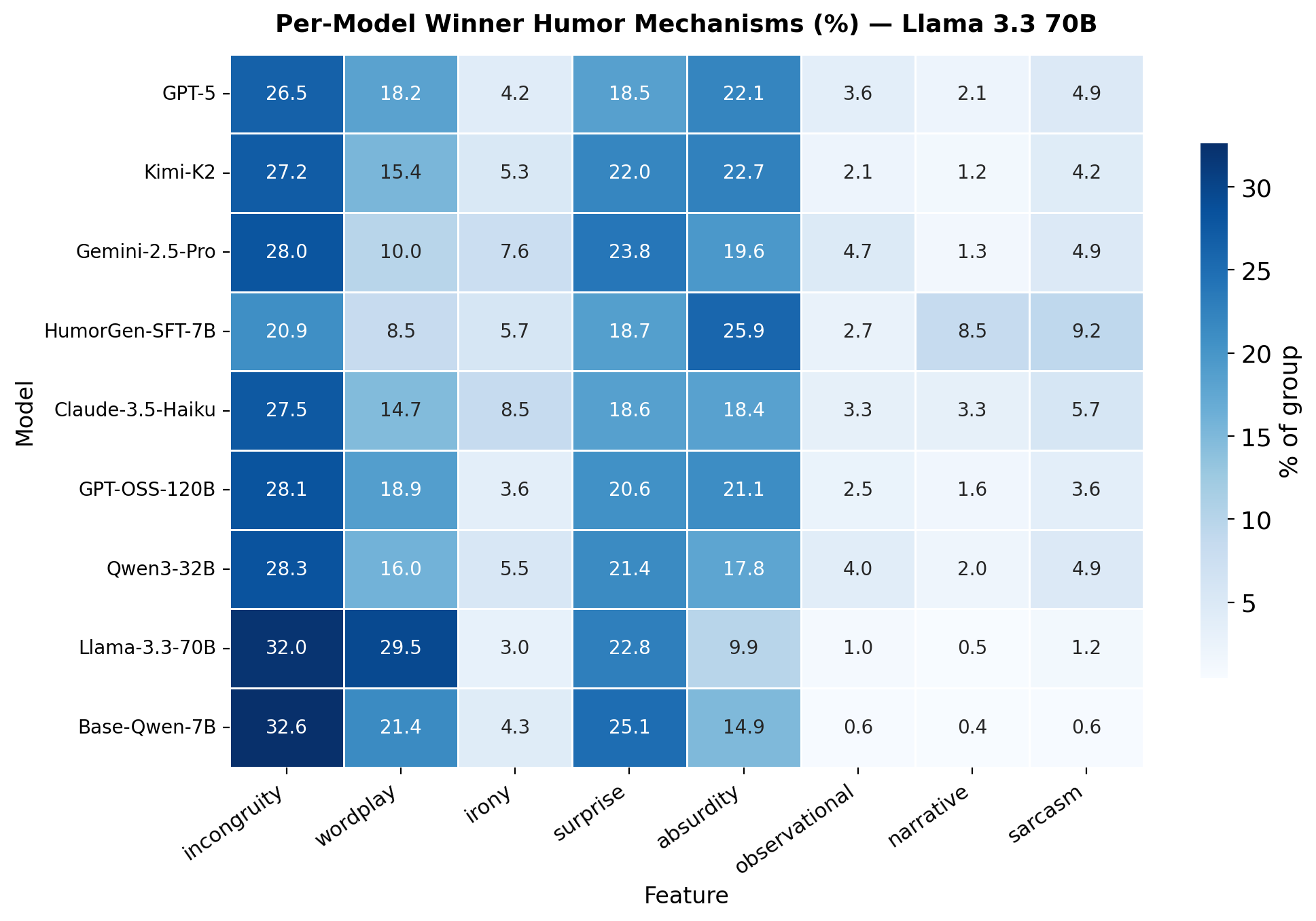}
    \caption{Per-model winning humor-mechanism distributions under the Llama judge (\% of wins). Frontier models emphasize \textit{Incongruity}, mid-tier specialists lead on \textit{Absurdity}, and baselines over-index on \textit{Wordplay}.}
    \label{fig:heatmap_llama_humor}
\end{figure}

\begin{figure}[t]
    \centering
    \includegraphics[width=0.7\columnwidth]{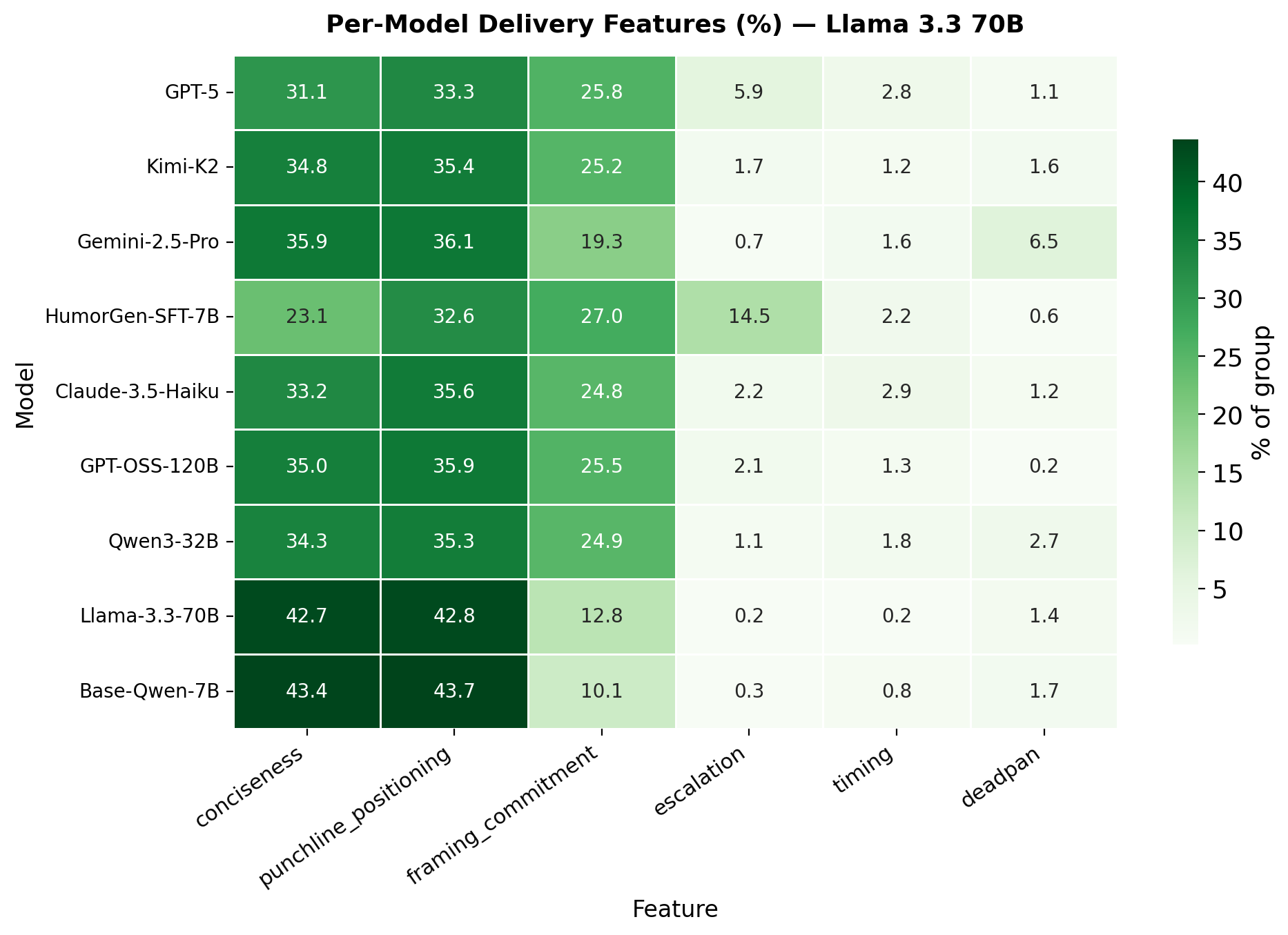}
    \caption{Per-model winning delivery-feature distributions under the Llama judge (\% of wins). Frontier models emphasize \textit{Conciseness}, while mid-tier specialists show elevated \textit{Escalation}.}
    \label{fig:heatmap_llama_delivery}
\end{figure}

\begin{figure}[H]
    \centering
    \includegraphics[width=0.7\columnwidth]{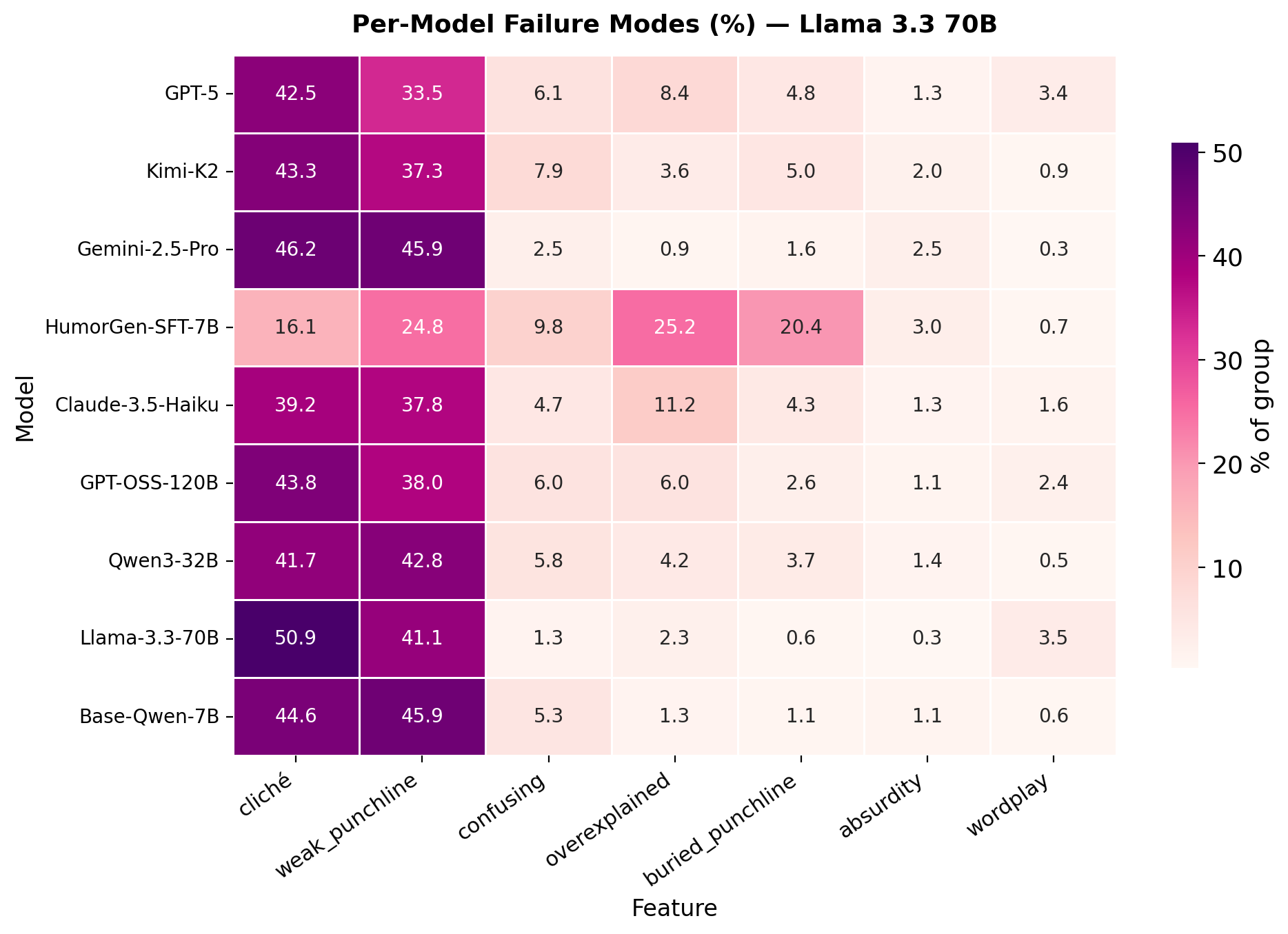}
    \caption{Per-model failure mode distributions (Llama judge). \textit{Clich\'e} and \textit{Weak Punchline} dominate most tiers. Mid-tier specialists show higher \textit{Overexplained} (25.2\%) and \textit{Buried Punchline} (20.4\%) rates.}
    \label{fig:heatmap_llama_loser}
\end{figure}

Figures~\ref{fig:heatmap_llama_humor}--\ref{fig:heatmap_llama_loser} visualize per-model tag frequencies from the primary Llama judge across the nine-model leaderboard.

These distributions reveal three recurring tiers:
\begin{itemize}\setlength{\itemsep}{1pt}\setlength{\parsep}{0pt}
    \item \textbf{Frontier generalists (e.g., GPT-5):} High \textit{Conciseness} (31.1\% of wins) and \textit{Incongruity} (26.5\%), with primary losses via \textit{Clich\'e}.
    \item \textbf{Mid-tier specialists (e.g., HumorGen-7B):} Lead on \textit{Absurdity} (25.9\%) and \textit{Sarcasm} (9.2\%), with elevated \textit{Escalation} (14.5\%).
    \item \textbf{Weak baselines (e.g., Llama 3.3 70B):} Over-index on \textit{Wordplay} (29.5\%) and accumulate \textit{Weak Punchline} (41.1\%) and \textit{Clich\'e} (50.9\%) when losing.
\end{itemize}

The same tier structure holds under the Qwen judge (Appendix~\ref{sec:appendix_qualitative}). Representative judge rationales are in Appendix~\ref{sec:appendix_llama_judge}.

\section{Conclusion}

We introduced HumorRank, a leaderboard-oriented framework for theory-grounded comparison of humor-generation systems. Rather than reducing humor to a universal scalar score, HumorRank organizes pairwise judgments into system-level rankings and interpretable model profiles. Budget-aware Adaptive Swiss Pairing reduces comparisons, while global Bradley--Terry estimation produces stable ratings; comparative rationales and mechanism, delivery, and failure annotations retain information omitted by scalar leaderboards. Across two benchmarks, stability under independent judges and reduced budgets, together with human calibration, indicates a consistent comparative signal despite humor's subjectivity. Results further suggest that humor-generation capability reflects comedic specialization and mechanism mastery, not scale alone. HumorRank thus provides a reusable basis for comparing systems, diagnosing successes and failures, and tracking progress in computational humor generation. HumorRank is maintained at \url{https://humorrank-leaderboard.pages.dev/}.

\section{Limitations}
\label{sec:limitations}

The limitations of this study fall into three categories:

\begin{itemize}
    \item \textbf{Evaluation Scope:} Evaluation is restricted to English data and nine models. Consequently, the study does not examine cross-lingual or cross-cultural humor, and the evaluated systems do not span the full range of contemporary models. Neither HTB nor SemEval covers interactive or multimodal humor. Real humor-tournament ASP fidelity is validated at $K{=}9$; larger pools are stress-tested only under a synthetic Bradley--Terry preference model (Appendix~\ref{sec:appendix_asp_scaling}), not with jokes from $10^{4}$--$10^{5}$ generators.

    \item \textbf{Human Validation Scale:} Exhaustive annotation of all 25,200 tournament duels is impractical because of cost and annotator fatigue. We therefore conduct a targeted 90-pair blind evaluation of difficult funny-vs-funny comparisons (Table~\ref{tab:human_eval_design}, Appendix~\ref{sec:appendix_human_eval}) as a post-hoc calibration of judge behavior. Human--Llama alignment does not differ significantly from human--human agreement on H$_2$+H$_3$ (Fisher exact $p = 0.808$). This provides practical validation of judge reliability without exhaustive annotation of the full tournament.

    \item \textbf{Broader Judge Evaluation:} Our judge-selection process screens for self- and family-preference biases and validates selected judges through cross-judge stability and human alignment (Appendix~\ref{sec:appendix_judge_failures}). The study focuses on these dimensions. Future work can extend evaluation to cross-cultural and stylistic settings and independently assess GTVH feature annotations with human experts.
\end{itemize}
\section{Reproducibility Statement}
To ensure full reproducibility of the HumorRank framework, we detail the core hyperparameter configurations and computational hardware requirements necessary to execute the generative tournament.

\textbf{Generation Hyperparameters:} For all candidate models evaluated in the tournament, we standardized the generation settings to prioritize creative diversity while maintaining structural coherence. Specifically, we configured all candidate models with a unified sampling temperature of $T=0.7$ and nucleus sampling of top-$p=0.9$ (set explicitly wherever the provider API exposes it). Token limits were inherited from respective model APIs to preserve native instructional adherence without imposing artificial truncation.

\textbf{Judge Hyperparameters:} The LLM judges (Llama 3.3 70B and Qwen 2.5 72B) were configured with a highly constrained sampling temperature of $T=0.1$ alongside a maximum retry threshold of 3 (with exponential backoff) for all pairwise JSON evaluation calls. Given the substantial financial and computational cost of the expansive generative tournament, this near-deterministic setting ensures that the LLM judges maintain stability and does not yield erratic or contradictory evaluations to the same prompt upon reassessment, thereby firmly preserving the integrity of the Bradley-Terry ratings.

\textbf{Computational Hardware:} Orchestrating full round-robin judging on SemEval and HTB (25,200 pairwise calls per LLM judge) required approximately 48 hours of dedicated NVIDIA H100 (80GB) GPU compute for the primary Llama judge pipeline. Qwen-judge replication required additional inference budget. Tournament code, evaluation scripts, and the Adaptive Swiss pairing implementation are provided in the supplementary materials.

\section{Ethics Statement}

This work proposes a framework for the systematic evaluation and ranking of humor generation across large language models. It does not itself constitute a humor generation system. Two ethical considerations warrant explicit acknowledgment. First, humor is a culturally and contextually variable phenomenon whose boundaries with offensive or exclusionary expression are highly sensitive to audience and setting. Evaluation frameworks that rank models on comedic output implicitly surface content generated by those models, and practitioners adapting such pipelines for downstream applications bear responsibility for enforcing appropriate content-moderation constraints. Second, the validity of automated ranking is constrained by the cultural and stylistic distribution of the LLM judge model's pretraining corpus. LLM-based evaluators trained predominantly on high-resource, Western-centric text may systematically disadvantage humor conventions from linguistically or culturally underrepresented communities, a limitation shared by the broader LLM-as-a-judge literature. HumorRank should therefore be interpreted as a reproducible diagnostic benchmark rather than a definitive assessment of comedic or creative quality.



\newpage
\bibliography{colm2026_conference}
\bibliographystyle{colm2026_conference}

\newpage
\appendix
\setcounter{topnumber}{3}
\setcounter{bottomnumber}{2}
\setcounter{totalnumber}{5}
\renewcommand{\topfraction}{0.90}
\renewcommand{\bottomfraction}{0.80}
\renewcommand{\textfraction}{0.07}
\renewcommand{\floatpagefraction}{0.80}
\setlength{\textfloatsep}{8pt plus 2pt minus 2pt}
\setlength{\floatsep}{6pt plus 2pt minus 2pt}
\setlength{\intextsep}{8pt plus 2pt minus 2pt}
\floatplacement{table}{!tbp}
\floatplacement{figure}{!tbp}
\newpage
\section*{Appendix: Table of Contents}
\addcontentsline{toc}{section}{Contents of the Appendix}

\newcommand{\apptoc}[2]{\noindent\textbf{#1} \dotfill \pageref{#2}\\[0.35ex]}
\newcommand{\apptocsub}[2]{\noindent\hspace{1.5em}#1 \dotfill \pageref{#2}\\[0.35ex]}

\apptoc{\ref{sec:appendix_humor_definition} Humor Definition}{sec:appendix_humor_definition}
\apptoc{\ref{sec:appendix_judge_failures} Failure Modes of LLM-as-a-Judge on Humor Evaluation}{sec:appendix_judge_failures}
\apptocsub{\ref{sec:appendix_judge_failures_program} Experimental program}{sec:appendix_judge_failures_program}
\apptocsub{\ref{sec:appendix_failure_mode_a} Failure mode A: score anchoring (Experiments 1--3)}{sec:appendix_failure_mode_a}
\apptocsub{\ref{sec:appendix_failure_mode_b} Failure mode B: judge-model self-preference (5-headline pilot)}{sec:appendix_failure_mode_b}
\apptocsub{\ref{sec:appendix_failure_mode_c} Failure mode C: spurious ties (early pairwise prompt)}{sec:appendix_failure_mode_c}
\apptocsub{\ref{sec:appendix_failure_pairs_qual} Illustrative failure pairs (qualitative audit)}{sec:appendix_failure_pairs_qual}
\apptocsub{\ref{sec:appendix_production_config} Production configuration (what survived)}{sec:appendix_production_config}
\apptoc{\ref{sec:appendix_judge_prompts} LLM-as-a-Judge Prompting Framework}{sec:appendix_judge_prompts}
\apptocsub{\ref{sec:feature_taxonomy} Feature Taxonomy Definitions}{sec:feature_taxonomy}
\apptoc{\ref{sec:appendix_budget_ablation} Tournament Budget Ablation (Full RR / Swiss 2RR / Swiss 3RR)}{sec:appendix_budget_ablation}
\apptocsub{\ref{sec:appendix_budget_tau} Cross-judge $\tau$ by budget mode}{sec:appendix_budget_tau}
\apptocsub{\ref{sec:appendix_swiss3rr_semeval_ranks} Swiss 3RR rank stability (SemEval, Llama judge)}{sec:appendix_swiss3rr_semeval_ranks}
\apptocsub{\ref{sec:appendix_swiss3rr_htb_ranks} Swiss 3RR rank stability (HTB, Llama judge)}{sec:appendix_swiss3rr_htb_ranks}
\apptocsub{\ref{sec:appendix_semeval_budget} SemEval budget ablation (Llama judge)}{sec:appendix_semeval_budget}
\apptocsub{\ref{sec:appendix_htb_budget_llama} HTB budget ablation (Llama judge)}{sec:appendix_htb_budget_llama}
\apptocsub{\ref{sec:appendix_htb_qwen_budget} HTB budget ablation (Qwen judge)}{sec:appendix_htb_qwen_budget}
\apptoc{\ref{sec:appendix_asp_algorithm} Adaptive Swiss Pairing Algorithm}{sec:appendix_asp_algorithm}
\apptoc{\ref{sec:appendix_asp_scaling} Synthetic ASP Scaling Stress Test ($K$ up to $10^{5}$)}{sec:appendix_asp_scaling}
\apptoc{\ref{sec:appendix_qwen_leaderboard} HumorRank Leaderboard Performance with Qwen 2.5 72B LLM Judge}{sec:appendix_qwen_leaderboard}
\apptoc{\ref{sec:appendix_llama_judge} Llama Judge Sample Decisions}{sec:appendix_llama_judge}
\apptoc{\ref{sec:appendix_hyperparameters} Hyperparameter Configurations}{sec:appendix_hyperparameters}
\apptoc{\ref{sec:appendix_qualitative} Qualitative Examples and Feature Reasoning}{sec:appendix_qualitative}
\apptocsub{\ref{sec:appendix_qualitative_observations} Key Observations (Qwen vs.\ Llama LLM judges)}{sec:appendix_qualitative_observations}
\apptoc{\ref{sec:appendix_human_eval} Human Evaluation Details}{sec:appendix_human_eval}
\apptocsub{\ref{sec:appendix_human_eval_selection} Pair selection and $\alpha$ computation}{sec:appendix_human_eval_selection}
\apptocsub{\ref{sec:appendix_human_eval_instructions} Instructions to participants}{sec:appendix_human_eval_instructions}
\apptocsub{\ref{sec:appendix_human_eval_participants} Participants}{sec:appendix_human_eval_participants}
\apptocsub{\ref{sec:appendix_human_eval_reliability} Inter-Annotator Reliability}{sec:appendix_human_eval_reliability}
\apptoc{\ref{sec:appendix_elo_stability} Stable Elo Shuffle Audit}{sec:appendix_elo_stability}
\vspace{1.5em}

\newpage

\section{Humor Definition}
\label{sec:appendix_humor_definition}

Building upon classic Incongruity Theory, psychological frameworks~\citep{larkin2017overview}, and Normative-Violation theory~\citep{veatch1998theory}, we require a rigorous working definition that can be applied to text-based evaluation. For the purposes of this research, we define humor explicitly as:

\begin{quote}
Humor is the cognitive reward (experienced as amusement) arising when an interlocutor successfully resolves a deliberately constructed incongruity, such as the narrative shift between a joke's setup and punchline, within a harmless and non-threatening context.
\end{quote}

This definition anchors psychological consensus into the practical reality of evaluating generated text. It explicitly requires four distinct components:

\begin{itemize}
    \item \textbf{The Joke Mechanism (Setup \& Punchline):} We evaluate humor not as random surprise, but as a structured linguistic narrative. The setup creates a logical expectation, and the punchline deliberately subverts it.
    \item \textbf{The ``Cognitive Reward'':} This maps to the cognitive appraisal process, describing the computational or intellectual achievement of bridging the logical gap between the setup and punchline.
    \item \textbf{Experienced as Amusement:} The cognitive resolution must trigger a pleasant response (mirth) rather than confusion.
    \item \textbf{Harmless Context:} Drawn from benign violation theory, the structural incongruity only produces amusement if it is appraised as non-threatening.
\end{itemize}

Because humor exists on a continuous spectrum determined by these mechanisms rather than as a discrete label, our methodology utilizes \textit{pairwise preference ranking}. By prompting the LLM judge to evaluate which generation produces a stronger cognitive reward, we effectively treat humor evaluation as a reward modeling paradigm across the multi-dimensional feature space of human amusement.

\section{Failure Modes of LLM-as-a-Judge on Humor Evaluation}
\label{sec:appendix_judge_failures}

HumorRank's evaluation protocol was not chosen a priori: it was the survivor of a deliberate search through LLM-as-a-Judge configurations that \textit{fail} on humor generation ranking. We document two distinct failure classes: \textbf{(A) paradigm failure}, where numeric or absolute scoring collapses despite careful rubrics; and \textbf{(B) judge-model failure}, where pairwise structure is correct but the judge model exhibits self-preference or family bias. Section~\ref{sec:why_pairwise} in the main paper summarizes the conclusion; this appendix holds the full experimental program and evidence.

\subsection{Experimental program}\label{sec:appendix_judge_failures_program}

We ran four tracks before locking the production tournament (Llama + Qwen72, structured pairwise judge prompt, 10,800 SemEval duels):

\begin{itemize}
    \item \textbf{Track A: numeric judges (Exp.\ 1--4).} Absolute 0--100, structured $4{\times}1$--$5$ $\to$ 0--100, and scalar 1--20 paradigms over 1,000--1,200 headlines with $\sim$11--15 humor candidates each ($\sim$41k scores total). Exp.\ 4 was a 120-headline Gemini absolute pilot (4 models/headline).
    \item \textbf{Track B: pairwise judge screening (9 SemEval models).} Same-prompt pairwise duels on the paper's nine contestants using GPT-5, Gemini 2.5 Pro, GPT-OSS 120B, and Qwen3-32B as judges. GPT-5 and Gemini were evaluated on \textbf{five shared headlines} (\texttt{en\_2001}--\texttt{en\_2005}; 36 pairs/headline $\Rightarrow$ 180 duels each); OSS and Qwen32 runs used ten headlines for additional coverage.
    \item \textbf{Track C: early pairwise prompt.} Verbose chain-of-thought pairwise pilot preceding the final structured prompt.
    \item \textbf{Production.} Screened open-weight judges (Llama 3.3 70B, Qwen 2.5 72B), structured pairwise prompt, position swap, full SemEval round-robin ($\tau = 0.889$).
\end{itemize}

Table~\ref{tab:judge_failures} summarizes outcomes.

\begin{table}[H]
\centering
\small
\begin{tabular}{p{0.22\linewidth} p{0.28\linewidth} p{0.32\linewidth} p{0.08\linewidth}}
\toprule
\textbf{Paradigm} & \textbf{Symptom} & \textbf{Response} & \textbf{n} \\
\midrule
Absolute 0--100 rubric & High-band clustering; mean within-HL spread 20.6 & Abandoned & 17k \\
Structured $4{\times}1$--$5$ $\to$ 0--100 & 88.5\% scores $=$ 70.0; 30.8\% headlines all tie & Abandoned & 12k \\
Scalar 1--20 funniness & 82.7\% in band 13--16; 22.2\% spread $\le 1$ & Abandoned & 12k \\
Early pairwise prompt (verbose CoT) & $\sim$62\% tie rate in pilot & Revised prompt & pilot \\
GPT/Gemini judge (pairwise) & 75--88\% self-win on shared pool & Rejected & 180--360 \\
\textbf{Structured pairwise + Llama/Qwen72} & Cross-judge $\tau = 0.889$ & \textbf{Production} & 10,800 \\
\bottomrule
\end{tabular}
\caption{LLM-as-a-Judge configurations tested on humor evaluation. Failed rows document \textit{why} HumorRank does not use absolute scoring or proprietary self-preferring judges.}
\label{tab:judge_failures}
\end{table}

\subsection{Failure mode A: score anchoring (Experiments 1--3)}\label{sec:appendix_failure_mode_a}

\textbf{Experiment 1: absolute 0--100 rubric.} An expert-style 0--100 funniness rubric with explicit bands (e.g., 85--100 ``excellent'', 70--84 ``good'') was applied by \textbf{Llama 3.3 70B Instruct} to $\sim$11--15 joke candidates per headline across 1,200 headlines (17,317 individual scores). Scores clustered in a narrow high band despite diverse candidates; mean within-headline spread was only 20.6 points on a 0--100 scale, and rubric bands compressed most outputs into ``good'' rather than separating models. This is a \textit{paradigm failure}: absolute scoring did not reliably discriminate humor quality among same-headline candidates, so it is unsuitable for leaderboard evaluation.

\textbf{Illustrative example (\texttt{en\_0004}).} Table~\ref{tab:exp1_en0004} provides one full headline-level sample. The same prompt receives only six score anchors (87/83/82/77/74/62), with verbose 161--163-word setups placed at the top band while tighter ``weigh in''/scale punchlines are pushed to 62. The issue is not isolated ranking noise; the evaluation interface itself fails to separate which jokes land.

\begingroup
\footnotesize
\setlength{\LTleft}{0pt}
\setlength{\LTright}{0pt}
\setlength{\tabcolsep}{6pt}
\renewcommand{\arraystretch}{1.18}
\begin{longtable}{@{}>{\bfseries}r>{\centering\arraybackslash}p{0.055\linewidth}>{\centering\arraybackslash}r>{\raggedright\arraybackslash}p{0.905\linewidth}@{}}
\caption{\textbf{Exp.~1 Failure Case (\texttt{en\_0004}).} Fifteen same-headline candidates scored by Llama 3.3 70B Instruct. \textbf{Metrics:} \textbf{Score} = Llama-assigned 0--100 rating; \textbf{Words} = word count. \textbf{Row Colors (score anchors):} gold/peach 87--83, lilac 82, blue 77--74, mint 62. \textbf{Error Markers:} coral (\textsuperscript{\textit{o}}) = verbose build rated too high; emerald (\textsuperscript{\textit{u}}) = tight punchline rated too low. \textbf{Takeaway:} absolute scoring collapses distinct jokes onto six anchors and misjudges which responses land.}
\label{tab:exp1_en0004} \\
\toprule
\textbf{Rk} & \textbf{Score} & \textbf{Words} & \textbf{Joke (full text)} \\
\midrule
\endfirsthead
\multicolumn{4}{l}{\footnotesize\textit{Table~\ref{tab:exp1_en0004} --- continued from previous page}} \\[2pt]
\toprule
\textbf{Rk} & \textbf{Score} & \textbf{Words} & \textbf{Joke (full text)} \\
\midrule
\endhead
\midrule
\multicolumn{4}{r}{\footnotesize continued on next page\ldots} \\
\endfoot
\bottomrule
\endlastfoot
\rowcolor{Anchor87} 1 & \cellcolor{Anchor87}{87} & 44 &
You know, ``Experts weigh in'' on body wipes. I guess that means they literally stepped on them with their lab coats on to see if they'd break. Good thing the wipes held up, otherwise we'd have a whole bunch of scientists doing the splits. \\
\rowcolor{AnchorOver} 2\textsuperscript{\textit{o}} & \cellcolor{AnchorOver}{87} & 161 &
You know, body wipes... because sometimes you need a quick fix for that lingering funk, like when you've been sitting in traffic for three hours and your deodorant decided to take a vacation. And let me tell you, experts are super excited about this groundbreaking issue. They're probably the same `experts' who got their degree from the University of ``I Can't Shower Because I'm Too Important To Be Clean.'' Listen, if these wipes really worked as well as advertised, we'd all just stop using showers altogether and the water crisis would be solved. We could rename them ``body band-aids''---they don't cure the smell, they just give you a little time before everyone starts backing away slowly. But hey, if you're in a rush and can't decide between a wipe and a shower, here's a tip: If you think a wipe will cut it, you're already too late. The cat's out of the bag and it's not the furry kind. \\
\rowcolor{AnchorOver} 3\textsuperscript{\textit{o}} & \cellcolor{AnchorOver}{87} & 163 &
So, do body wipes actually work? Let's ask the experts! I heard they can tell you not only if you smell good enough to eat (which is totally a thing now, apparently), but also if you weigh more than your cat (and who wouldn't want to know that)? Just imagine it---an expert body wipe panel sitting around discussing your cleanliness level with cups of tea and judging your body odor on a scale from `newly mown hay' to `forgotten gym socks'. And if you fail their sniff test, fear not, because there's always the option of becoming a human body wipe for the rest of the team---think of all the scrubbing you could do! The best part is, even if body wipes don't work, at least we know the experts will be busy and not questioning life's bigger mysteries like why the remote never works when you need it most. Now that's a silver lining I can get behind! \\
\rowcolor{Anchor83} 4 & \cellcolor{Anchor83}{83} & 95 &
Have you ever noticed how we have experts for everything now? Like, ``Experts weigh in on whether body wipes work.'' I mean, hello! If you're wiping yourself with a wet cloth, odds are you're not getting dirtier. It's like having a panel of experts to discuss whether a hair dryer makes your hair dry. You know what? I'm calling it: if you're using a body wipe, you are already an expert on body wipes. You don't need some other guy telling you if it works or not. You're the boss of your own smell situation! \\
\rowcolor{AnchorOver} 5\textsuperscript{\textit{o}} & \cellcolor{AnchorOver}{83} & 154 &
So I tried one of these wipes the other day, you know, just to test out the efficacy. I mean, I'm pretty sure it's just a ploy to make me buy more deodorant. But what if the real secret is in the corners, like the little nooks where your bra straps dig in or behind your knees? I mean, who even thinks about those places? And now, every time I use a body wipe, I'm like, `Did it get the corner under my left armpit?' And then I find myself spending the next 30 minutes just staring at that spot in the mirror, turning my arm around and around, going, `Did I miss it? Did I really miss it?!' And I swear, if anyone catches me doing this, they'll think I'm totally insane. But hey, I'd rather be the guy with the obsessive-compulsive corner-wiping routine than the guy who smells like old gym socks! \\
\rowcolor{AnchorUnder} 6\textsuperscript{\textit{u}} & \cellcolor{AnchorUnder}{82} & 50 &
You know, they say ``experts weigh in'' on whether body wipes actually work, but I'm pretty sure these so-called experts were using the heavy-duty kind---the ones that add pounds just from the packaging. At least now we know why the debate on body wipes has gotten so... weighty! \\
\rowcolor{Anchor82} 7 & \cellcolor{Anchor82}{82} & 111 &
So, experts have finally weighed in on body wipes, and guess what folks? They kinda work, if your definition of ``work'' is smearing last week's mystery stains into an even layer across your skin. You know you're doing it wrong when you catch yourself using them after a particularly sweaty spin class, thinking, `This will do. I mean, who needs water and soap when you've got the magic of questionable chemical concoctions?' But hey, at least we all feel a bit less guilty about not showering, right? Until the moment you accidentally sit down on something sticky and realize your body wipe game has failed you once again. Happy stinking, everybody! \\
\rowcolor{AnchorOver} 8\textsuperscript{\textit{o}} & \cellcolor{AnchorOver}{82} & 118 &
Experts have finally weighed in on body wipes, and guess what? They're not just effective at cleaning; they might just be the answer to world hunger. You see, after using these wipes, people suddenly found themselves feeling lighter---about 20 pounds lighter! It turns out, the secret ingredient isn't just magic cleaning solution, but a newfangled appetite suppressant. Who knew a swipe could be a diet plan? So next time you grab a body wipe, remember: it's not just cleaning you up, it's cutting down your weight. Just don't try to eat them---unless you want to add a new layer of flavor to your meal that includes a hint of... let's call it `sanitizing freshness.' \\
\rowcolor{Anchor77} 9 & \cellcolor{Anchor77}{77} & 97 &
You know, I used one of those body wipes the other day, and it was so effective it almost started folding my clothes and putting them away. But honestly, the only thing it missed was cleaning under my nails. Which made me wonder, if it got between every crevice of my skin, why didn't it pick up that one olive pit I've had lodged there since summer? It's like it read the manual wrong and thought its job was to prepare me for a trip to the dry cleaners instead of just giving me a quick clean. \\
\rowcolor{Anchor77} 10 & \cellcolor{Anchor77}{77} & 119 &
You know, I was at the gym the other day, and there was this guy who had this whole routine with his body wipes. He'd wipe down the machine before using it, then after every 10 minutes of exercise he stopped to wipe himself off. I asked him how it worked out for him, and he said, ``Honestly, I just feel like I'm being used by a giant baby wipe.'' I mean, if experts can't agree on whether they even work, I think the only thing these wipes guarantee is that you'll leave a trail of mystery scents wherever you go. Like, ``Hey, what's that smell? Oh, it's probably Jerry from accounting and his endless supply of body wipes!'' \\
\rowcolor{Anchor77} 11 & \cellcolor{Anchor77}{77} & 73 &
You know, after using those fancy new body wipes, I've been feeling lighter on my feet. But don't worry, it's not that kind of weight loss---though if they really did work that way, I'd never run out of excuses to wipe down at the grocery store! After all, `I'm just trying these new body wipes; they said they might help me slim down!' Who knew cleanliness could be so misleadingly slimming? \\
\rowcolor{Anchor74} 12 & \cellcolor{Anchor74}{74} & 75 &
You know, I tried using a body wipe once. Didn't work too well. So I asked my neighbor, ``Hey, could I use your body as a wipe instead?'' He looked at me funny, but then he said, ``Sure, just don't leave any lint.'' Turns out, experts say body wipes aren't very effective anyway, but they didn't mention the potential for inter-body cleaning alternatives. Who knew we had a whole neighborhood of walking, talking cleaning supplies? \\
\rowcolor{Anchor74} 13 & \cellcolor{Anchor74}{74} & 143 &
So, I was reading this article about how body wipes are great for, you know, cleaning yourself when you don't have access to a shower. But I had to ask myself, do they really get everything? Like, do they work in all the places where you can barely reach with a washcloth? I started wondering, if I use one of these things, could I be leaving microscopic bits of wipe in areas that I'd rather keep as pristine as possible? I mean, is there an emergency service for extracting rogue wipe fibers from awkward spots? Because, based on my calculations, if you use a body wipe just once a day, by the end of the month, statistically speaking, you might have a small colony of these fibers living comfortably in your underbelly. It's like inviting lint to come live rent-free in your navel! \\
\rowcolor{AnchorUnder} 14\textsuperscript{\textit{u}} & \cellcolor{AnchorUnder}{62} & 90 &
So, I asked these experts, ``Hey, when you `weigh in' on body wipes, do you stand on a scale with one in your hand?'' And you know what? They didn't laugh. They just said, ``Well, we did consider the weight of the product, but not in the way you might think.'' Because let's face it, the last thing you want after a long day is to feel weighed down by a shower, right? Just grab a wipe, and voil\`{a}---no weight gain, no water waste, and no need for a scale! \\
\rowcolor{AnchorUnder} 15\textsuperscript{\textit{u}} & \cellcolor{AnchorUnder}{62} & 113 &
So, I was reading this article on body wipes and experts weighing in, and I thought to myself, ``Experts, huh? Because obviously, the best person to consult when you've run out of shower gel is a... body wipe expert!'' And you know what? If these body wipes don't get your back as clean as a whistle, at least you can say you had a nice conversation with a cloth. After all, you never know when a friendly wipe might become your new best friend. They won't judge you for the gunk you picked up at the gym, but hey, they might introduce you to their lintly cousin who could use some love too! \\
\end{longtable}
\endgroup

\textbf{Experiment 2: structured $4 \times 1$--$5$ dimensions.} Hypothesis: decomposing humor into incongruity, resolution, linguistic, and punchline dimensions would force discrimination. Over 1,000 headlines ($\sim$12k scores), the judge repeated nearly identical dimension patterns (e.g., 4/4/3/4) on most jokes, producing a weighted total of \textbf{70.0 on 88.5\% of all scores}. On 30.8\% of headlines, \textit{every} candidate received the identical score, a complete ranking failure.

\textbf{Experiment 3: scalar 1--20 funniness.} Hypothesis: a smaller scale with a ``seasoned comedy judge'' persona would reduce anchoring. Result: \textbf{82.7\%} of scores fell in a four-point band (13--16); 22.2\% of headlines had within-headline spread $\le 1$ point. Reducing scale width did not fix the problem.

These failures are \textit{paradigm-level}: the judge assigns similar numbers to different jokes on the same headline when every candidate is already a humor attempt, the setting SemEval MWAHAHA uses. Absolute interfaces cannot produce a cross-model leaderboard here regardless of rubric quality.

\subsection{Failure mode B: judge-model self-preference (5-headline pilot)}\label{sec:appendix_failure_mode_b}

Pairwise comparison removes score anchoring but introduces a second failure mode: \textbf{the judge model favors its own family}. On five shared SemEval headlines (\texttt{en\_2001}--\texttt{en\_2005}), we ran full round-robin pairwise tournaments ($36$ pairs/headline) with GPT-5 and Gemini 2.5 Pro serving as judges over the same nine contestants used in the paper.

When a judge faced its own model's output, self-win rates reached \textbf{87.5\%} (GPT-5 as judge, GPT-5 as contestant) and \textbf{87.5\%} (Gemini), with both judges ranking themselves \#1 overall. GPT-OSS-as-judge showed \textbf{$\sim$75\%} self-win; Qwen3-32B-as-judge ranked itself \#4 at \textbf{$\sim$57.5\%} self-win. The four rejected judges agreed on the same winner in only \textbf{$\sim$44\%} of shared duels (79/180); no stable cross-judge humor ordering emerged. These runs motivated screening out proprietary and same-family judges; production evaluation uses Llama 3.3 70B and Qwen 2.5 72B, which showed more coherent humor judgments in this setup and no strong same-family self-preference (under full SemEval coverage, the Llama judge ranks Llama 3.3 70B at \#8, while the Qwen 2.5 72B judge does not elevate Qwen-family contestants: Qwen 3 32B is \#7 and Qwen 2.5 7B Instruct is \#9; Tables~\ref{tab:leaderboard} and~\ref{tab:leaderboard_qwen}).

\subsection{Failure mode C: spurious ties (early pairwise prompt)}\label{sec:appendix_failure_mode_c}

An early pairwise prompt encouraging extended chain-of-thought reasoning produced a \textbf{$\sim$62\% tie rate}, as the judge defaulted to ``equal'' rather than committing to a preference. Subsequent prompt revisions tightened instructions (``trust your first impression'', TIE only when genuinely equal) and enforced structured JSON outputs, reducing spurious ties while retaining GTVH feature tags (Appendix~\ref{sec:appendix_judge_prompts}).

\subsection{Illustrative failure pairs (qualitative audit)}\label{sec:appendix_failure_pairs_qual}

Table~\ref{tab:exp1_en0004} (Failure mode~A) gives a full headline-level example; additional Track~B and V1 pilot duels remain available for manual curation (paths below). Aggregate statistics above establish that failed configurations are unusable for leaderboard construction; qualitative inspection confirms \textit{why}.

\subsection{Production configuration (what survived)}\label{sec:appendix_production_config}

The production HumorRank judge stack combines four properties absent in failed configurations:

\begin{enumerate}
    \item \textbf{Relative, not absolute.} Each judgment is a preference between two jokes on the \textit{same} headline under the structured pairwise prompt.
    \item \textbf{Theory-grounded structure.} GTVH mechanism and delivery tags make preferences auditable.
    \item \textbf{Bias mitigation.} Position swapping, $T{=}0.1$, and rejection of self-preferring proprietary judges (Failure mode B).
    \item \textbf{Cross-judge validation.} Llama 3.3 70B + Qwen 2.5 72B yield Kendall $\tau = 0.889$ on full round-robin for both SemEval and HTB ($25{,}200$ pooled duels).
\end{enumerate}


\newpage

\section{LLM-as-a-Judge Prompting Framework}
\label{sec:appendix_judge_prompts}

The following prompt template was used for all pairwise comparisons in the HumorRank evaluation pipeline. The judge models received a system prompt establishing their role as comedy critics, followed by a structured user prompt presenting two jokes for comparison. All 10,800 automated tournament comparisons for each judge utilized this exact template.

\vspace{0.4em}

\definecolor{prompthdr}{RGB}{15, 23, 42}
\definecolor{systemcolor}{RGB}{20, 110, 80}
\definecolor{systembg}{RGB}{240, 250, 245}
\definecolor{usercolor}{RGB}{20, 90, 160}
\definecolor{userbg}{RGB}{237, 245, 255}
\definecolor{tagcolor}{RGB}{140, 60, 180}
\definecolor{keycolor}{RGB}{180, 90, 0}

\begin{figure}[!ht]
\centering
\begin{tcolorbox}[
  enhanced, colback=white, colframe=black!10,
  arc=6pt, boxrule=0.4pt,
  left=0pt, right=0pt, top=0pt, bottom=8pt,
  after skip=8pt,
  drop shadow={black!8, opacity=0.4, shadow xshift=0pt, shadow yshift=-1.5pt}
]
  \begin{tcolorbox}[
    enhanced, colback=prompthdr, colframe=prompthdr,
    arc=5pt, sharp corners=south, boxrule=0pt,
    left=8pt, right=8pt, top=5pt, bottom=5pt
  ]
  \begin{tabular}{@{}l@{\hspace{10pt}}l@{}}
    {\scriptsize\ttfamily\color{white!40} system} &
    {\scriptsize\sffamily\bfseries\color{white} Pairwise Judge Prompt: System}
  \end{tabular}
  \end{tcolorbox}
  \vspace{4pt}
  \hspace{8pt}\begin{minipage}{0.94\linewidth}
  \begin{tcolorbox}[
    enhanced, colback=systembg, colframe=systemcolor!25,
    leftrule=3pt, rightrule=0.4pt, toprule=0.4pt, bottomrule=0.4pt,
    arc=2pt, left=8pt, right=6pt, top=5pt, bottom=5pt
  ]
{\ttfamily\scriptsize\color{black!85}
"You are a comedy critic judging which of two jokes is funnier.\textbackslash n"\\
"Analyze both the underlying logic (humor mechanisms) and the presentation (delivery).\textbackslash n"\\
"Be direct and honest. If one joke is clearly better, pick it. "\\
"If they are genuinely equal in quality, say TIE.\textbackslash n"\\
"Do not overthink it; trust your first impression. Output JSON only."
}
  \end{tcolorbox}
  \end{minipage}
  \vspace{2pt}
\end{tcolorbox}

\begin{tcolorbox}[
  enhanced, colback=white, colframe=black!10,
  arc=6pt, boxrule=0.4pt,
  left=0pt, right=0pt, top=0pt, bottom=8pt,
  after skip=12pt,
  drop shadow={black!8, opacity=0.4, shadow xshift=0pt, shadow yshift=-1.5pt}
]
  \begin{tcolorbox}[
    enhanced, colback=prompthdr, colframe=prompthdr,
    arc=5pt, sharp corners=south, boxrule=0pt,
    left=8pt, right=8pt, top=5pt, bottom=5pt
  ]
  \begin{tabular}{@{}l@{\hspace{10pt}}l@{}}
    {\scriptsize\ttfamily\color{white!40} user} &
    {\scriptsize\sffamily\bfseries\color{white} Pairwise Judge Prompt: User}
  \end{tabular}
  \end{tcolorbox}
  \vspace{4pt}
  \hspace{8pt}\begin{minipage}{0.94\linewidth}
  \begin{tcolorbox}[
    enhanced, colback=userbg, colframe=usercolor!25,
    leftrule=3pt, rightrule=0.4pt, toprule=0.4pt, bottomrule=0.4pt,
    arc=2pt, left=8pt, right=6pt, top=5pt, bottom=5pt
  ]
{\ttfamily\scriptsize\color{black!85}
'Prompt: "\textcolor{tagcolor}{\{headline\}}"\textbackslash n\textbackslash n'\\
"JOKE A: \textcolor{tagcolor}{\{joke\_a\}}\textbackslash n\textbackslash n"\\
"JOKE B: \textcolor{tagcolor}{\{joke\_b\}}\textbackslash n\textbackslash n"\\
"Which is funnier? Return JSON:\textbackslash n"\\
"\{\{\textbackslash n"\\
'\quad \textcolor{keycolor}{"reasoning"}: \textcolor{usercolor}{"brief explanation"},\textbackslash n'\\
'\quad \textcolor{keycolor}{"decision"}: \textcolor{usercolor}{"A"} or \textcolor{usercolor}{"B"} or \textcolor{usercolor}{"TIE"},\textbackslash n'\\
'\quad \textcolor{keycolor}{"winner\_humor\_features"}: [list ALL that apply, 1-3, from: \textcolor{tagcolor}{\{mech\_features\}}],\textbackslash n'\\
'\quad \textcolor{keycolor}{"winner\_delivery\_features"}: [list ALL that apply, 1-3, from: \textcolor{tagcolor}{\{deliv\_features\}}],\textbackslash n'\\
'\quad \textcolor{keycolor}{"loser\_features"}: [list ALL that apply, 1-3, from: \textcolor{tagcolor}{\{loser\_features\}}]\textbackslash n'\\
"\}\}"
}
  \end{tcolorbox}
  \end{minipage}
  \vspace{2pt}
\end{tcolorbox}

\noindent
\begin{minipage}[t]{0.32\textwidth}
\begin{tcolorbox}[
  enhanced, colback=systembg, colframe=systemcolor!30,
  leftrule=3pt, rightrule=0.4pt, toprule=0.4pt, bottomrule=0.4pt,
  arc=3pt,
  attach boxed title to top left={yshift=-2pt, xshift=6pt},
  boxed title style={
    colback=systemcolor, colframe=systemcolor,
    arc=2pt, boxrule=0pt,
    left=4pt, right=4pt, top=2pt, bottom=2pt
  },
  title={\color{white}\bfseries\sffamily\scriptsize HUMOR MECHANISMS},
  left=7pt, right=6pt, top=5pt, bottom=5pt
]
{\ttfamily\scriptsize\color{black!80}
"incongruity", "wordplay",\\
"absurdity", "surprise",\\
"irony", "sarcasm",\\
"observational", "narrative"
}
\end{tcolorbox}
\end{minipage}\hfill
\begin{minipage}[t]{0.32\textwidth}
\begin{tcolorbox}[
  enhanced, colback=userbg, colframe=usercolor!30,
  leftrule=3pt, rightrule=0.4pt, toprule=0.4pt, bottomrule=0.4pt,
  arc=3pt,
  attach boxed title to top left={yshift=-2pt, xshift=6pt},
  boxed title style={
    colback=usercolor, colframe=usercolor,
    arc=2pt, boxrule=0pt,
    left=4pt, right=4pt, top=2pt, bottom=2pt
  },
  title={\color{white}\bfseries\sffamily\scriptsize DELIVERY FEATURES},
  left=7pt, right=6pt, top=5pt, bottom=5pt
]
{\ttfamily\scriptsize\color{black!80}
"timing", "conciseness",\\
"deadpan", "escalation",\\
"punchline\_positioning",\\
"framing\_commitment"
}
\end{tcolorbox}
\end{minipage}\hfill
\begin{minipage}[t]{0.32\textwidth}
\begin{tcolorbox}[
  enhanced, colback=botredbg, colframe=botred!30,
  leftrule=3pt, rightrule=0.4pt, toprule=0.4pt, bottomrule=0.4pt,
  arc=3pt,
  attach boxed title to top left={yshift=-2pt, xshift=6pt},
  boxed title style={
    colback=botred, colframe=botred,
    arc=2pt, boxrule=0pt,
    left=4pt, right=4pt, top=2pt, bottom=2pt
  },
  title={\color{white}\bfseries\sffamily\scriptsize LOSER FEATURES},
  left=7pt, right=6pt, top=5pt, bottom=5pt
]
{\ttfamily\scriptsize\color{black!80}
"cliché", "confusing",\\
"offensive", "overexplained",\\
"buried\_punchline",\\
"weak\_punchline"
}
\end{tcolorbox}
\end{minipage}

\vspace{0.6em}
\caption{\textnormal{Prompt template used for pairwise comparisons in all full round-robin runs.} Template variables (\textcolor{tagcolor}{\texttt{\{headline\}}}, \textcolor{tagcolor}{\texttt{\{joke\_a\}}}, \textcolor{tagcolor}{\texttt{\{joke\_b\}}}) are instantiated per comparison. The three feature lists (humor mechanisms, delivery, and loser features) enforce structured and consistent JSON outputs across all evaluations.}
\label{fig:judge_prompt}
\end{figure}
\clearpage
\vspace{2em}
\subsection{Feature Taxonomy Definitions}
\label{sec:feature_taxonomy}

To ensure conceptual clarity regarding the theoretical grounding of the LLM-as-a-judge, the exact linguistic definitions mapped by the judge's JSON schema are provided in Table~\ref{tab:feature_definitions}.

\begin{table}[ht]
\centering
\small
\renewcommand{\arraystretch}{1.2}
\begin{tabular}{lp{12.5cm}}
\toprule
\textbf{Feature} & \textbf{Definition} \\
\midrule
\multicolumn{2}{l}{\textbf{Humor Mechanisms (Deep Structure / GTVH-Aligned)}} \\
\midrule
\texttt{incongruity} & Violation of expectations or semantic mismatch. \\
\texttt{wordplay} & Puns, double meanings, or clever syntactic manipulation. \\
\texttt{absurdity} & Bizarre, surreal, or hilariously illogical situations. \\
\texttt{surprise} & Sudden misdirection or sharp pivot in the punchline. \\
\texttt{irony} & Contrast between expectation and reality, often subverting literal meaning. \\
\texttt{sarcasm} & Mocking or contemptuous irony. \\
\texttt{observational} & Finding humor in universally relatable, everyday situations. \\
\texttt{narrative} & Storytelling structure with characters or extended premise. \\
\midrule
\multicolumn{2}{l}{\textbf{Delivery Features (Surface / Stylistic)}} \\
\midrule
\texttt{timing} & Rhythmic pacing, effective beat control via punctuation. \\
\texttt{conciseness} & Economy of delivery; punchy and no wasted words. \\
\texttt{deadpan} & Flat affect, understated phrasing of absurd content. \\
\texttt{escalation} & Progressive build-up of absurdity or tension before the payoff. \\
\texttt{punchline\_positioning} & The punchline lands at the absolute, optimal structure-end. \\
\texttt{framing\_commitment} & Total consistency of the comedic voice or bit without wavering. \\
\midrule
\multicolumn{2}{l}{\textbf{Loser Features (Flaws / Incongruity-Resolution Failures)}} \\
\midrule
\texttt{cliché} & Overused, tired premise or punchline. \\
\texttt{confusing} & Incongruity was unresolvable; didn't make sense. \\
\texttt{offensive} & Mean-spirited or crosses acceptable bounds without comedic payoff. \\
\texttt{overexplained} & Kills the joke by spelling it out too explicitly. \\
\texttt{buried\_punchline} & Punchline exists, but is hidden mid-sentence or poorly placed. \\
\texttt{weak\_punchline} & Structural setup was okay, but the payoff was trivial or unfunny. \\
\bottomrule
\end{tabular}
\caption{Taxonomy of Humor Mechanisms, Delivery Features, and Failure Modes evaluated by the LLM-as-a-judge. These definitions align the evaluation protocol with General Theory of Verbal Humor (GTVH) constructs.}
\label{tab:feature_definitions}
\end{table}

\newpage
\FloatBarrier
\section{Tournament Budget Ablation}
\label{sec:appendix_budget_ablation}

We evaluate Swiss 2RR and Swiss 3RR schedules with the same Adaptive Swiss Pairing code and budget definitions used in the main pipeline. Tables and figures below are grouped by benchmark (SemEval, then HTB) and, within each benchmark, by budget mode (Full RR, Swiss 3RR, Swiss 2RR). All Swiss budget tables use the Llama judge labels unless noted; HTB Qwen judge Swiss tables appear in \ref{sec:appendix_htb_qwen_budget}.

\subsection{Cross-judge $\tau$ by budget mode}\label{sec:appendix_budget_tau}

Table~\ref{tab:appendix_budget_tau} summarizes the Kendall $\tau$ rank correlation across different budget modes to measure cross-judge agreement.

\begin{table}[H]
\centering
\small
\begin{tabular}{lccc}
\toprule
\textbf{Comparison} & \textbf{Full RR} & \textbf{Swiss 2RR} & \textbf{Swiss 3RR} \\
\midrule
SemEval: Llama vs Qwen & 0.889 & 0.667 & \textbf{0.889} \\
HTB: Llama vs Qwen & 0.889 & 1.000 & 0.889 \\
SemEval vs HTB (Llama) & 0.889 & 0.778 & 0.778 \\
SemEval vs HTB (Qwen) & 0.889 & 0.778 & 0.889 \\
\midrule
\textbf{Average} & \textbf{0.889} & 0.806 & \textbf{0.861} \\
\bottomrule
\end{tabular}
\caption{Kendall $\tau$ across budget modes (four cross-judge / cross-benchmark cells). Swiss 3RR restores SemEval Llama$\leftrightarrow$Qwen agreement to Full RR levels.}
\label{tab:appendix_budget_tau}
\end{table}

Cross-judge stability is a budget effect: both benchmarks agree at Full RR and 3RR ($\tau = 0.889$); Swiss 2RR ($\sim$22\% budget) is an under-budget stress test where mid-tier rankings become volatile and cross-judge $\tau$ drops on SemEval (0.667) while remaining at 1.000 on HTB.
\FloatBarrier

\subsection{Swiss 3RR rank stability (SemEval, Llama judge)}\label{sec:appendix_swiss3rr_semeval_ranks}

Table~\ref{tab:appendix_budget_ranks} presents the rank stability of evaluated models on the SemEval benchmark using the Llama judge under the Swiss 3RR schedule.

\begin{table}[H]
\centering
\small
\begin{tabular}{@{}lp{2.8in}@{}}
\toprule
\textbf{Model} & \textbf{Rank (Full RR / Swiss 3RR / Swiss 2RR)} \\
\midrule
GPT-5 & 1 / 1 / 1 \\
Kimi K2 & 2 / 2 / 4 \\
Gemini 2.5 Pro & 3 / 3 / 2 \\
HumorGen-7B & 4 / 5 / 6 \\
Claude 3.5 Haiku & 5 / 7 / 5 \\
GPT OSS 120B & 6 / 4 / 3 \\
Qwen 3 32B & 7 / 6 / 7 \\
Llama 3.3 70B & 8 / 8 / 8 \\
Qwen 2.5 7B Instruct & 9 / 9 / 9 \\
\bottomrule
\end{tabular}
\caption{SemEval Llama judge ranks under Full RR, Swiss 3RR, and Swiss 2RR. Ranks \#1 (GPT-5), \#8 (Llama 3.3 70B), and \#9 (Qwen 2.5 7B Instruct) are invariant across all three schedules; mid-tier models reorder most under Swiss 2RR.}
\label{tab:appendix_budget_ranks}
\end{table}
\FloatBarrier

\subsection{Swiss 3RR rank stability (HTB, Llama judge)}\label{sec:appendix_swiss3rr_htb_ranks}

Table~\ref{tab:appendix_htb_budget_ranks} presents the rank stability of evaluated models on the HTB benchmark using the Llama judge under the Swiss 3RR schedule.

\begin{table}[H]
\centering
\small
\begin{tabular}{@{}lp{2.8in}@{}}
\toprule
\textbf{Model} & \textbf{Rank (Full RR / Swiss 3RR / Swiss 2RR)} \\
\midrule
GPT-5 & 1 / 1 / 1 \\
Kimi K2 & 2 / 2 / 2 \\
HumorGen-7B & 3 / 3 / 6 \\
Claude 3.5 Haiku & 4 / 6 / 4 \\
Gemini 2.5 Pro & 5 / 5 / 5 \\
GPT OSS 120B & 6 / 4 / 3 \\
Qwen 3 32B & 7 / 7 / 7 \\
Llama 3.3 70B & 8 / 8 / 8 \\
Qwen 2.5 7B Instruct & 9 / 9 / 9 \\
\bottomrule
\end{tabular}
\caption{HTB Llama judge ranks under Full RR, Swiss 3RR, and Swiss 2RR. Ranks \#1 (GPT-5), \#8 (Llama 3.3 70B), and \#9 (Qwen 2.5 7B Instruct) are invariant; HumorGen-7B drops from \#3 to \#6 under Swiss 2RR.}
\label{tab:appendix_htb_budget_ranks}
\end{table}
\FloatBarrier

\subsection{SemEval budget ablation (Llama judge)}
\label{sec:appendix_semeval_budget}

SemEval uses 300 prompts and $K{=}9$ contestants. Full RR runs 36 pairs per prompt (10,800 judgments); Swiss 3RR and Swiss 2RR subsample to 12 and 8 pairs per prompt (3,600 and 2,400 judgments). Table~\ref{tab:appendix_budget_ranks} summarizes rank stability across modes on this benchmark.

\begin{figure}[htbp]
    \centering
    \includegraphics[width=0.72\linewidth]{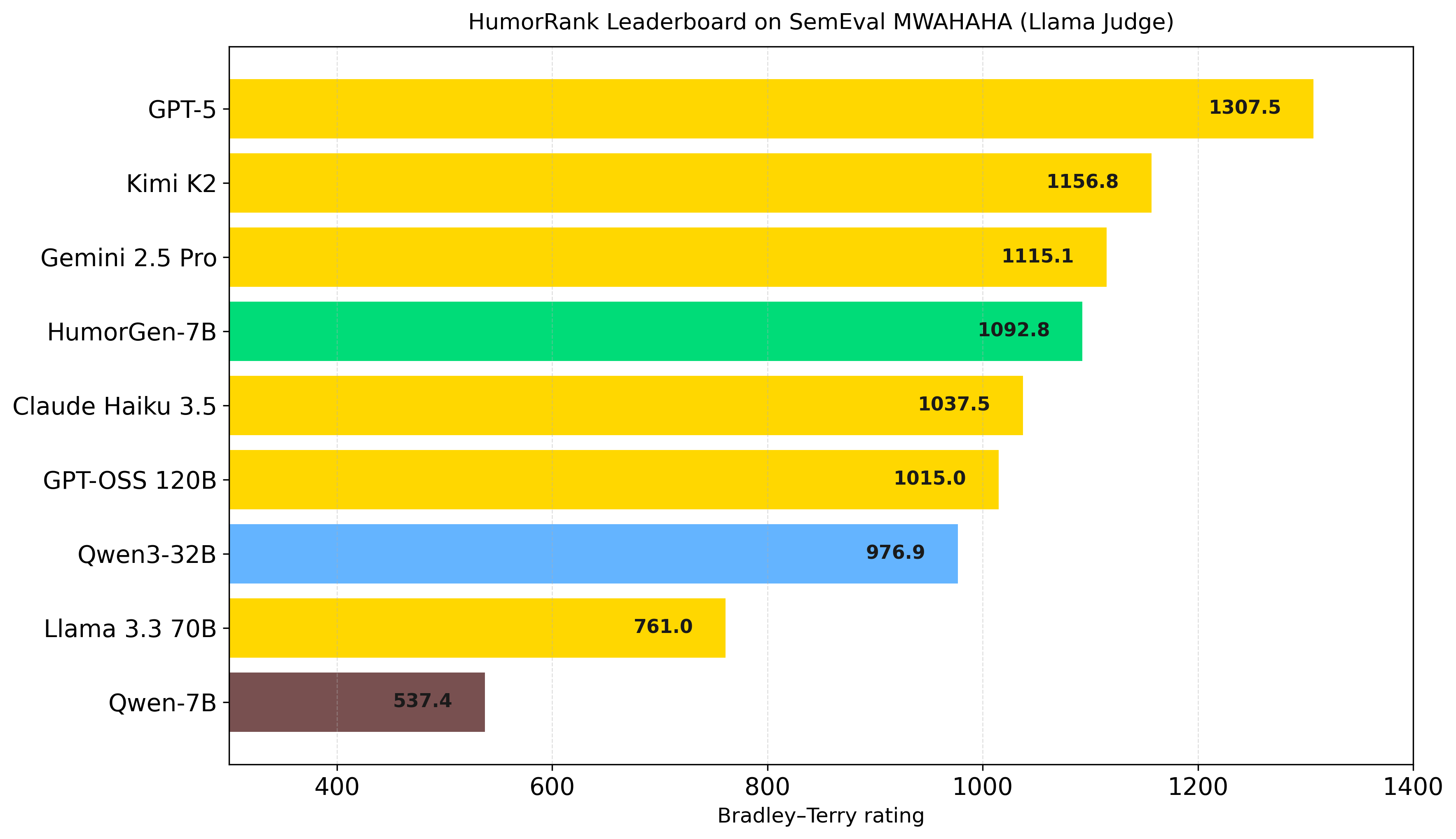}
    \vspace{0.4em}
    \includegraphics[width=0.72\linewidth]{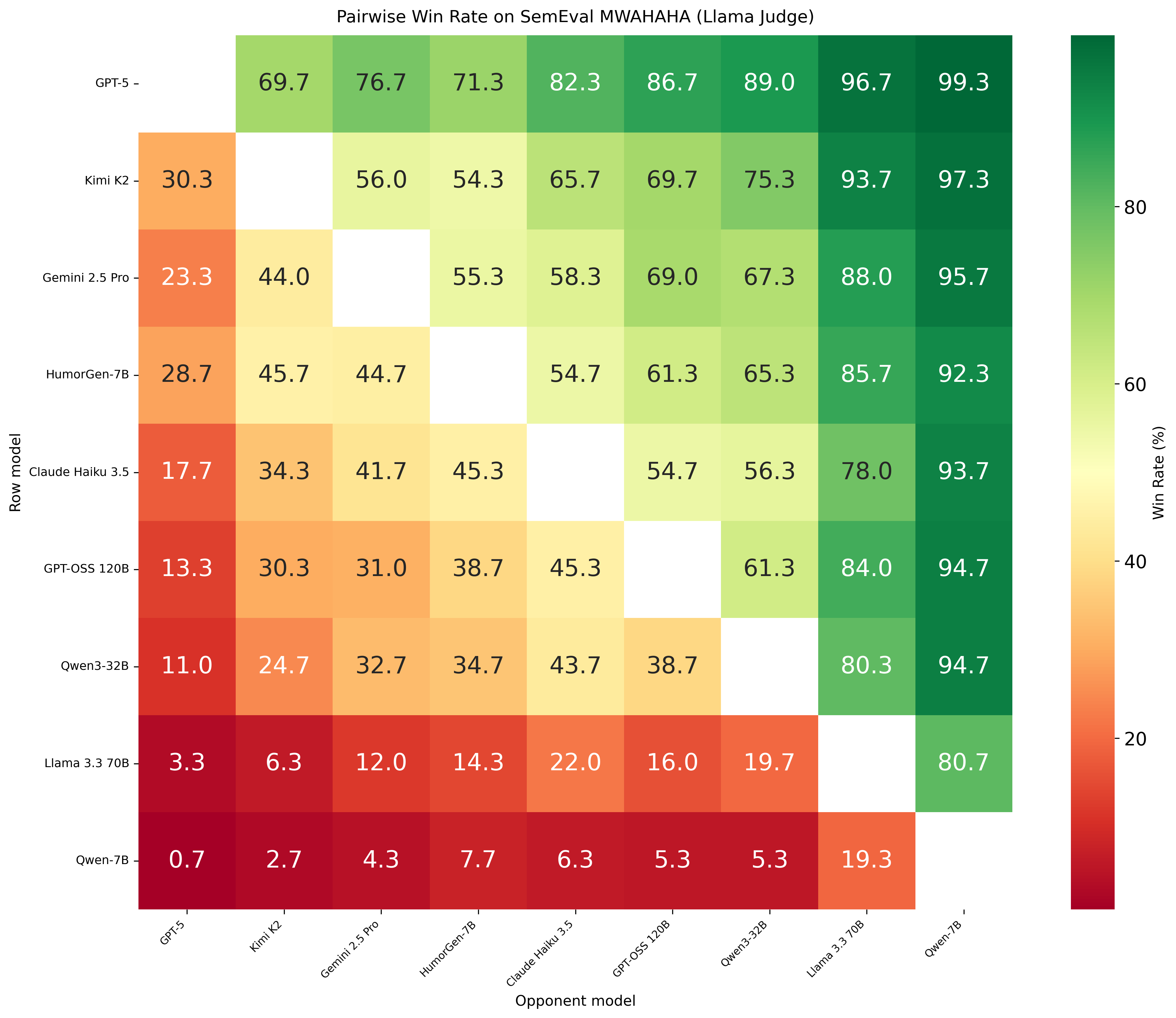}
    \caption{\textbf{SemEval, Full RR, Llama judge} (Table~\ref{tab:leaderboard}). \textbf{Top:} Bradley--Terry leaderboard with 95\% confidence intervals (10,800 judgments). \textbf{Bottom:} Pairwise win-rate heatmap. \textit{Observation:} clear frontier/mid/baseline separation; cross-judge agreement with Qwen Full RR is Kendall $\tau = 0.889$.}
    \label{fig:appendix_full_rr_llama}
\end{figure}

\paragraph{Full RR (100\% budget).}
The SemEval rows in Table~\ref{tab:leaderboard} report the primary Full RR leaderboard. Figure~\ref{fig:appendix_full_rr_llama} visualizes the same Full RR run: GPT-5 and Kimi K2 lead the field; HumorGen-7B ranks 4th on SemEval; Llama 3.3 70B and Qwen 2.5 7B Instruct anchor the bottom tier.

\paragraph{Swiss 3RR ($\sim$33\% budget).}
At 12 pairs/prompt, Swiss 3RR preserves the Full RR ordering at the top and bottom: GPT-5 remains \#1; Llama 3.3 70B and Qwen 2.5 7B Instruct remain \#8 and \#9. HumorGen-7B shifts one rank (4$\rightarrow$5) while mid-tier models show modest reordering. Kendall $\tau$ vs.\ SemEval Full RR is 0.889, and SemEval Llama$\leftrightarrow$Qwen cross-judge $\tau$ matches Full RR (0.889; Table~\ref{tab:appendix_budget_tau}).

\begin{figure}[htbp]
    \centering
    \includegraphics[width=0.72\linewidth]{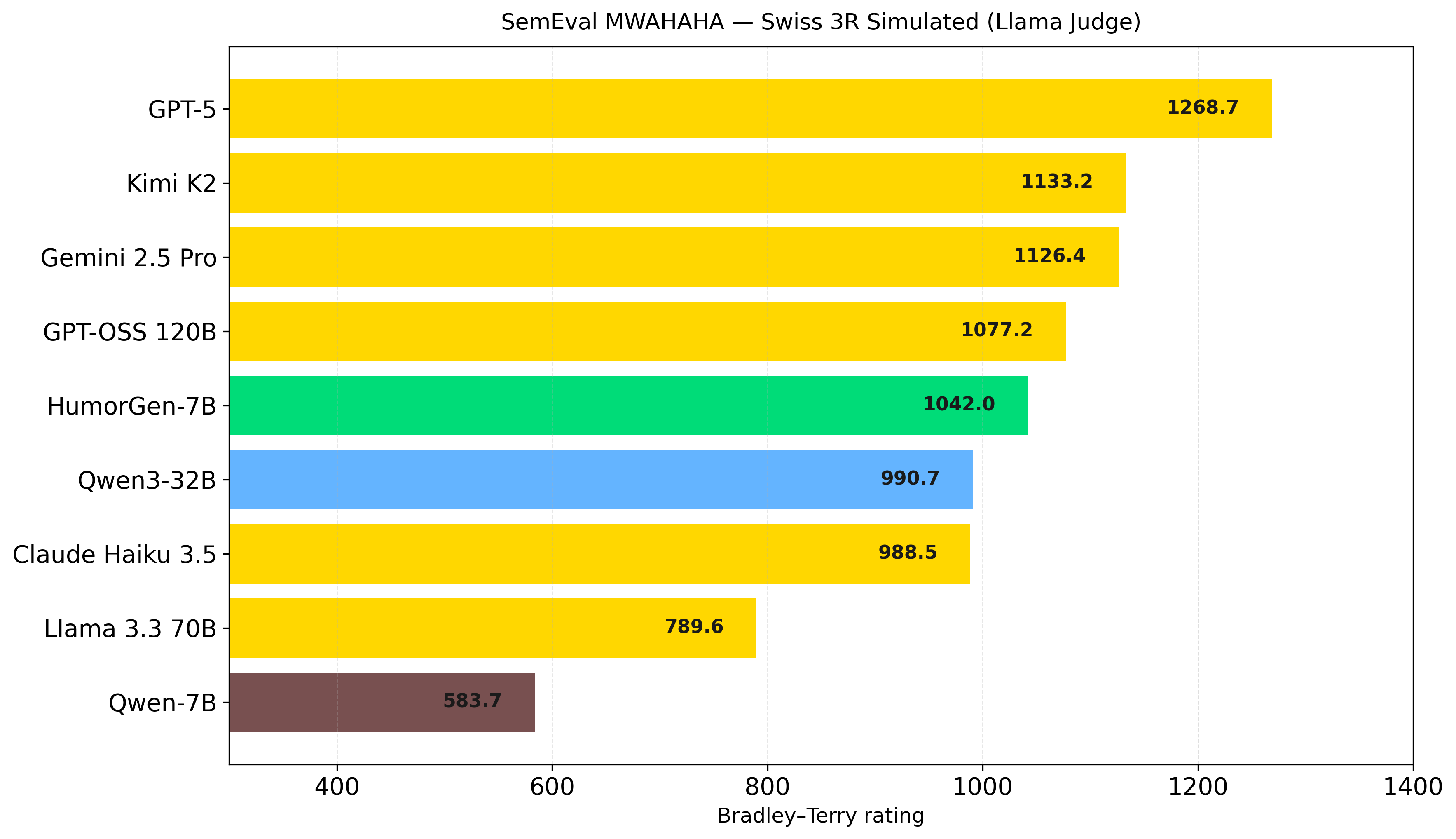}
    \vspace{0.4em}
    \includegraphics[width=0.72\linewidth]{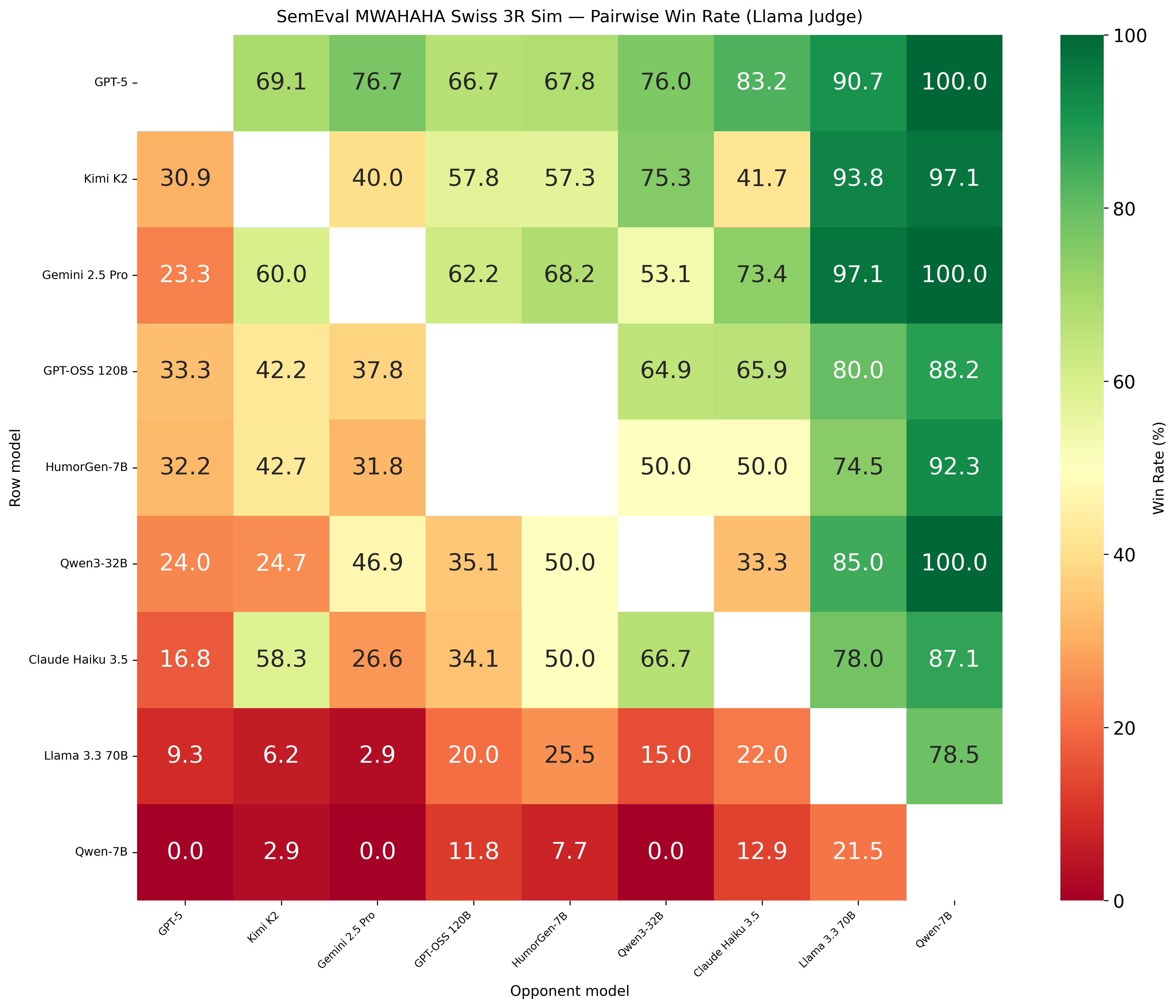}
    \caption{\textbf{SemEval, Swiss 3RR, Llama judge} (Table~\ref{tab:appendix_swiss_3r_leaderboard}). \textbf{Top:} BT leaderboard from 3,600 judged pairs. \textbf{Bottom:} Win-rate heatmap. \textit{Observation:} Swiss 3RR at one-third of Full RR comparisons recovers the same cross-judge rank correlation ($\tau = 0.889$) and keeps anchor ranks \#1/\#8/\#9 fixed.}
    \label{fig:budget_swiss_3r}
\end{figure}

\begin{table}[H]
\centering
\small
\begin{tabular}{llcccc}
\toprule
\textbf{Rank} & \textbf{Model} & \textbf{BT Rating} & \textbf{Stable Elo} & \textbf{95\% CI} & \textbf{Win Rate} \\
\midrule
1 & GPT-5 & 1268.7 & 1253.2 & $[1243.0, 1302.7]$ & 76.7\% \\
2 & Kimi-K2 & 1133.2 & 1132.7 & $[1110.8, 1160.9]$ & 63.9\% \\
3 & Gemini 2.5 Pro & 1126.4 & 1108.6 & $[1105.0, 1151.3]$ & 52.8\% \\
4 & GPT OSS 120B & 1077.3 & 1086.3 & $[1048.5, 1106.8]$ & 49.3\% \\
5 & HumorGen-7B & 1042.0 & 1058.5 & $[1012.9, 1070.7]$ & 64.1\% \\
6 & Qwen 3 32B & 990.7 & 997.8 & $[965.5, 1021.9]$ & 35.2\% \\
7 & Claude 3.5 Haiku & 988.5 & 987.4 & $[962.1, 1014.6]$ & 47.9\% \\
8 & Llama 3.3 70B & 789.7 & 795.2 & $[758.1, 818.1]$ & 35.1\% \\
9 & Qwen 2.5 7B Instruct & 583.7 & 580.3 & $[537.6, 620.0]$ & 12.4\% \\
\bottomrule
\end{tabular}
\caption{\textbf{SemEval, Swiss 3RR, Llama judge.} Rank $\tau$ vs.\ Full RR $= 0.889$; cross-judge $\tau$ (Llama vs.\ Qwen on SemEval) $= 0.889$. Anchor ranks \#1, \#8, \#9 unchanged from Table~\ref{tab:appendix_budget_ranks}.}
\label{tab:appendix_swiss_3r_leaderboard}
\end{table}

\paragraph{Swiss 2RR ($\sim$22\% budget).}
At 8 pairs/prompt, anchor ranks \#1, \#8, and \#9 remain stable, but mid-tier ordering becomes noisier (e.g., Kimi K2 2$\rightarrow$4, Gemini 2.5 Pro 3$\rightarrow$2, HumorGen-7B 4$\rightarrow$6). SemEval Llama$\leftrightarrow$Qwen cross-judge $\tau$ drops to 0.667 (Table~\ref{tab:appendix_budget_tau}), so we treat Swiss 2RR as a minimum-budget stress test rather than the recommended scaling mode.

\begin{figure}[htbp]
    \centering
    \includegraphics[width=0.72\linewidth]{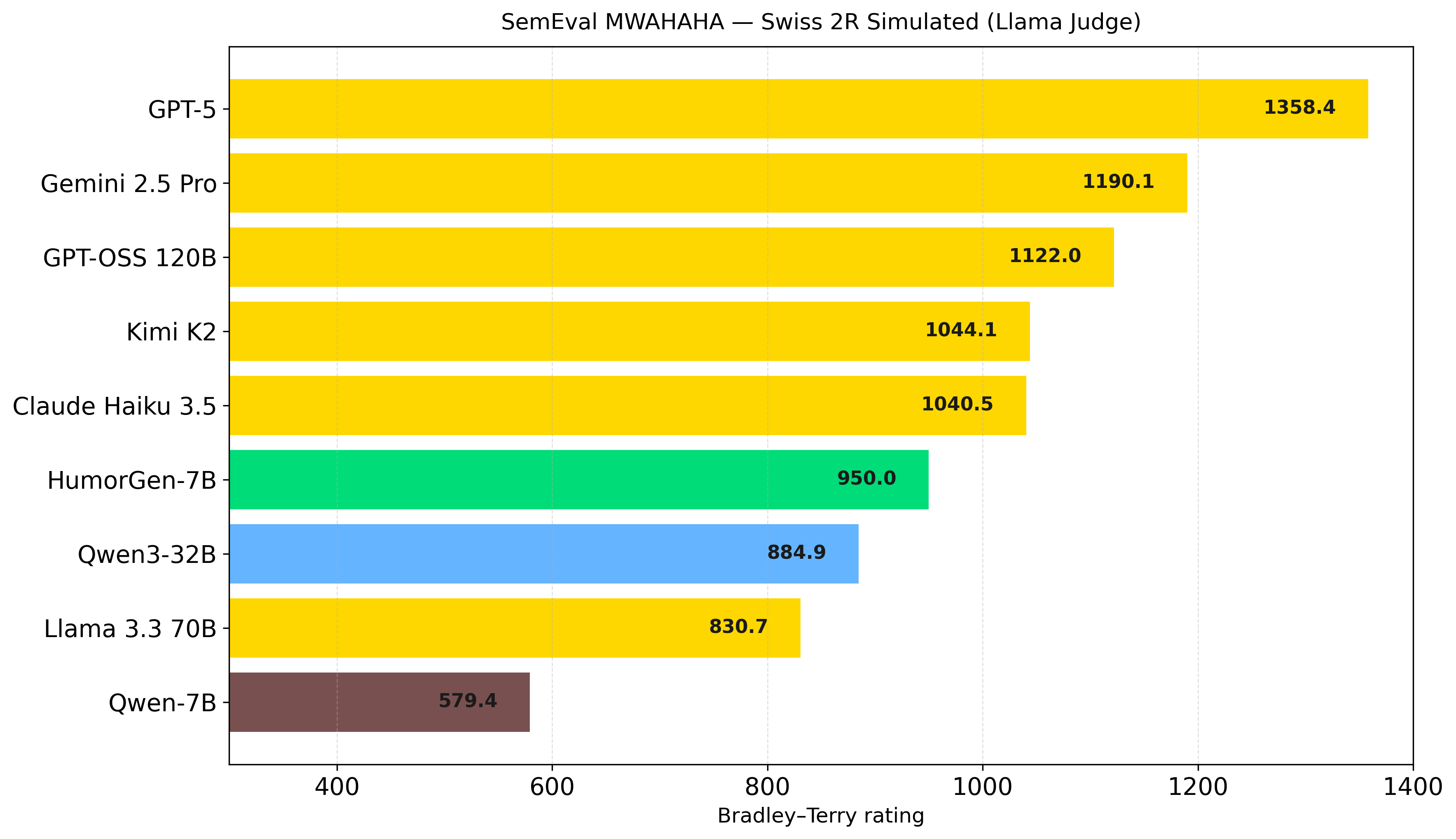}
    \vspace{0.4em}
    \includegraphics[width=0.72\linewidth]{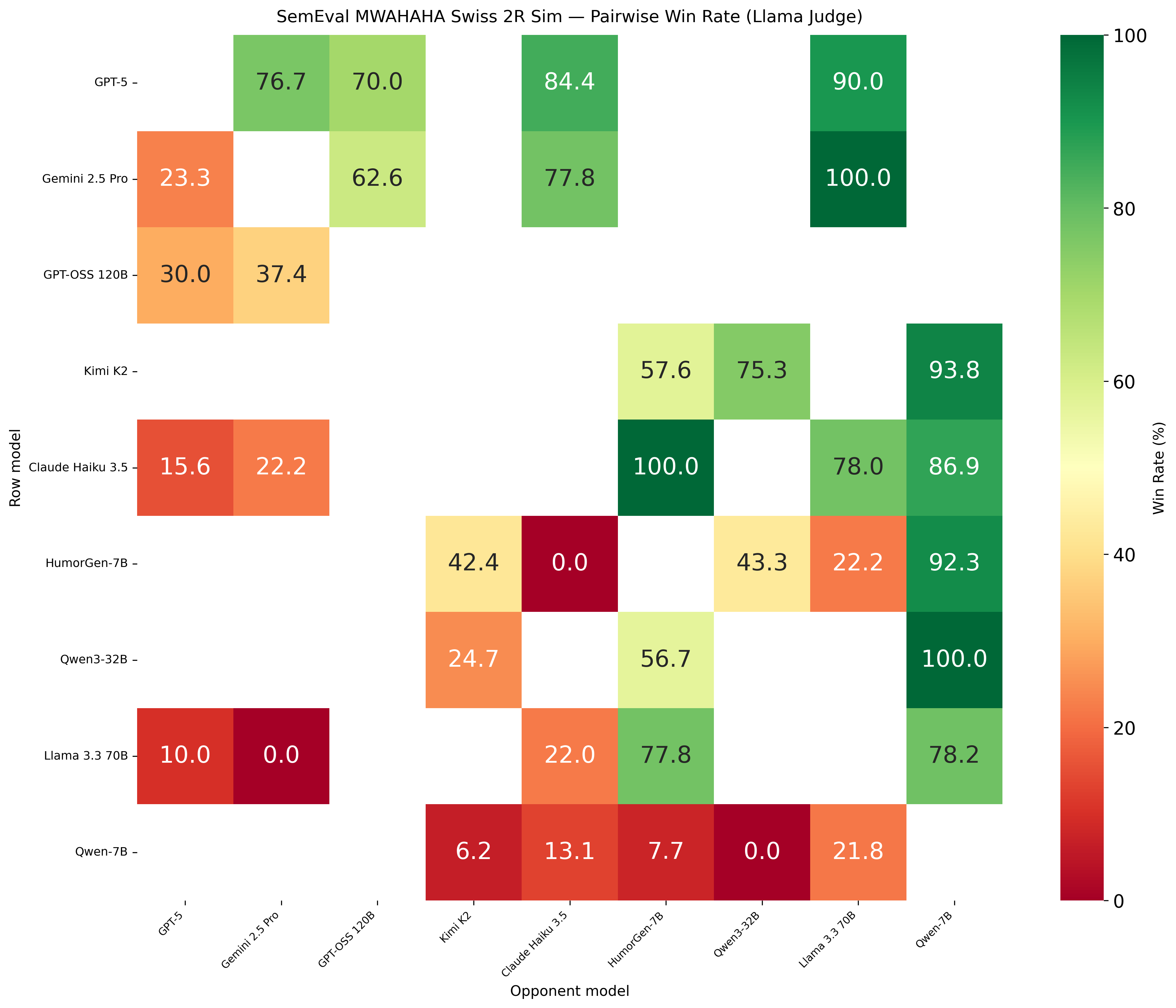}
    \caption{\textbf{SemEval, Swiss 2RR, Llama judge} (Table~\ref{tab:appendix_swiss_2r_leaderboard}). \textbf{Top:} BT leaderboard from 2,400 judged pairs. \textbf{Bottom:} Win-rate heatmap. \textit{Observation:} lowest-budget mode preserves top/bottom anchors but increases mid-tier rank volatility and reduces cross-judge agreement ($\tau = 0.667$).}
    \label{fig:budget_swiss_2r}
\end{figure}

\begin{table}[H]
\centering
\small
\begin{tabular}{llcccc}
\toprule
\textbf{Rank} & \textbf{Model} & \textbf{BT Rating} & \textbf{Stable Elo} & \textbf{95\% CI} & \textbf{Win Rate} \\
\midrule
1 & GPT-5 & 1358.4 & 1307.9 & $[1304.3, 1423.6]$ & 79.3\% \\
2 & Gemini 2.5 Pro & 1190.1 & 1103.0 & $[1135.1, 1246.6]$ & 45.3\% \\
3 & GPT OSS 120B & 1122.0 & 1037.6 & $[1056.8, 1207.7]$ & 35.7\% \\
4 & Kimi-K2 & 1044.1 & 1117.4 & $[985.3, 1111.5]$ & 68.8\% \\
5 & Claude 3.5 Haiku & 1040.5 & 1028.4 & $[993.9, 1085.8]$ & 55.3\% \\
6 & HumorGen-7B & 950.0 & 1000.7 & $[898.4, 1008.2]$ & 66.5\% \\
7 & Qwen 3 32B & 884.9 & 960.9 & $[816.5, 957.6]$ & 31.8\% \\
8 & Llama 3.3 70B & 830.7 & 841.7 & $[790.4, 861.7]$ & 42.3\% \\
9 & Qwen 2.5 7B Instruct & 579.4 & 602.4 & $[529.9, 631.3]$ & 13.2\% \\
\bottomrule
\end{tabular}
\caption{\textbf{SemEval, Swiss 2RR, Llama judge.} Cross-judge Llama$\leftrightarrow$Qwen $\tau = 0.667$; ranks \#1, \#8, and \#9 remain stable (Table~\ref{tab:appendix_budget_ranks}).}
\label{tab:appendix_swiss_2r_leaderboard}
\end{table}

\subsection{HTB budget ablation (Llama judge)}
\label{sec:appendix_htb_budget_llama}

HTB uses 400 held-out headline prompts with the same nine-model pool. Full RR requires 14,400 judgments per LLM judge; Swiss 3RR and 2RR use 4,800 and 3,200 judgments, respectively. Table~\ref{tab:appendix_htb_budget_ranks} summarizes rank stability across budget modes on this benchmark.

\paragraph{Full RR (100\% budget).}
On HTB Full RR, GPT-5 and Kimi K2 remain in the top two positions; HumorGen-7B ranks 3rd under the Llama judge; Llama 3.3 70B and Qwen 2.5 7B Instruct remain 8th and 9th. Cross-judge agreement with Qwen Full RR is Kendall $\tau = 0.889$ (Table~\ref{tab:appendix_budget_tau}).

\begin{table}[H]
\centering
\small
\begin{tabular}{llcccc}
\toprule
\textbf{Rank} & \textbf{Model} & \textbf{BT Rating} & \textbf{Stable Elo} & \textbf{95\% CI} & \textbf{Win Rate} \\
\midrule
1 & GPT-5 & 1314.7 & 1300.4 & $[1301.3, 1329.0]$ & 84.1\% \\
2 & Kimi K2 & 1242.0 & 1239.1 & $[1229.5, 1255.6]$ & 77.2\% \\
3 & HumorGen-7B & 1097.7 & 1122.8 & $[1084.4, 1109.9]$ & 60.6\% \\
4 & Claude 3.5 Haiku & 1054.2 & 1058.3 & $[1043.2, 1068.5]$ & 55.1\% \\
5 & Gemini 2.5 Pro & 1024.0 & 1009.0 & $[1010.6, 1038.2]$ & 51.3\% \\
6 & GPT OSS 120B & 1009.6 & 1017.2 & $[998.4, 1021.6]$ & 49.5\% \\
7 & Qwen 3 32B & 942.5 & 946.2 & $[928.1, 955.7]$ & 41.3\% \\
8 & Llama 3.3 70B & 791.9 & 795.4 & $[776.4, 808.0]$ & 24.9\% \\
9 & Qwen 2.5 7B Instruct & 523.5 & 511.8 & $[501.8, 549.8]$ & 5.9\% \\
\bottomrule
\end{tabular}
\caption{\textbf{HTB, Full RR, Llama judge} (14,400 judgments). \textit{Observation:} ordering mirrors SemEval at the extremes (GPT-5 \#1; Llama/Qwen 2.5 7B Instruct \#8/\#9); HumorGen-7B ranks 3rd on this held-out benchmark.}
\label{tab:htb_leaderboard_llama}
\end{table}

\paragraph{Swiss 3RR ($\sim$33\% budget).}
Swiss 3RR on HTB preserves ranks \#1, \#8, and \#9 and keeps HumorGen-7B at rank \#3. HTB Llama$\leftrightarrow$Qwen cross-judge $\tau$ remains 0.889 (Table~\ref{tab:appendix_budget_tau}).

\begin{figure}[htbp]
    \centering
    \includegraphics[width=0.72\linewidth]{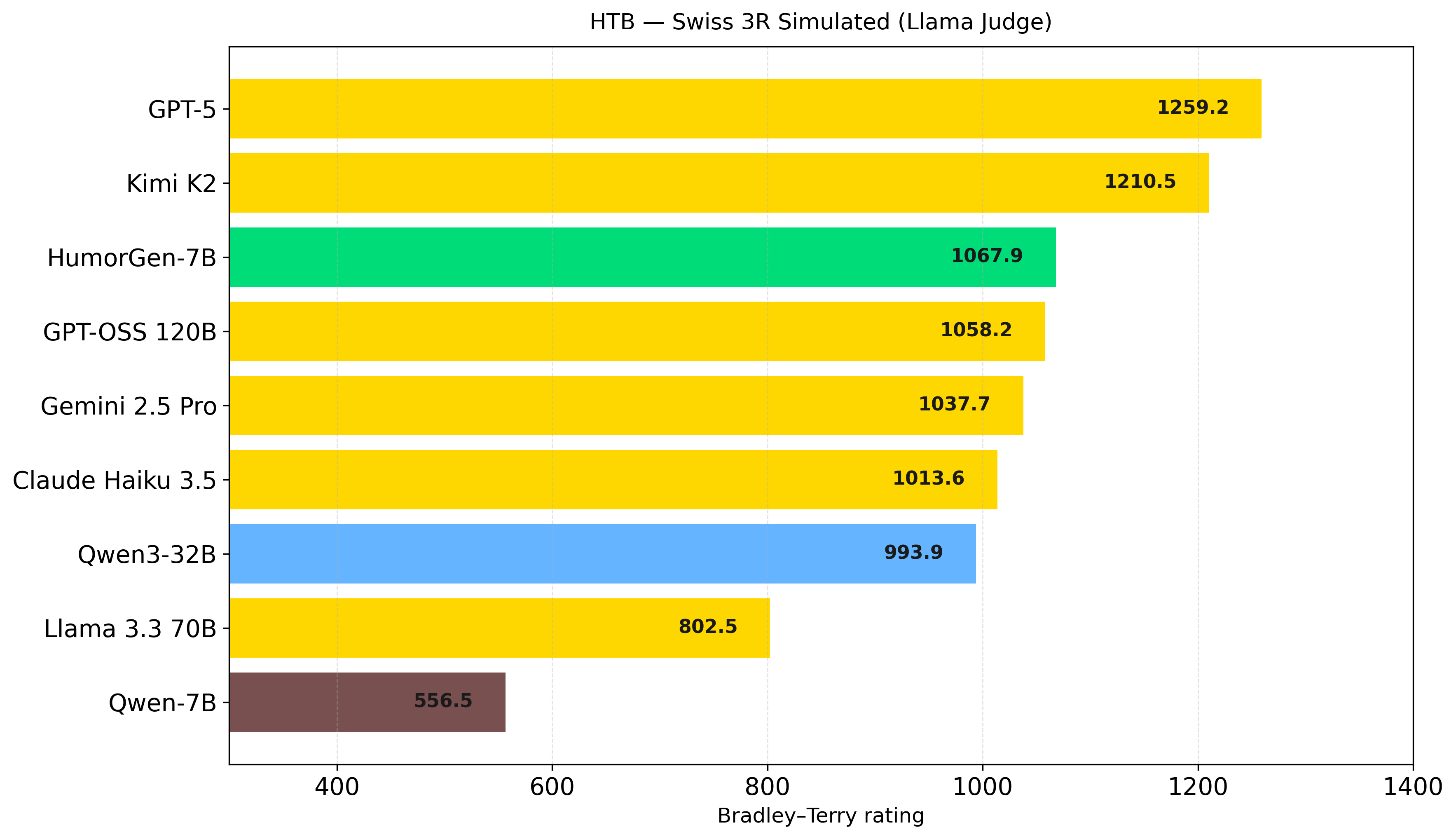}
    \vspace{0.4em}
    \includegraphics[width=0.72\linewidth]{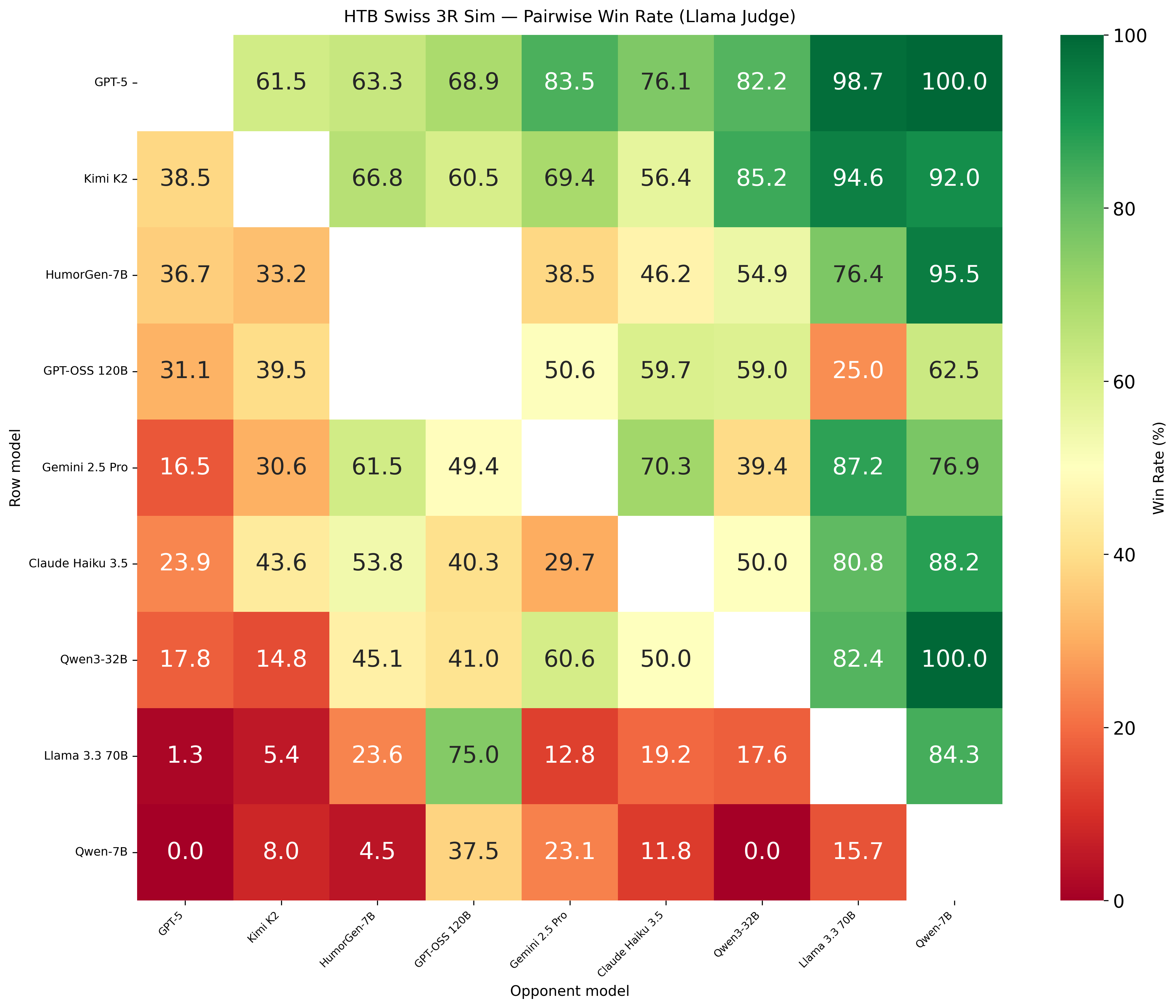}
    \caption{\textbf{HTB, Swiss 3RR, Llama judge} (Table~\ref{tab:htb_swiss_3rr_leaderboard_llama}). \textbf{Top:} BT leaderboard from 4,800 judged pairs. \textbf{Bottom:} Win-rate heatmap. \textit{Observation:} one-third budget retains HTB anchor ranks and cross-judge stability ($\tau = 0.889$).}
    \label{fig:htb_swiss_3rr_llama}
\end{figure}

\begin{table}[H]
\centering
\small
\begin{tabular}{llcccc}
\toprule
\textbf{Rank} & \textbf{Model} & \textbf{BT Rating} & \textbf{Stable Elo} & \textbf{95\% CI} & \textbf{Win Rate} \\
\midrule
1 & GPT-5 & 1259.2 & 1246.7 & $[1236.8, 1282.7]$ & 76.7\% \\
2 & Kimi K2 & 1210.5 & 1204.0 & $[1188.5, 1236.0]$ & 71.4\% \\
3 & HumorGen-7B & 1067.9 & 1055.4 & $[1040.1, 1094.1]$ & 63.9\% \\
4 & GPT OSS 120B & 1058.2 & 1062.8 & $[1032.3, 1088.9]$ & 50.9\% \\
5 & Gemini 2.5 Pro & 1037.7 & 1056.0 & $[1018.5, 1061.0]$ & 42.2\% \\
6 & Claude 3.5 Haiku & 1013.6 & 1025.1 & $[986.8, 1036.6]$ & 51.6\% \\
7 & Qwen 3 32B & 993.9 & 988.3 & $[971.5, 1020.2]$ & 32.5\% \\
8 & Llama 3.3 70B & 802.5 & 789.4 & $[778.3, 827.1]$ & 35.7\% \\
9 & Qwen 2.5 7B Instruct & 556.5 & 572.4 & $[507.5, 595.1]$ & 10.4\% \\
\bottomrule
\end{tabular}
\caption{\textbf{HTB, Swiss 3RR, Llama judge.} Ranks \#1, \#3 (HumorGen-7B), \#8, and \#9 match Full RR anchors; cross-judge HTB $\tau = 0.889$.}
\label{tab:htb_swiss_3rr_leaderboard_llama}
\end{table}

\paragraph{Swiss 2RR ($\sim$22\% budget).}
At minimum budget, HTB anchor ranks \#1, \#8, and \#9 remain fixed, but HumorGen-7B drops from \#3 to \#6 and mid-tier models reorder. Notably, HTB Llama$\leftrightarrow$Qwen $\tau$ rises to 1.000 at 2RR (Table~\ref{tab:appendix_budget_tau}), a benchmark-specific effect we attribute to reduced comparison density rather than improved ranking fidelity.

\begin{figure}[htbp]
    \centering
    \includegraphics[width=0.72\linewidth]{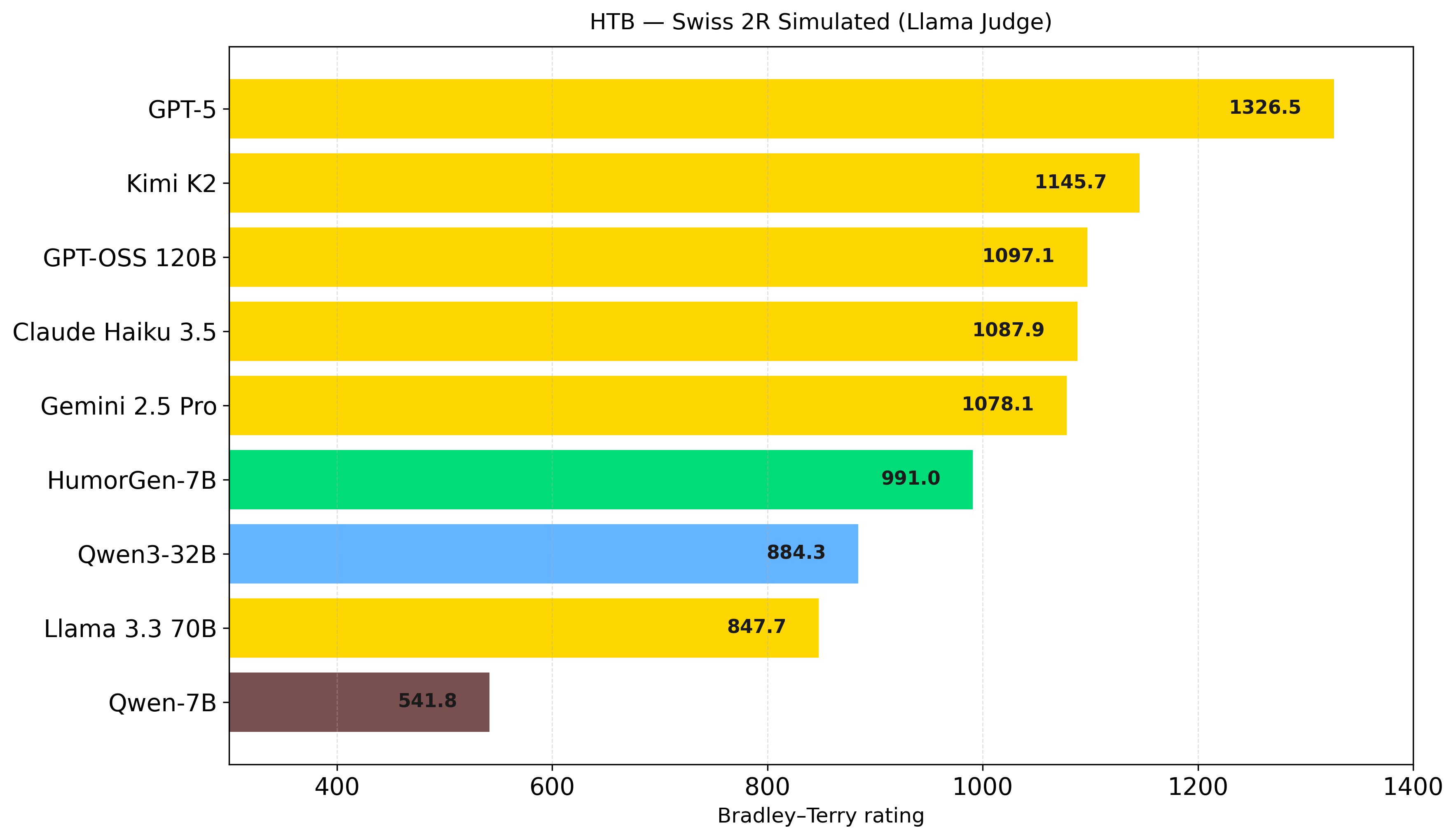}
    \vspace{0.4em}
    \includegraphics[width=0.72\linewidth]{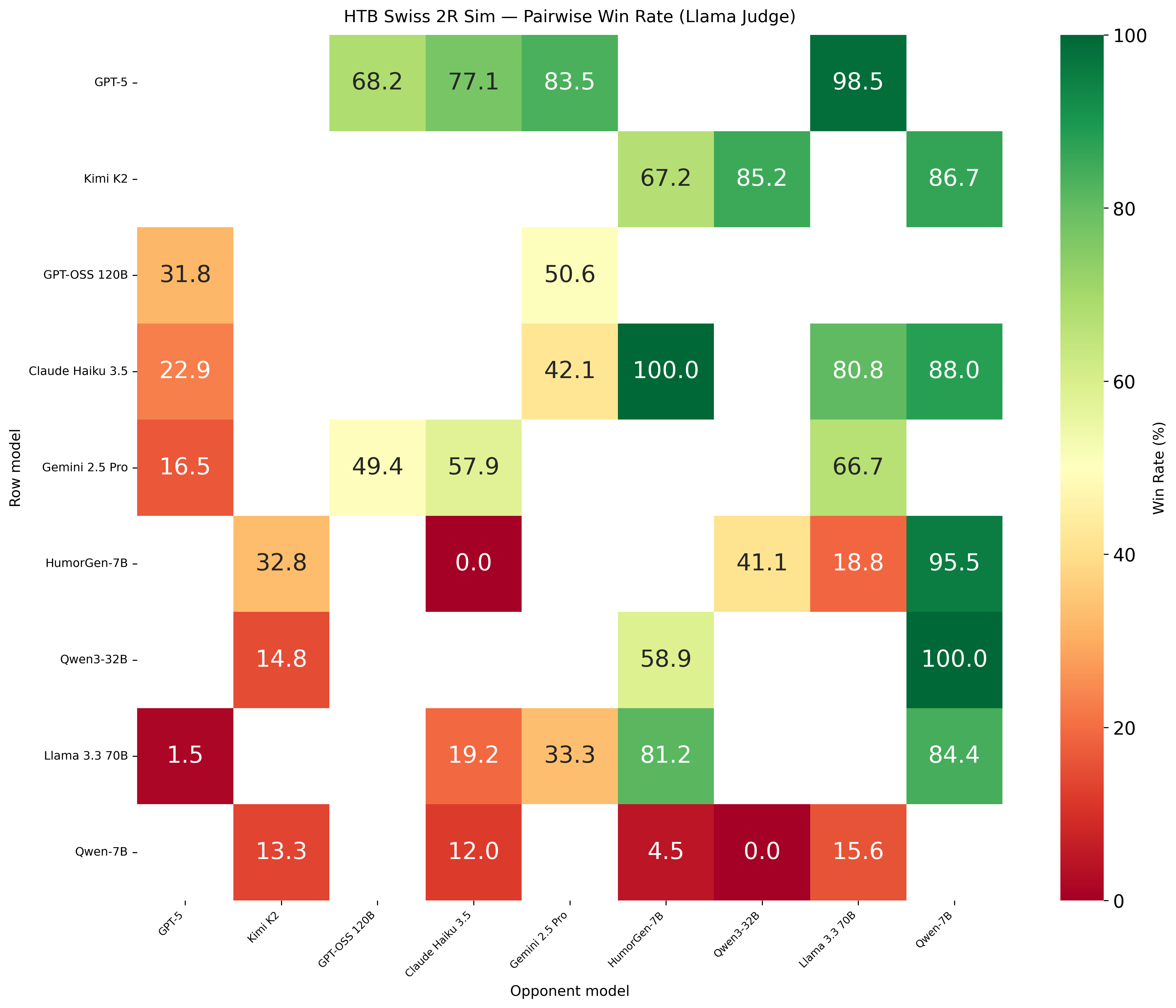}
    \caption{\textbf{HTB, Swiss 2RR, Llama judge} (Table~\ref{tab:htb_swiss_2rr_leaderboard_llama}). \textbf{Top:} BT leaderboard from 3,200 judged pairs. \textbf{Bottom:} Win-rate heatmap. \textit{Observation:} anchor ranks hold, but specialist mid-tier placement becomes less stable (HumorGen-7B 3$\rightarrow$6).}
    \label{fig:htb_swiss_2rr_llama}
\end{figure}

\begin{table}[H]
\centering
\small
\begin{tabular}{llcccc}
\toprule
\textbf{Rank} & \textbf{Model} & \textbf{BT Rating} & \textbf{Stable Elo} & \textbf{95\% CI} & \textbf{Win Rate} \\
\midrule
1 & GPT-5 & 1326.5 & 1303.3 & $[1274.5, 1386.9]$ & 81.4\% \\
2 & Kimi K2 & 1145.7 & 1180.1 & $[1077.0, 1208.1]$ & 77.3\% \\
3 & GPT OSS 120B & 1097.1 & 1043.6 & $[1032.3, 1163.1]$ & 47.5\% \\
4 & Claude 3.5 Haiku & 1087.9 & 1072.2 & $[1046.1, 1137.9]$ & 59.5\% \\
5 & Gemini 2.5 Pro & 1078.1 & 1051.7 & $[1024.0, 1138.7]$ & 33.8\% \\
6 & HumorGen-7B & 991.0 & 1025.6 & $[936.6, 1052.7]$ & 64.4\% \\
7 & Qwen 3 32B & 884.3 & 941.2 & $[807.2, 962.0]$ & 20.7\% \\
8 & Llama 3.3 70B & 847.7 & 829.9 & $[813.1, 888.8]$ & 44.1\% \\
9 & Qwen 2.5 7B Instruct & 541.8 & 552.5 & $[490.5, 583.7]$ & 9.6\% \\
\bottomrule
\end{tabular}
\caption{\textbf{HTB, Swiss 2RR, Llama judge.} Anchor ranks \#1/\#8/\#9 stable; HumorGen-7B drops 3$\rightarrow$6 vs.\ Full RR. Cross-judge HTB $\tau = 1.000$ at this budget (Table~\ref{tab:appendix_budget_tau}).}
\label{tab:htb_swiss_2rr_leaderboard_llama}
\end{table}

\subsection{HTB budget ablation (Qwen judge)}
\label{sec:appendix_htb_qwen_budget}

We replicate the HTB Full RR and Swiss budget modes under the Qwen judge to confirm cross-LLM-judge patterns on the held-out benchmark.

\paragraph{Full RR (100\% budget).}
Qwen judge HTB Full RR agrees with Llama judge HTB at Kendall $\tau = 0.889$; GPT-5 and Kimi K2 remain top-two, with HumorGen-7B 4th (vs.\ 3rd under the Llama judge).

\begin{table}[H]
\centering
\small
\begin{tabular}{llcccc}
\toprule
\textbf{Rank} & \textbf{Model} & \textbf{BT Rating} & \textbf{Stable Elo} & \textbf{95\% CI} & \textbf{Win Rate} \\
\midrule
1 & GPT-5 & 1245.7 & 1278.6 & $[1234.5, 1258.7]$ & 78.8\% \\
2 & Kimi K2 & 1199.5 & 1205.3 & $[1187.2, 1211.7]$ & 73.8\% \\
3 & Claude 3.5 Haiku & 1101.4 & 1094.1 & $[1089.4, 1114.0]$ & 61.8\% \\
4 & HumorGen-7B & 1099.8 & 1099.2 & $[1086.1, 1115.0]$ & 61.6\% \\
5 & GPT OSS 120B & 1024.3 & 1015.9 & $[1013.4, 1037.9]$ & 51.7\% \\
6 & Gemini 2.5 Pro & 1001.3 & 990.4 & $[989.4, 1013.7]$ & 48.8\% \\
7 & Qwen 3 32B & 951.4 & 951.5 & $[941.2, 964.7]$ & 42.4\% \\
8 & Llama 3.3 70B & 757.7 & 746.2 & $[743.1, 773.5]$ & 21.1\% \\
9 & Qwen 2.5 7B Instruct & 618.9 & 618.7 & $[601.0, 634.4]$ & 10.1\% \\
\bottomrule
\end{tabular}
\caption{\textbf{HTB, Full RR, Qwen judge} (14,400 judgments). \textit{Observation:} cross-judge HTB $\tau = 0.889$ vs.\ Table~\ref{tab:htb_leaderboard_llama}; bottom-two ranks unchanged.}
\label{tab:htb_leaderboard_qwen}
\end{table}

\paragraph{Swiss 3RR ($\sim$33\% budget).}

\begin{figure}[htbp]
    \centering
    \includegraphics[width=0.72\linewidth]{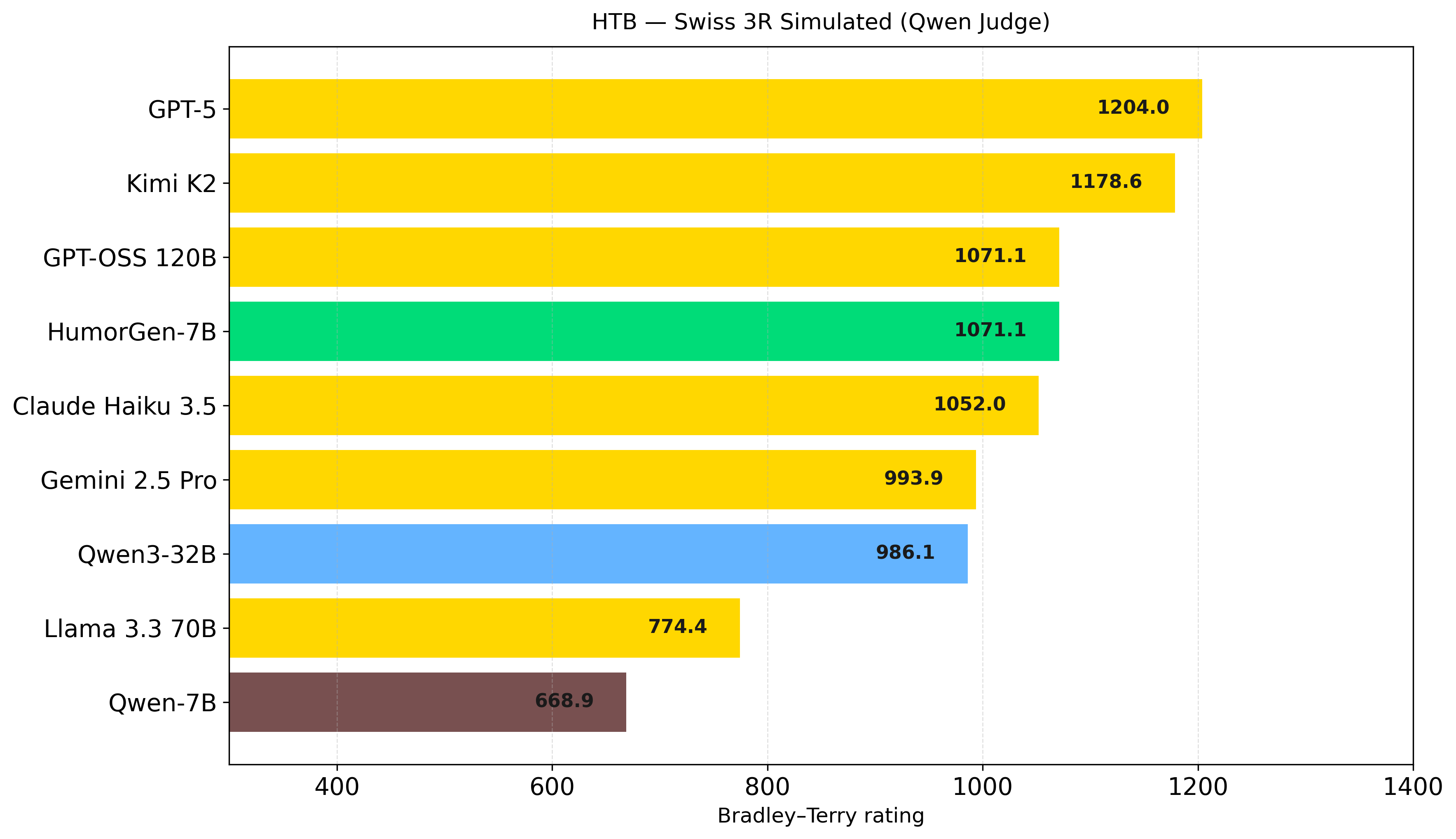}
    \vspace{0.4em}
    \includegraphics[width=0.72\linewidth]{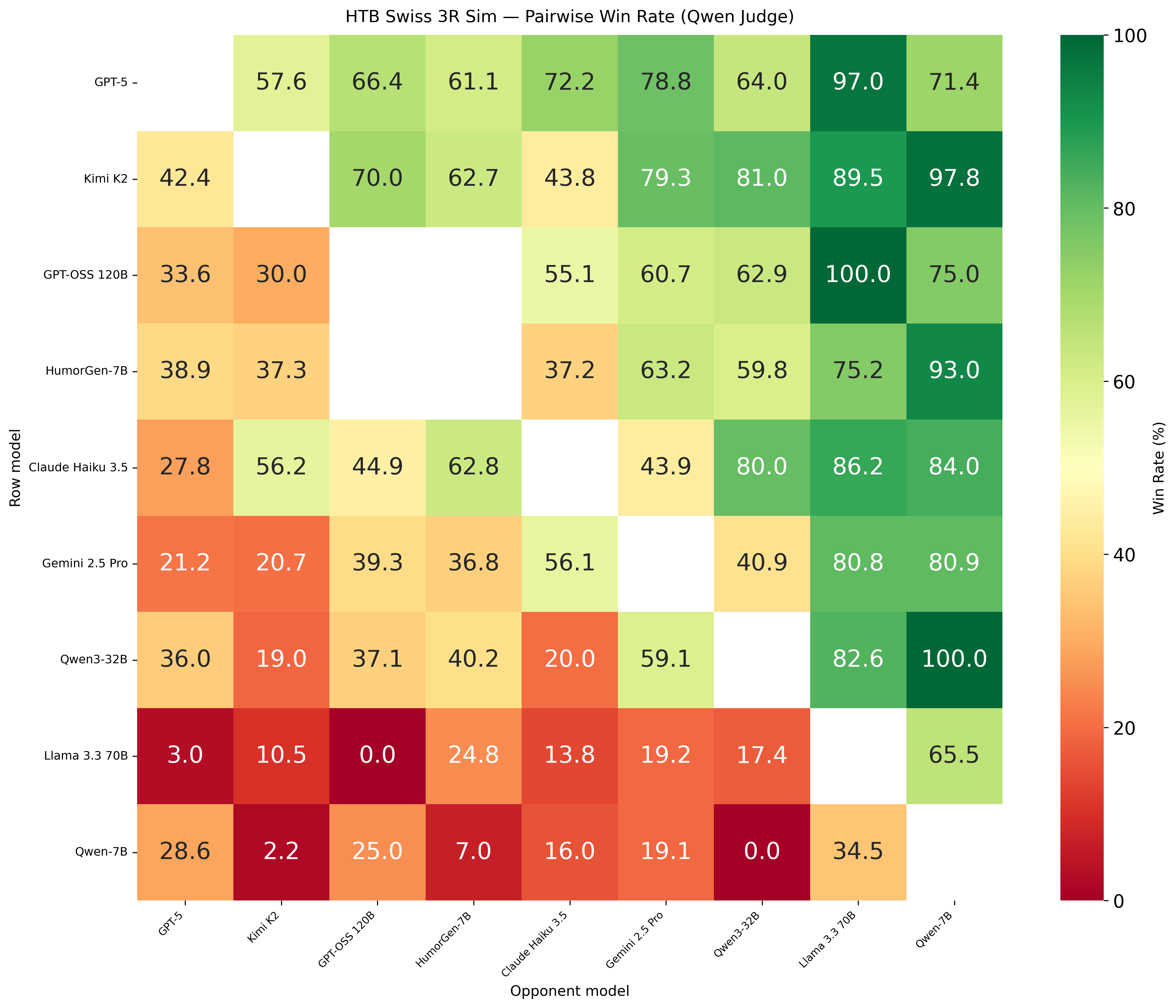}
    \caption{\textbf{HTB, Swiss 3RR, Qwen judge} (Table~\ref{tab:htb_swiss_3rr_leaderboard_qwen}). \textbf{Top:} BT leaderboard from 4,800 judged pairs. \textbf{Bottom:} Win-rate heatmap. \textit{Observation:} GPT-5/Kimi remain top-two; Llama/Base Qwen stay \#8/\#9; HumorGen-7B remains top-tier (\#4).}
    \label{fig:htb_swiss_3rr_qwen}
\end{figure}

\begin{table}[H]
\centering
\small
\begin{tabular}{llcccc}
\toprule
\textbf{Rank} & \textbf{Model} & \textbf{BT Rating} & \textbf{Stable Elo} & \textbf{95\% CI} & \textbf{Win Rate} \\
\midrule
1 & GPT-5 & 1204.0 & 1207.1 & $[1182.2, 1228.7]$ & 72.1\% \\
2 & Kimi K2 & 1178.6 & 1182.5 & $[1158.3, 1199.0]$ & 69.9\% \\
3 & GPT OSS 120B & 1071.1 & 1064.0 & $[1045.8, 1100.2]$ & 55.1\% \\
4 & HumorGen-7B & 1071.1 & 1081.3 & $[1043.8, 1098.1]$ & 64.4\% \\
5 & Claude 3.5 Haiku & 1052.0 & 1053.2 & $[1034.1, 1078.3]$ & 57.6\% \\
6 & Gemini 2.5 Pro & 993.9 & 1004.3 & $[975.0, 1016.3]$ & 37.4\% \\
7 & Qwen 3 32B & 986.1 & 984.4 & $[962.4, 1008.9]$ & 34.7\% \\
8 & Llama 3.3 70B & 774.4 & 760.6 & $[749.1, 797.7]$ & 30.8\% \\
9 & Qwen 2.5 7B Instruct & 668.9 & 662.6 & $[634.9, 699.7]$ & 18.5\% \\
\bottomrule
\end{tabular}
\caption{\textbf{HTB, Swiss 3RR, Qwen judge.} Cross-judge HTB $\tau = 0.889$; anchor ranks \#1/\#8/\#9 stable vs.\ Qwen Full RR.}
\label{tab:htb_swiss_3rr_leaderboard_qwen}
\end{table}

\paragraph{Swiss 2RR ($\sim$22\% budget).}

\begin{figure}[htbp]
    \centering
    \includegraphics[width=0.72\linewidth]{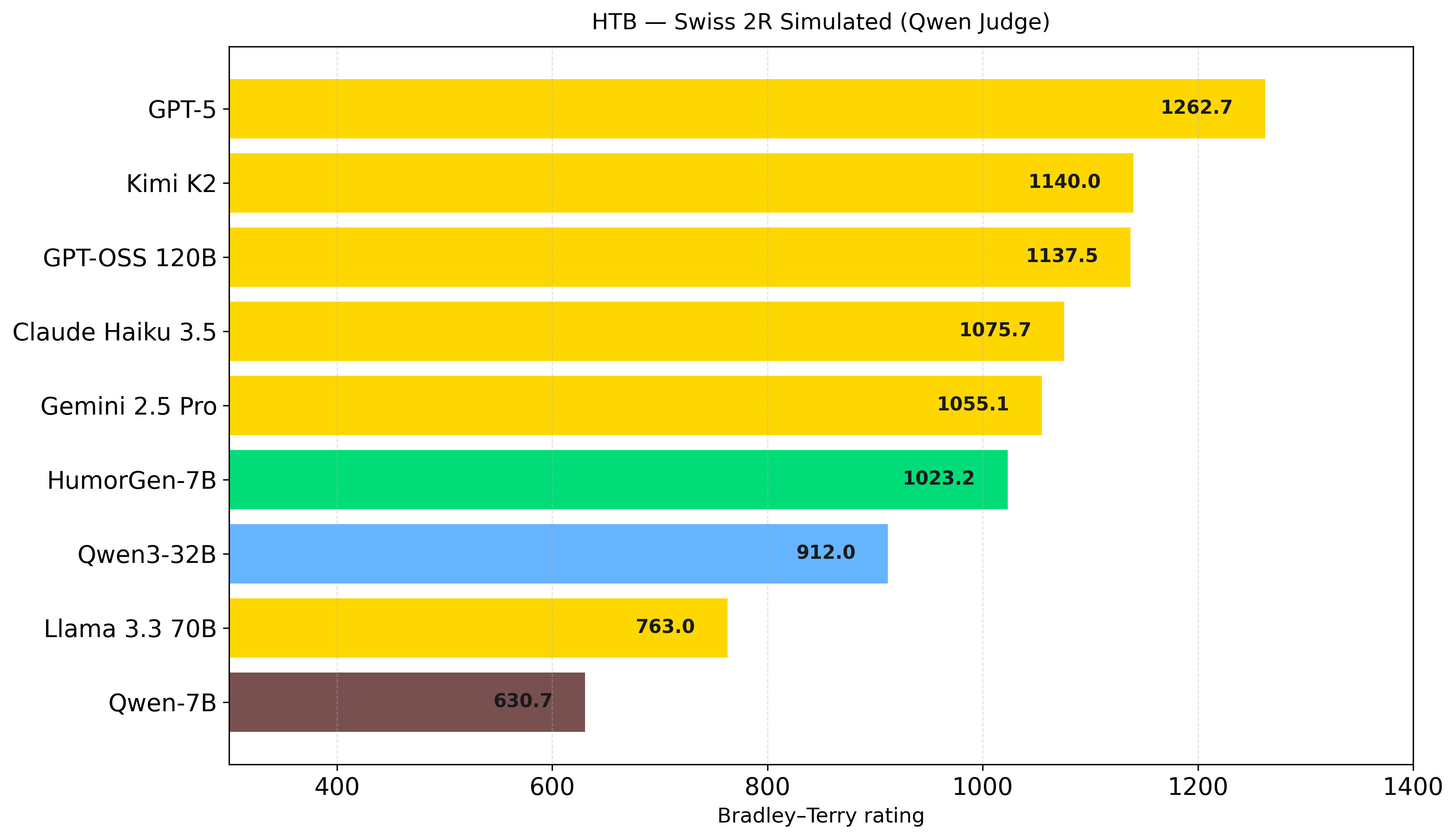}
    \vspace{0.4em}
    \includegraphics[width=0.72\linewidth]{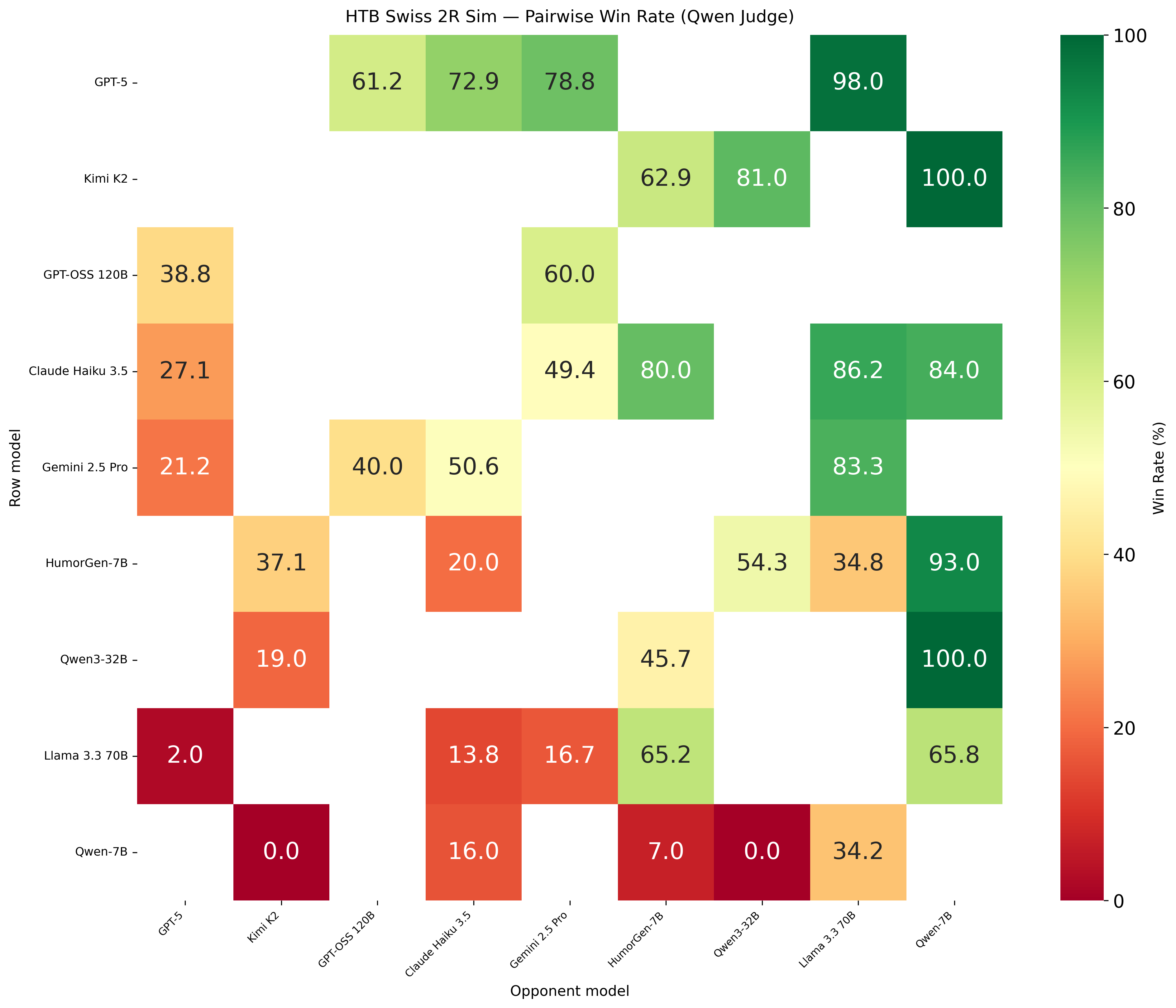}
    \caption{\textbf{HTB, Swiss 2RR, Qwen judge} (Table~\ref{tab:htb_swiss_2rr_leaderboard_qwen}). \textbf{Top:} BT leaderboard from 3,200 judged pairs. \textbf{Bottom:} Win-rate heatmap. \textit{Observation:} minimum-budget HTB run preserves top/bottom anchors but increases mid-tier volatility, consistent with the Llama judge Swiss 2RR pattern.}
    \label{fig:htb_swiss_2rr_qwen}
\end{figure}

\begin{table}[H]
\centering
\small
\begin{tabular}{llcccc}
\toprule
\textbf{Rank} & \textbf{Model} & \textbf{BT Rating} & \textbf{Stable Elo} & \textbf{95\% CI} & \textbf{Win Rate} \\
\midrule
1 & GPT-5 & 1262.7 & 1239.4 & $[1221.6, 1315.4]$ & 76.1\% \\
2 & Kimi K2 & 1140.0 & 1166.5 & $[1077.3, 1196.3]$ & 74.0\% \\
3 & GPT OSS 120B & 1137.6 & 1098.8 & $[1092.1, 1196.4]$ & 55.5\% \\
4 & Claude 3.5 Haiku & 1075.7 & 1057.5 & $[1035.0, 1113.9]$ & 62.7\% \\
5 & Gemini 2.5 Pro & 1055.1 & 1045.2 & $[1016.5, 1106.4]$ & 32.0\% \\
6 & HumorGen-7B & 1023.2 & 1057.8 & $[972.7, 1061.8]$ & 66.4\% \\
7 & Qwen 3 32B & 912.0 & 948.1 & $[850.6, 959.9]$ & 23.9\% \\
8 & Llama 3.3 70B & 763.0 & 746.8 & $[731.4, 797.7]$ & 35.5\% \\
9 & Qwen 2.5 7B Instruct & 630.7 & 639.9 & $[587.0, 668.7]$ & 18.2\% \\
\bottomrule
\end{tabular}
\caption{\textbf{HTB, Swiss 2RR, Qwen judge.} Anchor ranks \#1/\#8/\#9 stable; HumorGen-7B 4$\rightarrow$6 vs.\ Qwen Full RR. Cross-judge HTB $\tau = 1.000$ (Table~\ref{tab:appendix_budget_tau}).}
\label{tab:htb_swiss_2rr_leaderboard_qwen}
\end{table}
\FloatBarrier

\newpage
\section{Adaptive Swiss Pairing Algorithm}
\label{sec:appendix_asp_algorithm}

Algorithm~\ref{alg:swiss} specifies pair scheduling only; final leaderboard ratings are computed separately using global Bradley--Terry MLE on the collected match graph. In line with the main-text definition of \textit{under-sampled} pairs, the coverage count in line 901 is the observed duel count for $(i,j)$ in $G$; ASP prefers partners with the lowest count among near-equal-strength candidates.

\begin{algorithm}[H]
\caption{Adaptive Swiss Pairing for Tournament Evaluation}
\label{alg:swiss}
\begin{algorithmic}[1]
\Require Set of models $\mathcal{M}$, prompt set $\mathcal{H}$, max comparisons $C_{max}$, temporary pairing scores $S$
\State Initialize match-history multigraph $G$ with per-prompt pair counts $c_{ijh} \gets 0$
\While{total recorded matches in $G < C_{max}$}
    \State Sort $\mathcal{M}$ descending by temporary pairing scores $S$
    \State Initialize unmatched subset $U \gets \mathcal{M}$
    \State Initialize round pairings $Q \gets \emptyset$
    \While{$|U| \ge 2$}
        \State $i \gets$ highest-rated model in $U$
        \State Find $j \in U \setminus \{i\}$ minimizing $|S_i - S_j|$ with lowest coverage count in $G$
        \If{valid match $j$ exists}
            \State $Q \gets Q \cup \{(i,j)\}$
            \State $U \gets U \setminus \{i,j\}$
        \Else
            \State $j \gets$ least-played feasible partner in $U \setminus \{i\}$ \Comment{progress fallback}
            \State $Q \gets Q \cup \{(i,j)\}$
            \State $U \gets U \setminus \{i,j\}$
        \EndIf
    \EndWhile
    \State Execute LLM-as-a-judge evaluations for all pairs in $Q$ on selected prompts
    \State Increment corresponding per-prompt counts in $G$
    \State Update temporary pairing scores $S$ using observed outcomes
\EndWhile
\State \textbf{return} Global match history graph $G$ \Comment{final ratings computed later via BT MLE}
\end{algorithmic}
\end{algorithm}

\section{Synthetic ASP Scaling Stress Test}
\label{sec:appendix_asp_scaling}

The main paper validates ASP on a real humor tournament at $K{=}9$ (Full~RR leaderboards; Swiss~2RR/3RR budget ablation). Collecting jokes from $10^{4}$--$10^{5}$ generators is impractical, so we additionally stress-test the \emph{scheduler and rank estimator} in a synthetic Bradley--Terry (BT) world that requires no jokes and no LLM calls. This appendix supports ASP as a scaling contribution; it does \emph{not} claim a humor-quality ranking over $10^{5}$ real models.

\paragraph{Setup:}
Each of $K$ contestants has a known latent strength on an evenly spaced ladder. Every scheduled duel is drawn from the logistic BT model $\mathbb{P}(i\succ j)=\sigma(\theta_i-\theta_j)$. ASP builds the match graph under two budgets: (i)~fixed Swiss~3RR with $R{=}3$ (at $K{=}9$: $C_{\max}{=}12$, i.e.\ $\sim$33\% of Full~RR, matching Section~\ref{sec:budget_ablation}), which is $\mathcal{O}(K)$ pairs/prompt; (ii)~optional denser $R{=}\lceil\log_2 K\rceil$, which is $\mathcal{O}(K\log K)$ pairs/prompt. Rankings are recovered by BT~MLE for $K\le500$ and Elo for larger $K$. We report graph connectivity, Spearman~$\rho$, and Kendall~$\tau$ versus the latent order. Exact ASP (Algorithm~\ref{alg:swiss}) is used for $K\le1000$; a windowed coverage-aware approximation is used beyond that for computational tractability.

\paragraph{Results:}
Table~\ref{tab:asp_synthetic_scaling} summarizes the stress test. Under both budgets the union match graph remains connected through $K{=}10^{5}$. Swiss~3RR already recovers the latent order with high fidelity at the paper's operating point ($K{=}9$: $\rho{=}0.983$, $\tau{=}0.944$) and remains informative at extreme scale ($K{=}10^{5}$: $\rho{=}0.906$). The optional $\mathcal{O}(K\log K)$ schedule further improves large-$K$ recovery (e.g., $\rho{=}0.972$ at $K{=}10^{5}$) while still using orders of magnitude fewer comparisons than Full~RR ($\binom{10^{5}}{2}\!\approx\!5{\times}10^{9}$ pairs/prompt vs.\ $8.5{\times}10^{5}$ under $R{=}17$).

\begin{table}[t]
\centering
\small
\begin{tabular}{rrlrrcc}
\toprule
$K$ & $R$ & Mode & Pairs/prompt & Budget \% & $\rho$ & $\tau$ \\
\midrule
9 & 3 & exact & 12 & 33.3 & 0.983 & 0.944 \\
9 & 4 & exact & 18 & 50.0 & 0.983 & 0.944 \\
100 & 3 & exact & 150 & 3.0 & 0.998 & 0.971 \\
1{,}000 & 3 & exact & 1{,}500 & 0.30 & 0.987 & 0.904 \\
1{,}000 & 10 & exact & 5{,}000 & 1.0 & 0.997 & 0.955 \\
10{,}000 & 3 & windowed & 15{,}000 & 0.03 & 0.948 & 0.802 \\
10{,}000 & 14 & windowed & 70{,}000 & 0.14 & 0.990 & 0.911 \\
50{,}000 & 3 & windowed & 75{,}000 & 0.01 & 0.915 & 0.747 \\
50{,}000 & 16 & windowed & 400{,}000 & 0.03 & 0.962 & 0.829 \\
100{,}000 & 3 & windowed & 150{,}000 & 0.003 & 0.906 & 0.732 \\
100{,}000 & 17 & windowed & 850{,}000 & 0.017 & 0.972 & 0.851 \\
\bottomrule
\end{tabular}
\caption{Synthetic ASP scaling (no jokes/LLM). Outcomes from a latent BT ladder; rankings recovered by BT-MLE ($K\le500$) or Elo ($K>500$). At $K{=}9$, $R{=}3$ uses the paper Swiss~3RR budget of 12 pairs/prompt ($\sim$33\% of Full~RR). All listed graphs are connected. $R{=}3$ is $\mathcal{O}(K)$; $R{=}\lceil\log_2 K\rceil$ is $\mathcal{O}(K\log K)$.}
\label{tab:asp_synthetic_scaling}
\end{table}

\paragraph{Interpretation and scope:}
Swiss~3RR is the practical cheap mode ($\mathcal{O}(K)$), not an $\mathcal{O}(K\log K)$ claim. The $\mathcal{O}(K\log K)$ column is an optional denser schedule for very large pools. Real humor evidence for ASP remains the $K{=}9$ Full~RR/Swiss ablation; Table~\ref{tab:asp_synthetic_scaling} shows that the same budgeting rules continue to yield usable comparison graphs and recoverable rankings as $K$ grows under a controlled preference model.

\newpage
\section{HumorRank Leaderboard Performance with Qwen 2.5 72B LLM Judge}
\label{sec:appendix_qwen_leaderboard}

\begin{figure}[htbp]
    \centering
    \includegraphics[width=0.72\linewidth]{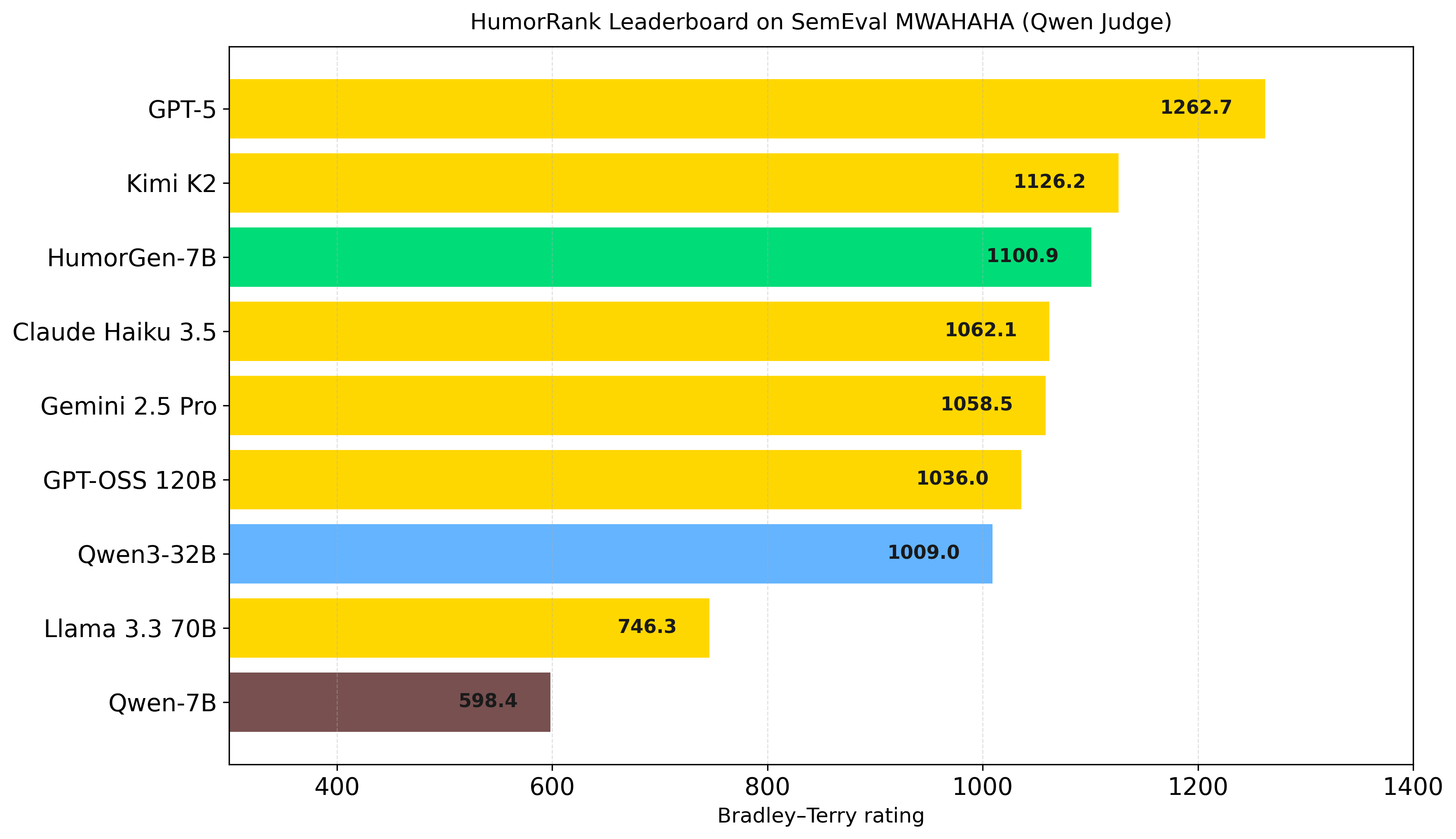}
    \vspace{0.4em}
    \includegraphics[width=0.72\linewidth]{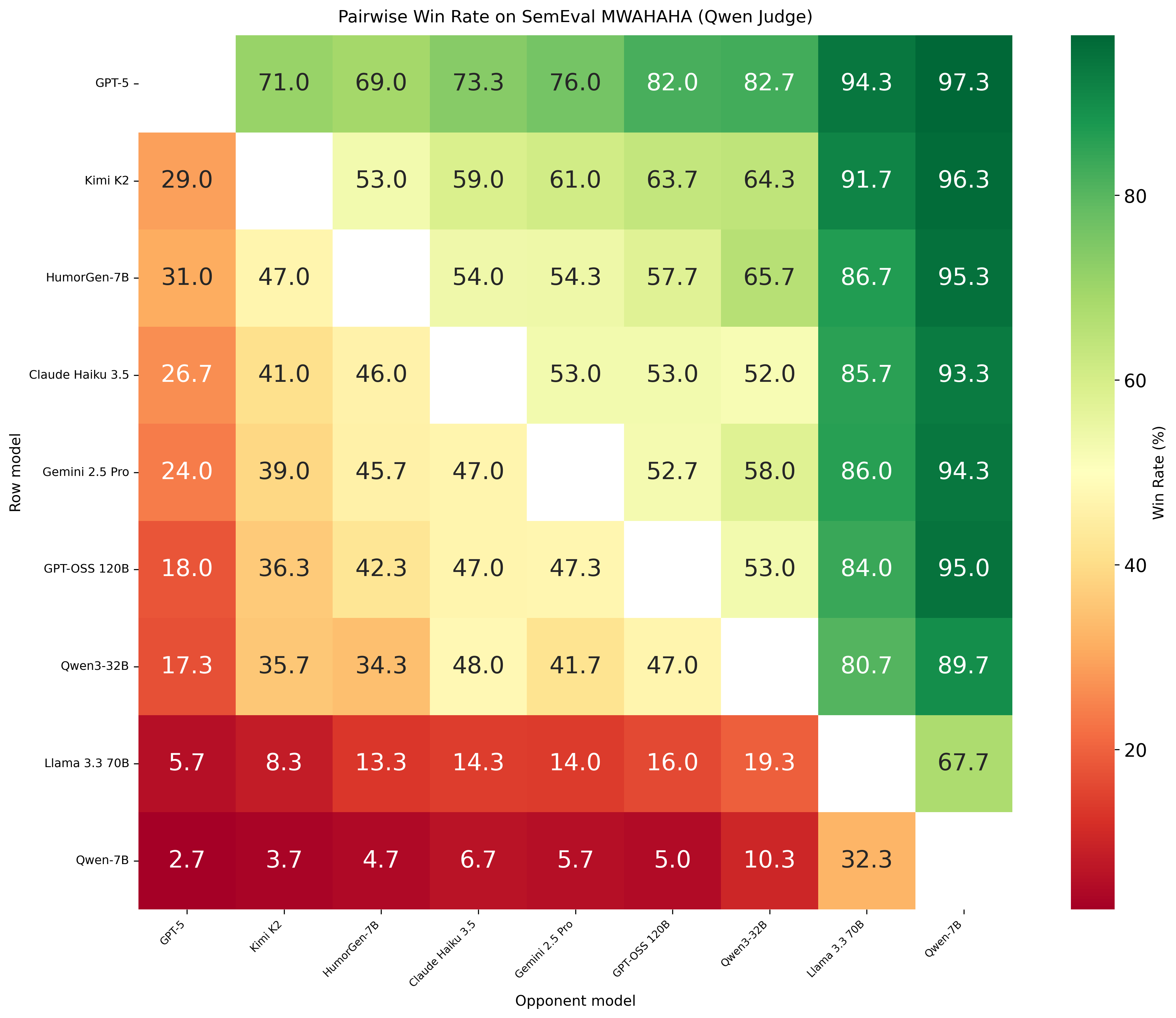}
    \caption{\textbf{SemEval, Full RR, Qwen judge} (Table~\ref{tab:leaderboard_qwen}; Kendall $\tau = 0.889$ vs.\ SemEval panel of Table~\ref{tab:leaderboard}). \textbf{Top:} Bradley--Terry leaderboard with 95\% confidence intervals (10,800 judgments). \textbf{Bottom:} Pairwise win-rate heatmap. \textit{Observation:} GPT-5 and Kimi K2 remain ranks 1--2; Llama 3.3 70B and Qwen 2.5 7B Instruct remain ranks 8--9; HumorGen-7B moves 4$\rightarrow$3 and Gemini 2.5 Pro 3$\rightarrow$5 vs.\ the Llama judge, consistent with modest mid-tier reordering at $\tau = 0.889$.}
    \label{fig:leaderboard_qwen}
\end{figure}

To validate the stability of our primary SemEval leaderboard (Llama judge), we replicate the same 10,800 SemEval duels with Qwen 2.5 72B Instruct as a secondary LLM judge. Figure~\ref{fig:leaderboard_qwen} and Table~\ref{tab:leaderboard_qwen} present the resulting Bradley--Terry leaderboard ($\tau = 0.889$ vs.\ SemEval panel of Table~\ref{tab:leaderboard}).

\begin{table}[H]
\centering
\small
\begin{tabular}{llcccc}
\toprule
\textbf{Rank} & \textbf{Model} & \textbf{BT Rating} & \textbf{Stable Elo} & \textbf{95\% CI} & \textbf{Win Rate} \\
\midrule
1 & GPT-5 & 1262.7 & 1267.7 & $[1246.2, 1279.8]$ & 80.7\% \\
2 & Kimi-K2 & 1126.2 & 1135.8 & $[1113.8, 1140.2]$ & 64.8\% \\
3 & HumorGen-7B\footnotemark[1] & 1100.9 & 1092.1 & $[1088.5, 1114.9]$ & 61.5\% \\
4 & Claude 3.5 Haiku & 1062.1 & 1062.4 & $[1049.4, 1074.9]$ & 56.3\% \\
5 & Gemini 2.5 Pro & 1058.5 & 1068.4 & $[1044.0, 1070.4]$ & 55.8\% \\
6 & GPT OSS 120B & 1036.0 & 1036.2 & $[1023.5, 1054.1]$ & 52.9\% \\
7 & Qwen 3 32B & 1009.0 & 997.1 & $[993.8, 1026.2]$ & 49.3\% \\
\midrule
8 & Llama 3.3 70B & 746.3 & 740.7 & $[725.1, 761.7]$ & 19.8\% \\
9 & Qwen 2.5 7B Instruct & 598.4 & 599.6 & $[578.4, 624.4]$ & 8.9\% \\
\bottomrule
\end{tabular}
\caption{\textbf{SemEval, Full RR, Qwen judge} (10,800 judgments). BT ratings are primary; Stable Elo is a sequence-robust audit metric (Appendix~\ref{sec:appendix_elo_stability}). \textit{Observation:} Kendall $\tau = 0.889$ vs.\ SemEval panel of Table~\ref{tab:leaderboard}; anchor ranks \#1/\#8/\#9 match the Llama judge; HumorGen-7B ranks 3rd (vs.\ 4th under the Llama judge).}
\label{tab:leaderboard_qwen}
\end{table}
\footnotetext[1]{Referred to HumorGen-7B as HumorGen SFT 7B in plots and figures.}

\newpage
\newpage
\section{Llama Judge Sample Decisions}
\label{sec:appendix_llama_judge}

Below are four representative Llama judge (Llama 3.3 70B Instruct) decisions from SemEval match logs, selected to span outcome types (frontier vs.\ mid-tier, specialist vs.\ baseline, and tie-adjacent splits). Each pair shows the preferred joke (green) against the rejected entry (red), together with the LLM judge's reasoning and approximate Elo updates from the online pairing log (audit only; final leaderboard ratings use global Bradley--Terry MLE).

\definecolor{winnercolor}{RGB}{22, 120, 80}
\definecolor{losercolor}{RGB}{180, 50, 50}
\definecolor{deliverycolor}{RGB}{30, 90, 180}
\definecolor{winnerbg}{RGB}{236, 253, 245}
\definecolor{loserbg}{RGB}{255, 241, 241}
\definecolor{judgemetabg}{RGB}{30, 41, 59}

\newcommand{\featureline}[3]{%
\noindent
\begin{tcolorbox}[
  enhanced,
  colback=#1!12,
  colframe=#1!25,
  arc=3pt, boxrule=0.4pt,
  left=7pt, right=7pt, top=2pt, bottom=2pt,
  before skip=2pt, after skip=2pt,
]
\colorbox{#1}{\color{white}\bfseries\sffamily\scriptsize\strut~#2~}%
\hspace{4pt}{\sffamily\scriptsize\color{black!70} #3}
\end{tcolorbox}%
}

\newcommand{\winnerbox}[2]{%
\noindent
\begin{tcolorbox}[
  enhanced,
  colback=winnerbg,
  colframe=winnercolor!50,
  leftrule=5pt, rightrule=0.4pt, toprule=0.4pt, bottomrule=0.4pt,
  arc=3pt,
  left=8pt, right=8pt, top=5pt, bottom=5pt,
  after skip=2pt, before skip=5pt,
  attach boxed title to top left={yshift=-2pt, xshift=8pt},
  boxed title style={
    colback=winnercolor, colframe=winnercolor,
    arc=2pt, boxrule=0pt,
    left=5pt, right=5pt, top=2pt, bottom=2pt
  },
  title={\color{white}\bfseries\sffamily\scriptsize #1
         \normalfont\sffamily\scriptsize\color{white!90}~$\cdot$~Winner~{\large\checkmark}}
]
{\sffamily\small #2}
\end{tcolorbox}%
}

\newcommand{\loserbox}[2]{%
\noindent
\begin{tcolorbox}[
  enhanced,
  colback=loserbg,
  colframe=losercolor!50,
  leftrule=5pt, rightrule=0.4pt, toprule=0.4pt, bottomrule=0.4pt,
  arc=3pt,
  left=8pt, right=8pt, top=5pt, bottom=5pt,
  after skip=6pt, before skip=2pt,
  attach boxed title to top left={yshift=-2pt, xshift=8pt},
  boxed title style={
    colback=losercolor, colframe=losercolor,
    arc=2pt, boxrule=0pt,
    left=5pt, right=5pt, top=2pt, bottom=2pt
  },
  title={\color{white}\bfseries\sffamily\scriptsize #1
         \normalfont\sffamily\scriptsize\color{white!90}~$\cdot$~Loser~{\large$\times$}}
]
{\sffamily\small\color{black!80} #2}
\end{tcolorbox}%
}

\noindent
\begin{tcolorbox}[
  enhanced, colback=judgemetabg!8, colframe=judgemetabg!20,
  arc=4pt, boxrule=0.4pt,
  left=8pt, right=8pt, top=4pt, bottom=4pt,
  after skip=10pt
]
\sffamily\scriptsize
\textcolor{winnercolor}{\rule{9pt}{7pt}}\;
\textbf{\color{winnercolor} Winning Features}
\quad
\textcolor{deliverycolor}{\rule{9pt}{7pt}}\;
\textbf{\color{deliverycolor} Delivery Features}
\quad
\textcolor{losercolor}{\rule{9pt}{7pt}}\;
\textbf{\color{losercolor} Loser Features}
\end{tcolorbox}

\noindent
\begin{tcolorbox}[
  enhanced, colback=judgemetabg, colframe=judgemetabg,
  arc=3pt, boxrule=0pt,
  left=8pt, right=8pt, top=3pt, bottom=3pt,
  after skip=4pt, before skip=8pt,
]
{\color{white}\sffamily\scriptsize
  \textbf{Decision \#1} \quad \texttt{en\_2051}
  \quad\textbf{·}\quad Tie: No
  \quad\textbf{·}\quad Confidence: \textsc{Medium}
  \quad\textbf{·}\quad Score: 1.0
  \quad\textbf{·}\quad
  ELO$_\Delta$: {\color{green!60!white}$+$7.66} ~/~ {\color{red!60!white}$-$7.66}
}
\end{tcolorbox}

\noindent{\sffamily\small\itshape\color{black!55}
  Headline: `We did it!': Countries celebrate World Cup qualification
}\vspace{2pt}

\noindent{\sffamily\footnotesize\color{black!55}
  Reasoning: Joke~A has a more relatable and unexpected twist: the concept of
  ``not washing lucky socks'' as a claimed contribution to the team's success
  makes it more humorous and engaging.
}\vspace{3pt}

\featureline{winnercolor}{Winning Features}{incongruity $\cdot$ absurdity $\cdot$ surprise $\cdot$ irony}
\featureline{deliverycolor}{Delivery Features}{conciseness $\cdot$ punchline\_positioning $\cdot$ framing\_commitment}
\featureline{losercolor}{Loser Features}{clich\'e $\cdot$ weak\_punchline}

\winnerbox{Gemini-2.5-Pro}{%
  After my team qualified for the World Cup, I ran into the street yelling,
  ``WE DID IT!'' My neighbor looked at me and said, ``What exactly did \textit{you}
  do?'' I told him, ``I successfully avoided jinxing them by not washing my lucky
  socks for six months. You're welcome.''
}
\loserbox{Claude-3.5-Haiku}{%
  Why did the soccer player bring a ladder to celebrate World Cup qualification?
  Because they wanted to take their excitement to the next level\ldots\ literally,
  they were already on cloud nine!
}

\noindent
\begin{tcolorbox}[
  enhanced, colback=judgemetabg, colframe=judgemetabg,
  arc=3pt, boxrule=0pt,
  left=8pt, right=8pt, top=3pt, bottom=3pt,
  after skip=4pt, before skip=8pt,
]
{\color{white}\sffamily\scriptsize
  \textbf{Decision \#2} \quad \texttt{en\_2021}
  \quad\textbf{·}\quad Tie: No
  \quad\textbf{·}\quad Confidence: \textsc{Medium}
  \quad\textbf{·}\quad Score: 1.0
  \quad\textbf{·}\quad
  ELO$_\Delta$: {\color{green!60!white}$+$7.14} ~/~ {\color{red!60!white}$-$7.14}
}
\end{tcolorbox}

\noindent{\sffamily\small\itshape\color{black!55}
  Headline: `Mind-blowing' population of feral goldfish in Ottawa pond to be euthanized
}\vspace{2pt}

\noindent{\sffamily\footnotesize\color{black!55}
  Reasoning: Joke~A uses a clever play on words with `mind-blowing' and creates a
  vivid, absurd image; Joke~B relies on a more straightforward and less surprising
  punchline.
}\vspace{3pt}

\featureline{winnercolor}{Winning Features}{wordplay $\cdot$ absurdity $\cdot$ incongruity}
\featureline{deliverycolor}{Delivery Features}{conciseness $\cdot$ punchline\_positioning $\cdot$ framing\_commitment}
\featureline{losercolor}{Loser Features}{clich\'e $\cdot$ weak\_punchline}

\winnerbox{HumorGen-SFT-7B}{%
  So, they're going to euthanize these `mind-blowing' feral goldfish in Ottawa's pond?
  I hope they have anesthetic for fish because if you don't do it right, those goldfish
  might just get their last swim as they float up to heaven saying,
  ``Mind-blowing, indeed!''
}
\loserbox{GPT-OSS-120B}{%
  Ottawa's pond got so many feral goldfish that the city had to ``euthanize''
  them; turns out they were just trying to pay the water bill with their own scales.
}

\noindent
\begin{tcolorbox}[
  enhanced, colback=judgemetabg, colframe=judgemetabg,
  arc=3pt, boxrule=0pt,
  left=8pt, right=8pt, top=3pt, bottom=3pt,
  after skip=4pt, before skip=8pt,
]
{\color{white}\sffamily\scriptsize
  \textbf{Decision \#3} \quad \texttt{en\_2034}
  \quad\textbf{·}\quad Tie: No
  \quad\textbf{·}\quad Confidence: \textsc{Medium}
  \quad\textbf{·}\quad Score: 1.0
  \quad\textbf{·}\quad
  ELO$_\Delta$: {\color{green!60!white}$+$7.63} ~/~ {\color{red!60!white}$-$7.63}
}
\end{tcolorbox}

\noindent{\sffamily\small\itshape\color{black!55}
  Headline: Three-quarters of parents let children miss school for `duvet day'
}\vspace{2pt}

\noindent{\sffamily\footnotesize\color{black!55}
  Reasoning: Joke~B has a more complex and layered structure: the fraction wordplay
  (``three-quarters is exactly the blanket-to-face ratio'') adds an extra dimension
  of sarcasm and irony over Joke~A's simpler setup.
}\vspace{3pt}

\featureline{winnercolor}{Winning Features}{incongruity $\cdot$ wordplay $\cdot$ absurdity $\cdot$ irony $\cdot$ sarcasm}
\featureline{deliverycolor}{Delivery Features}{timing $\cdot$ conciseness $\cdot$ punchline\_positioning $\cdot$ framing\_commitment}
\featureline{losercolor}{Loser Features}{weak\_punchline}

\winnerbox{GPT-5}{%
  Three-quarters of parents let kids miss school for ``duvet day''; teachers are
  fuming: ``How are we supposed to cover the syllabus when everyone's already under
  the covers?'' On the bright side, kids finally understand fractions: three-quarters
  is exactly the blanket-to-face ratio required to breathe.
}
\loserbox{Qwen 2.5 7B Instruct}{%
  Why did the kid stay home from school? Because when asked if he wanted to go,
  he replied, ``I'd rather be sick three-quarters of the time!''
}

\noindent
\begin{tcolorbox}[
  enhanced, colback=judgemetabg, colframe=judgemetabg,
  arc=3pt, boxrule=0pt,
  left=8pt, right=8pt, top=3pt, bottom=3pt,
  after skip=4pt, before skip=8pt,
]
{\color{white}\sffamily\scriptsize
  \textbf{Decision \#4} \quad \texttt{en\_2115}
  \quad\textbf{·}\quad Tie: No
  \quad\textbf{·}\quad Confidence: \textsc{Medium}
  \quad\textbf{·}\quad Score: 1.0
  \quad\textbf{·}\quad
  ELO$_\Delta$: {\color{green!60!white}$+$7.66} ~/~ {\color{red!60!white}$-$7.66}
}
\end{tcolorbox}

\noindent{\sffamily\small\itshape\color{black!55}
  Headline: How to design an aesthetic home office that boosts productivity
}\vspace{2pt}

\noindent{\sffamily\footnotesize\color{black!55}
  Reasoning: Both jokes rely on irony and sarcasm, but Joke~A's darker,
  more unexpected twist, ``the exact shade of my laptop screen at 3\,a.m. (black
  with a faint glow of regret)'', is more surprising and absurd.
}\vspace{3pt}

\featureline{winnercolor}{Winning Features}{incongruity $\cdot$ absurdity $\cdot$ surprise $\cdot$ irony $\cdot$ sarcasm}
\featureline{deliverycolor}{Delivery Features}{conciseness $\cdot$ deadpan $\cdot$ punchline\_positioning}
\featureline{losercolor}{Loser Features}{clich\'e $\cdot$ weak\_punchline}

\winnerbox{Kimi-K2}{%
  My productivity skyrocketed after I designed my aesthetic home office: I painted
  the walls the exact shade of my laptop screen at 3\,a.m.: black with a faint
  glow of regret.
}
\loserbox{Gemini-2.5-Pro}{%
  I spent all weekend creating the perfect aesthetic home office to boost my
  productivity. It worked! I am now incredibly productive at taking pictures of
  my aesthetic home office for Instagram.
}

\vspace{0.5em}
\captionof{figure}{Four representative LLaMA judge decisions. Winner~{\checkmark}
(green) and Loser~{$\times$} (red) are labelled directly on each joke box.
Feature rows indicate winning humor traits (green), delivery strengths (blue),
and loser weaknesses (red). ELO deltas are approximated from the evaluation log.}
\label{fig:llama_judge_decisions}
\newpage

\section{Hyperparameter Configurations}
\label{sec:appendix_hyperparameters}

Standardized hyperparameters across the HumorRank tournament are detailed in Table~\ref{tab:hyperparams_gen} (candidate generation), Table~\ref{tab:hyperparams_judge} (LLM pairwise judging), and Table~\ref{tab:hyperparams_tournament} (Bradley--Terry and Elo audit settings).
Primary Llama-judge Full RR pipelines (SemEval and HTB; 25{,}200 pairwise calls per LLM judge across both benchmarks) used NVIDIA H100 (80GB) GPU inference and/or hosted LLM APIs; Qwen-judge replication required additional inference budget.
Offline rating replay from the anonymized supplementary match logs requires only CPU Python\,$\ge$\,3.10 with NumPy (and the \texttt{krippendorff} package for human-eval $\alpha$).

\begin{table}[H]
\centering
\small
\begin{tabular}{lp{0.6\linewidth}}
\toprule
\textbf{Parameter} & \textbf{Value} \\
\midrule
Temperature & 0.7 \\
Top-$p$ & 0.9 \\
Max New Tokens & 256 \\
System Prompt & "You are a joke generator. Given a headline or topic, generate a funny joke. Output ONLY the joke text. No thinking tags, no reasoning, no explanation, no extra words." \\
\bottomrule
\end{tabular}
\caption{Hyperparameters for candidate humor generation across all local and API-based models.}
\label{tab:hyperparams_gen}
\end{table}

\begin{table}[H]
\centering
\small
\begin{tabular}{lp{0.6\linewidth}}
\toprule
\textbf{Parameter} & \textbf{Value} \\
\midrule
Primary Judge & Llama 3.3 70B Instruct \\
Ablation Judge & Qwen 2.5 72B Instruct \\
Temperature & 0.1 \\
Max New Tokens & 512 \\
Max Retries & 3 \\
Backoff Base & 2.0 \\
Retry Cap & 4.0 \\
\bottomrule
\end{tabular}
\caption{Hyperparameters for LLM-as-a-Judge adjudication.}
\label{tab:hyperparams_judge}
\end{table}

\begin{table}[H]
\centering
\small
\begin{tabular}{ll}
\toprule
\textbf{Parameter} & \textbf{Value} \\
\midrule
Initial Elo Rating & 1000.0 \\
$K$-factor & 32 \\
Min Rounds per Model & 2 \\
Max Rounds per Model & 3 \\
Stable Elo Shuffles & 10 \\
BT Convergence ($\epsilon$) & $10^{-6}$ \\
Bootstrap Iterations & 200 \\
Bootstrap Resampling Seed & 42 \\
\bottomrule
\end{tabular}
\caption{Tournament configuration and Bradley--Terry global MLE parameters.}
\label{tab:hyperparams_tournament}
\end{table}

\newpage
\section{Qualitative Examples and Feature Reasoning}
\label{sec:appendix_qualitative}

The main paper focuses on aggregate feature patterns (Figures~\ref{fig:heatmap_llama_humor}--\ref{fig:heatmap_llama_delivery}). Appendix~\ref{sec:appendix_llama_judge} provides representative LLM judge rationales drawn from SemEval match logs. Qwen judge feature distributions are below.

\begin{figure}[htbp]
    \centering
    \includegraphics[width=0.72\linewidth]{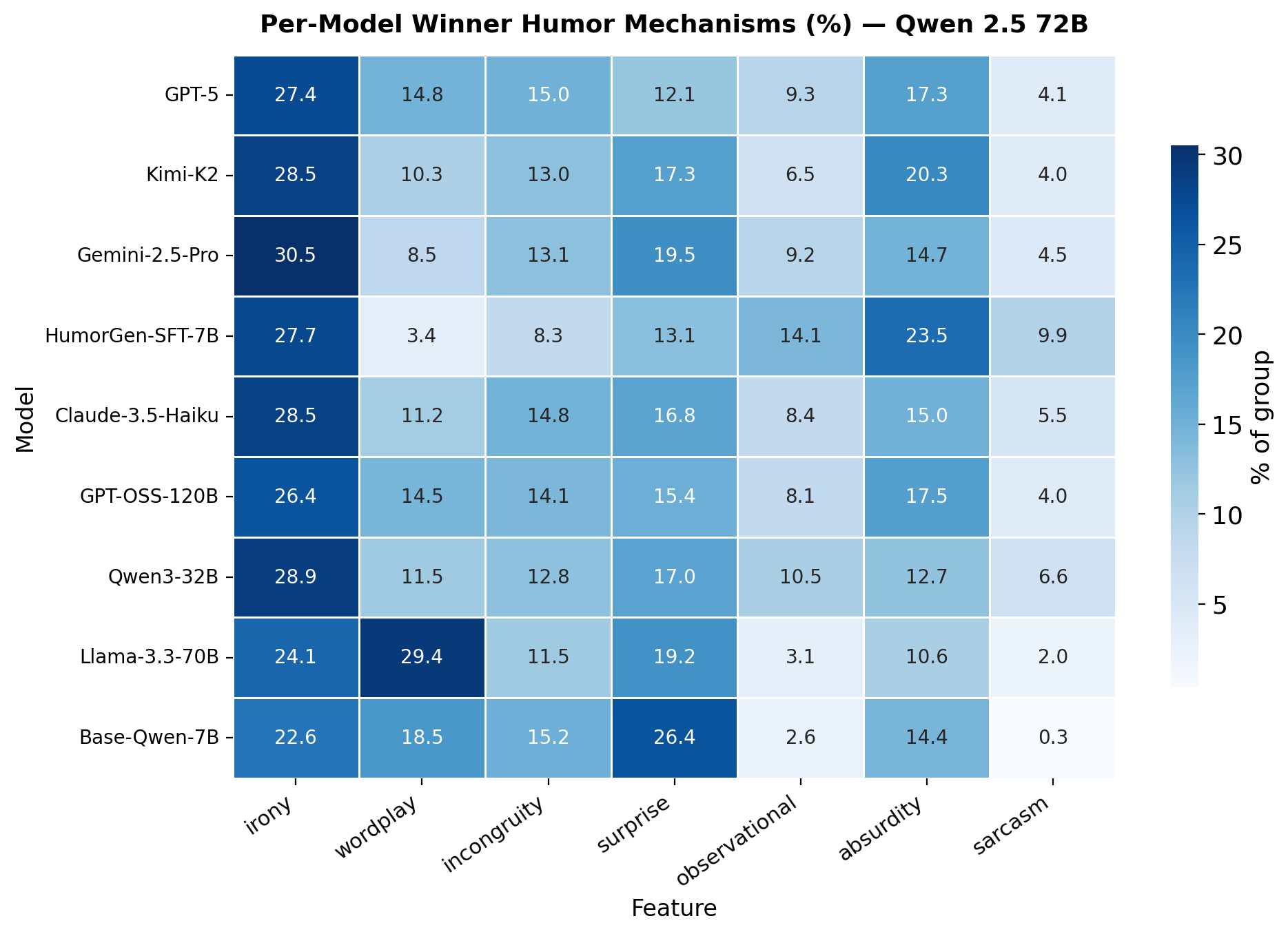}
    \vspace{0.4em}
    \includegraphics[width=0.72\linewidth]{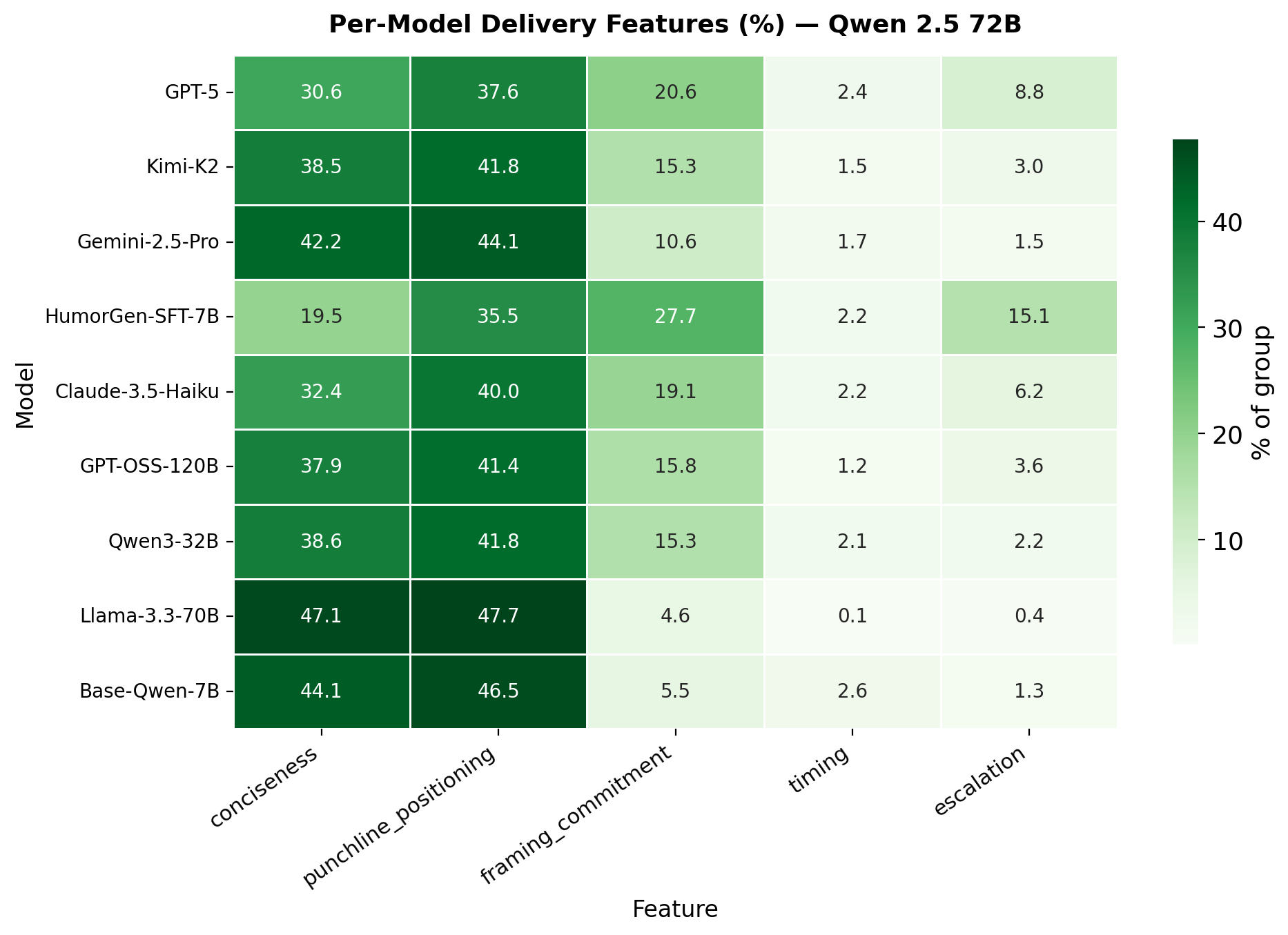}
    \caption{Qwen judge: per-model winning feature distributions. \textbf{Top:} Humor mechanisms (\% of wins). \textbf{Bottom:} Delivery features (\% of wins). Tags are co-emitted with duel outcomes in the structured LLM judge response. Rank patterns are consistent with the primary Llama judge (Figures~\ref{fig:heatmap_llama_humor} and~\ref{fig:heatmap_llama_delivery}).}
    \label{fig:heatmaps_qwen_appendix}
\end{figure}

\begin{figure}[htbp]
    \centering
    \includegraphics[width=0.72\linewidth]{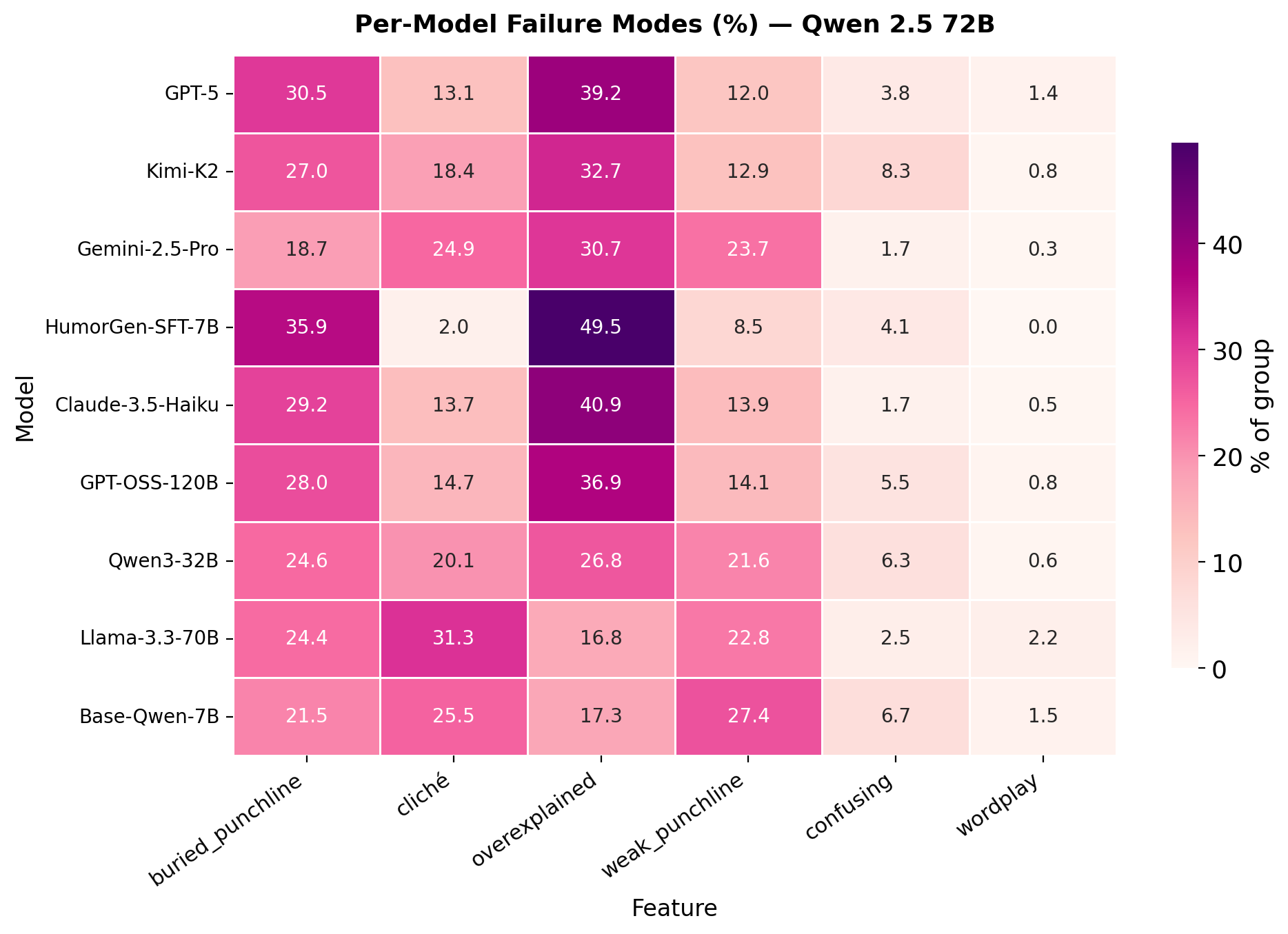}
    \caption{Qwen judge: per-model failure mode distributions (\% of losses). HumorGen-7B again shows markedly higher \textit{Overexplained} (49.5\%) rates than other models, consistent with the primary Llama judge (Figure~\ref{fig:heatmap_llama_loser}).}
    \label{fig:heatmap_qwen_loser}
\end{figure}

\subsection{Key Observations (Qwen vs.\ Llama LLM judges)}\label{sec:appendix_qualitative_observations}

Relative to the primary Llama judge (Figures~\ref{fig:heatmap_llama_humor} and~\ref{fig:heatmap_llama_delivery}), the Qwen judge preserves the same tier structure but shifts tag weights modestly on mid-tier models:

\begin{enumerate}
    \item \textbf{HumorGen-SFT-7B} shows the largest judge-specific gap: \textit{Absurdity} is 25.9\% under the Llama judge vs.\ 23.5\% under the Qwen judge, while \textit{Overexplained} loser tags remain elevated under both LLM judges (Figure~\ref{fig:heatmap_qwen_loser}).

    \item \textbf{Baseline open-weight models} (Base-Qwen-7B, Llama-3.3-70B) show near-identical mechanism and delivery profiles across judges, reinforcing that bottom-tier separation is judge-invariant.

    \item \textbf{Frontier models} (GPT-5, Kimi-K2) retain high \textit{Incongruity} and \textit{Conciseness} under both LLM judges; the Qwen judge assigns slightly higher \textit{Wordplay} shares to GPT-5 wins than the Llama judge does.
\end{enumerate}

\section{Human Evaluation Details}
\label{sec:appendix_human_eval}

\paragraph{Pair selection and $\alpha$ computation.}\label{sec:appendix_human_eval_selection}
The 90-pair evaluation set (Tables~\ref{tab:human_eval_design} and~\ref{tab:alpha_breakdown}) comprises curated funny-versus-funny blind comparisons over 75 unique headlines, stratified by comparison type as in Table~\ref{tab:human_eval_design}. Three blind human annotators (denoted H$_{1}$, H$_{2}$, and H$_{3}$) independently re-rated anonymized joke pairs; each vote is coded as a nominal winner-model label. Krippendorff's $\alpha$~\citep{krippendorff2011computing} is computed on the resulting annotator$\times$pair matrix, with incomplete overlap handled natively. In Table~\ref{tab:alpha_breakdown}, H$_{i}$ indexes human annotators, while \textit{Llama} and \textit{Qwen} denote the production LLM judges (Llama 3.3 70B and Qwen 2.5 72B Instruct).

\paragraph{Instructions to participants.}\label{sec:appendix_human_eval_instructions}
Annotators saw the headline, two anonymized jokes (Option A and Option B), and chose which was funnier or declared a tie. Model identities were hidden; left/right order was randomized per pair.

\paragraph{Participants.}\label{sec:appendix_human_eval_participants}
Three annotators were recruited by invitation (Master's students with native or near-native English proficiency and prior coursework or research exposure to humor and NLP). They rated the 90-pair evaluation set without payment. We index them as H$_{1}$--H$_{3}$ in Table~\ref{tab:alpha_breakdown}.

\paragraph{Inter-Annotator Reliability.}\label{sec:appendix_human_eval_reliability}
We use Krippendorff's Alpha ($\alpha$) for nominal data with multiple annotators and incomplete overlap:
\begin{equation}
\alpha = 1 - \frac{D_o}{D_e}
\end{equation}
where $D_o$ is observed disagreement and $D_e$ is expected chance disagreement. Cohort-level $\alpha$ values are reported in Table~\ref{tab:alpha_breakdown}. The Fisher exact test in \S\ref{sec:human_eval} compares human--human and human--Llama agreement on annotator dyad H$_2$+H$_3$ over the same judge-labeled pairs ($p = 0.808$).

\newpage
\section{Stable Elo Shuffle Audit}
\label{sec:appendix_elo_stability}


To validate the robustness of the derived Elo ratings against sequence-dependence (often referred to as ``late-winner bias'' in streamed continuous tournaments), HumorRank uses a Stable Elo variant grounded in order-independent aggregation~\citep{albers2001elo}. 

In standard sequential Elo implementations, a model $m$ updates its rating $R_m$ after a sequence of matches based on the standard iterative update rule:
\begin{equation}
    R_{m}^{(t+1)} = R_{m}^{(t)} + K \cdot (S - E_{m})
\end{equation}
where $t$ indexes the chronological order of the match. Consequently, a model earning a win at the end of the match history block gains an inherently outsized advantage over a model that earned an identical win early in the sequence.

To substantially reduce this temporal artifact, we strip the time dependencies by evaluating the match history $H$ across $N$ independently shuffled topological permutations. The stable terminal rating $\bar{R}_m$ for each model $m$ is defined as the arithmetic mean across all sequences:
\begin{equation}
    \bar{R}_m = \frac{1}{N} \sum_{k=1}^N R_{m, k}^{(T)}
\end{equation}
where $R_{m, k}^{(T)}$ represents the final rating of model $m$ after iterating through all $T = 10,800$ matches in the $k$-th shuffled permutation. 

In our experiments, we set $N=10$ as the shuffle count used for the reported audit statistics. To empirically quantify residual order sensitivity, we measured the standard deviation of final ratings across the permutations:
\begin{equation}
    \sigma_m = \sqrt{ \frac{1}{N} \sum_{k=1}^N \left( R_{m, k}^{(T)} - \bar{R}_m \right)^2 }
\end{equation}

Tracking this distribution across both the primary (Llama 3.3 70B) and validation (Qwen 2.5 72B) judges for all 9 contestants yielded the following internal stability metrics on a base 1000-point scale:
\begin{itemize}
    \item \textbf{Maximum Variance:} Bounded strictly at $\sigma_{max} = 37.5$ Elo points across all models.
    \item \textbf{Mean Variance:} Clustered tightly around $\bar{\sigma} \approx 29.5$ Elo points.
\end{itemize}

Given that inter-model spreads on the leaderboard exceed 200 points, this stringent empirical result ($\sigma < 38.0$) indicates that ordering effects are small relative to between-model separation in this study.

\end{document}